\documentclass[10pt,a4paper]{article}

\usepackage[margin=1.5cm,footskip=0.75cm]{geometry}

\usepackage[table]{xcolor}

\usepackage{mathptmx}
\usepackage[T1]{fontenc}
\usepackage{microtype}

\usepackage{amssymb}

\usepackage{graphicx}

\usepackage{lineno}
\usepackage{ragged2e}
\usepackage{setspace}

\usepackage{booktabs}
\usepackage{multirow}
\usepackage[flushleft]{threeparttable}

\usepackage{float}

\usepackage{amsmath}

\usepackage{siunitx}
\DeclareSIUnit\inch{in.}
\DeclareSIUnit\sample{S}

\usepackage{subfiles}

\usepackage[hidelinks,pdfpagemode=UseNone]{hyperref}
\usepackage{bookmark}
\makeatletter
\renewcommand\@seccntformat[1]{}
\makeatother
\usepackage[capitalise]{cleveref}
\usepackage[superscript,noadjust]{cite}

\usepackage{newfloat}
\usepackage{caption}
\DeclareCaptionLabelSeparator{bar}{ | }
\DeclareFloatingEnvironment[fileext=sfig,name=Supplementary Fig.]{supplementaryfigure}
\crefname{supplementaryfigure}{Supplementary Fig.}{Supplementary Figs.}
\usepackage{xcolor}

\begin{document}

\raggedright

\bfseries \huge
\centering
TumorMap: A Laser-based Surgical Platform for 3D Tumor Mapping and Fully-Automated Tumor Resection
\\[18pt]

\centering
\normalfont \normalsize
Guangshen Ma, PhD,\textsuperscript{1,2 $^{\dagger}$, *}
Ravi Prakash,\textsuperscript{1, $^{\dagger}$, *}
Beatrice Schleupner,\textsuperscript{3}
Jeffrey Everitt, DVM,\textsuperscript{4}
Arpit Mishra, PhD,\textsuperscript{1} \\
Junqin Chen, PhD, \textsuperscript{1} 
Brian Mann, PhD, \textsuperscript{1}
Boyuan Chen, PhD, \textsuperscript{1}
Leila Bridgeman, PhD, \textsuperscript{1}
Pei Zhong, PhD, \textsuperscript{1} \\
Mark Draelos, MD, PhD, \textsuperscript{2,5} 
William C. Eward, DVM, MD \textsuperscript{3}
and
Patrick J. Codd, MD\textsuperscript{1,6, $^{\dagger}$}
\\[18pt]

\itshape
\textsuperscript{$*$}G. Ma and R. Prakash contributed equally to this work. \\
\textsuperscript{$^{\dagger}$} Corresponding email: guangshe@umich.edu, ravi.prakash@duke.edu, patrick.codd@duke.edu. \\
\textsuperscript{1}Thomas Lord Department of Mechanical Engineering and Materials Science, Duke University\\
\textsuperscript{2}Department of Robotics, University of Michigan, Ann Arbor\\
\textsuperscript{3}Department of Orthopaedic Surgery, School of Medicine, Duke University\\
\textsuperscript{4}Department of Pathology, School of Medicine, Duke University\\
\textsuperscript{5}Department of Ophthalmology and Visual Sciences, University of Michigan Medical School, Ann Arbor\\
\textsuperscript{6}Department of Neurosurgery, School of Medicine, Duke University
\\[18pt]

\justify

\normalfont
\begin{abstract}
Surgical resection of malignant solid tumors is critically dependent on the surgeon's ability to accurately identify pathological tissue and remove the tumor while preserving surrounding healthy structures.
However, building an intraoperative 3D tumor model for subsequent removal faces major challenges due to the lack of high-fidelity tumor reconstruction, difficulties in developing generalized tissue models to handle the inherent complexities of tumor diagnosis, and the natural physical limitations of bimanual operation, physiologic tremor, and fatigue creep during surgery.
To overcome these challenges, we introduce ``TumorMap", a surgical robotic platform to formulate intraoperative 3D tumor boundaries and achieve autonomous tissue resection using a set of multifunctional lasers.
TumorMap integrates a three-laser mechanism (optical coherence tomography, laser-induced endogenous fluorescence, and cutting laser scalpel) combined with deep learning models to achieve fully-automated and noncontact tumor resection. 
We validated TumorMap in murine osteoscarcoma and soft-tissue sarcoma tumor models, and established a novel histopathological workflow to estimate sensor performance.
With submillimeter laser resection accuracy, we demonstrated multimodal sensor-guided autonomous tumor surgery without any human intervention.

\end{abstract}

\newpage

\section{Introduction} 
Robotic surgery for tumor resection has revolutionized conventional surgical procedures by offering greater precision and dexterity to surgeons, enabling smaller incisions, reduced trauma, faster recovery, and improved patient outcomes with surgical efficiency \cite{connor2020autonomous, dupont2021decade, yip2023artificial}. 
In surgical oncology, the goal of solid malignant tumor resection is to achieve a negative surgical margin, which is directly associated with improved disease-free survival and longevity and a reduced risk of recurrence \cite{stewart2019robotic}. 
Successful tumor removal relies on the surgeon's ability to accurately demarcate the pathological region and precisely resect the malignant tissue along with a discernible margin of surrounding tissue, known as the `surgical margin' \cite{wyld2015evolution}. 
This process focuses on an explicit understanding of the anatomy surrounding the tumor and factors derived from preoperative imaging and intraoperative consultation with pathologists. 
In essence, surgeons are tasked with generating an intricate 3D tumor reconstruction to enhance their understanding of the tumor's size, shape, and location relative to surrounding vital structures in an ever-changing and high-stakes surgical environment. 

However, the nature of three-dimensional (3D) tumor reconstruction poses significant challenges in intraoperative diagnosis, perception, and tumor resection.
First, interpretation of pathological boundaries for tumor removal still heavily relies on surgeons' subjective evaluations informed by preoperative images, visual inspection, and tactile feedback \cite{salcudean2022robot, su2022state}.
It is challenging to achieve intraoperative tumor diagnosis and boundary reconstruction using a gold standard of histopathology, as the integration of a histopathological workflow in surgical robotics can easily disrupt surgical flow and create hurdles to seamless data integration.
In addition to the diagnosis problem, tumor perception heavily relies on intraoperative sensing in surgery.
Although preoperative imaging systems, e.g., magnetic resonance imaging (MRI) or computed tomography (CT), provide holistic information on anatomical structures, they are not suitable for dynamic intraoperative imaging guidance due to low temporal resolution and time-consuming workflow \cite{su2022state}. 
Real-time vision systems, such as monocular or stereoscopic cameras, lack the resolution and volumetric capability required to guide instruments for precise tumor surgery \cite{ma2021comprehensive}.
In addition, conventional surgical tools, such as scalpels, forceps, and needles, inherently require direct contact with tissue, and thus can distort the surface geometry and obscure vision (e.g. tool-tissue interaction) for accurate tumor surface reconstruction. 
For tumor resection, conventional bimanual surgical methods present several challenges to achieve precise and accurate tissue manipulation.
Even highly skilled surgeons are affected by physiologic hand tremors at frequencies of $\sim$12 Hz \cite{verrelli2016intraoperative}, which are amplified in procedures involving precise and delicate interactions with small tissues \cite{fargen2016factors}. 
These existing challenges in tumor surgery, including perception, tumor interpretation, and tissue manipulation, highlight the need for a new surgical robot system to jointly perform real-time diagnosis, high-fidelity volumetric perception, and noncontact tumor removal to overcome current shortcomings and advance oncologic care. 

In light of these challenges, the development of a fast and efficient tumor diagnosis approach is an important first step in robotic-assisted surgical oncology. 
During (or prior to) the surgery, histological examination of biopsied tissues remains the gold standard for tumor diagnosis \cite{mullhaupt2011tumor, o2014effect}. 
However, traditional biopsy-based workflows inevitably disrupt intraoperative continuity, alter tissue geometry, and require manual tracking of excised samples \cite{low2018fluorescence}. 
Recent advances in endogenous fluorescence-guided surgery have enabled rapid, noncontact intraoperative imaging that helps surgeons locate tumor regions while preserving tissue integrity \cite{judy2015quantification, sperber2024blinded, tucker2020creation, tucker2021creation, tucker2022creation, zachem2025intraoperative}. 
This method leverages naturally occurring biomarkers within tissues to detect metabolic and compositional differences between healthy and malignant regions under a single-spot laser excitation \cite{sperber2024blinded}, thus providing label-free, real-time visualization without physical disruption. 
It presents a promising direction for integrating laser-based optical biosensors directly into robotic surgical systems for tumor identification and boundary mapping. 
Although the endogenous fluorescence technique uses a focused laser beam for point-based classification, leveraging the fast scanning robot capability and high-fidelity perception systems can enable time-efficient tumor mapping.

While effective tumor reconstruction models in surgical oncology require high efficiency and precision, tumor resolution becomes another critical factor in providing detailed structures for surgical planning and visualization. 
Thus, the 3D tumor perception system that can be coordinated with point-based tumor diagnosis becomes a critical component of system integration.
Conventional 3D sensors, including monocular cameras, structured light cameras, and dual-camera systems, are limited to surface-level information and lack the volumetric resolution required to reconstruct tumor structures with submillimeter accuracy \cite{schoob2017stereo, zhou2019real}. 
These challenges make optical coherence tomography (OCT), a non-invasive imaging modality that uses low-coherence interferometry to detect reflected light and generate depth-resolved (volumetric) tissue maps with micron-level precision \cite{huang1991optical, drexler2008optical}, suitable for intraoperative tumor perception. 
Recent adoptions of OCT in ophthalmic and microsurgical applications have demonstrated its effectiveness in solving fine anatomic details for surgical planning \cite{ahronovich2021review, ma2025robotics}. 
Thus, using OCT with lasers (e.g., laser-induced fluorescence and laser scalpel) in automated robotic surgery shows a promising direction to provide realtime and high-resolution subsurface imaging and overcome the limited visibility problem of contact-based scalpels \cite{gunalan2023towards}.

To work efficiently with the perception and diagnosis system, noncontact tumor resection is needed to preserve surface geometry during tissue manipulation, which motivates the use of laser scalpels. 
Existing surgical platforms, such as the da Vinci Research Kit for automated surgery \cite{kim2025srt}, the Da Vinci surgical system \cite{d2021accelerating} for remote surgery and the smart tissue autonomous robot (STAR) system \cite{saeidi2022autonomous} for intelligent laparoscopic suturing, use contact-based surgical tools (e.g., needles and forceps) for tissue manipulation.
Relying on complex actuators to handle biological tissues (e.g., cutting and grasping), these tools can inevitably change the tissue geometry through direct contact and cause vision occlusion of intraoperative sensors. 
Energy-based scalpels, such as lasers, provide a unique alternative to traditional scalpels due to their tunable properties for tissue resection and the ability to precisely direct energy delivery to selective anatomical structures. 
The laser-tissue interaction process inherently performs hemostasis and minimizes surgical trauma resulting from bleeding \cite{lee2022end, zhu2019dual}.
Recent work has shown the versatility of the integration of laser scalpels in robotic platforms for free-form and fiber-guided applications \cite{lee2022end, abdelaziz2024fiberbots, pacheco2024automatic, york2021microrobotic,ma2019novel,ma2021stereocnc}.
Considering the noncontact nature of laser-tissue interaction (to preserve tissue geometry) and the free-form system integration, the laser scalpel has become a promising tool in surgical robot systems. 

%%%%%%%%%%%%%%%%%%%%%%%%%%%%%%%%%%%%%%%%%%%%%

\begin{figure}[H]
\centering
\includegraphics[width = 0.72\textwidth]{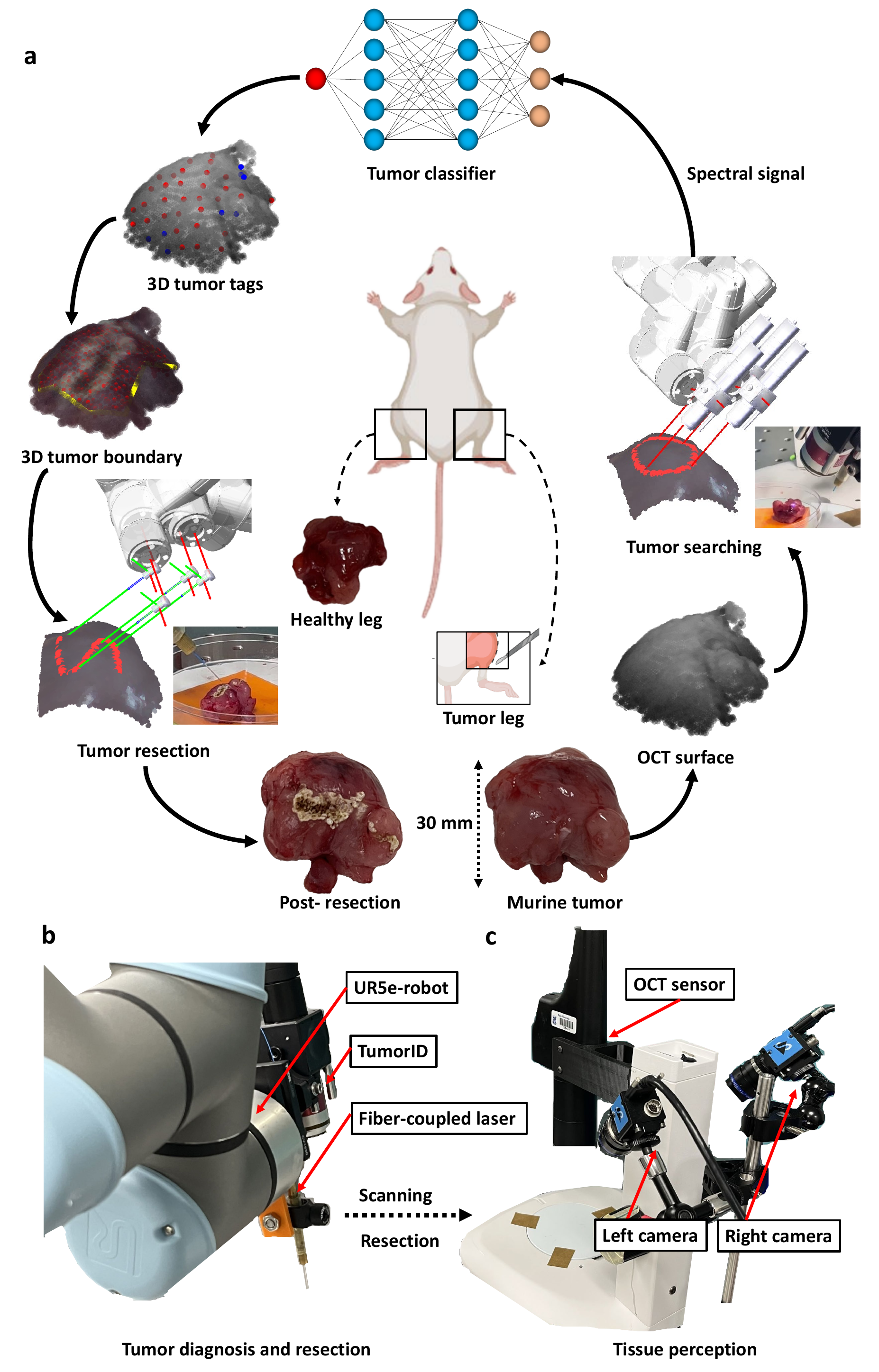}
\caption{
% (in a) 
\textbf{Overview of the TumorMap system and the fully-automated murine tumor resection workflow.} 
This system incorporates two sub-modules for tissue diagnosis and resection (in \textbf{b}) and tissue perception (in \textbf{c}). 
\textbf{a:} The workflow of 3D tumor map formulation. 
First, tumors are induced in the legs of genetically engineered mice.
Once the tumor grows to the expected palpable size, the tissue of interest is dissected to extract the superficial tumor targets from the right leg while keeping the left leg as a healthy (control) sample.
Based on the tumor and healthy datasets (Method.~\ref{murine_tumor_classification}), a tumor classifier is trained to later allow tumor boundary prediction during the realistic experiment. 
For the offline experiment, the tumor is scanned by the OCT sensor for surface reconstruction. 
The laser-induced endogenous fluorescence sensor (also referred to as \textbf{``TumorID"}, i.e., tumor identification sensor) is controlled by the robot to achieve tumor searching based on a raster scanning pattern. 
Each collected spectrum data is passed through the tumor classifier to generate a single-spot tumor label.
This formulates a sequence of 3D tumor tags.
A 3D tumor boundary is created from the tumor tags, which incorporates colorized, pathological, and geometric information, leading to the formulation of a tumor boundary based on the convex hull algorithm \cite{yap2013quantitative} (Method \ref{method_model_tumor_roi_geometry}).
Finally, the tumor boundary is post-processed to formulate a region for automated tumor removal using a fiber-coupled laser scalpel. 
\textbf{b:} The hardware setup for the tissue diagnosis and resection system. 
This system incorporates a 6-DOF UR5e robot arm, a laser-induced endogenous fluorescence sensor (TumorID), and a fiber-coupled cutting laser scalpel with a side-by-side configuration.
The working distance between the TumorID laser to the tissue is approximately 56.3 mm. 
This allows for efficient implementations of robot kinematics and trajectory modeling with a single robot arm,   
\textbf{c:} The tissue perception system of a $1310~nm$ tabletop OCT system with a fixed dual-camera system relative to the surgical scene. 
This system is used to capture the geometric and colorized information of the murine tumor \cite{BioRender}.
}
\label{fig_1_system_main}
\end{figure}

In this work, we present \textbf{''TumorMap''}, a novel surgical platform to leverage the versatility of lasers with complementary sensing modalities for fully-automated tumor diagnosis, reconstruction, and resection (Fig.~\ref{fig_1_system_main}a). 
TumorMap incorporates a 6-DOF robot-controlled diagnostic system using a noncontact endogenous fluorescence technique with a rigidly registered fiber-coupled laser scalpel to achieve free-space tissue resection, as well as a 3D perception system based on OCT imaging techniques and a dual camera system. 
With complementary sensing modalities, we developed a point-based tumor classifier for the endogenous fluorescence sensor, coupled with high-fidelity information from the OCT and cameras to create a \textbf{``3D multimodal tumor map''}, which encoded colorized, pathological, and geometric information on a unified surface to determine the tumor boundary for laser resection. 
We also proposed a computational framework to achieve multi-laser and sensor calibration as well as robot modeling for surgical planning and resection.
To demonstrate TumorMap's feasibility for fully-automated surgery, we conducted comprehensive experiments on tissue phantoms, \textit{ex vivo} tissue samples, and murine sarcoma tumors \cite{lazarides2016fluorescence}.
We also validated the efficacy of the deep learning model for tissue diagnosis through a novel histopathological analysis workflow. 
The successful development of TumorMap showed promising applications in surgical oncology and general 3D modeling of soft tissue in biomedical research.

\section{Results}
\subsection{TumorMap Platform}
The TumorMap platform incorporates two synergistic modules to formulate tissue diagnosis, perception, and resection systems.
The first module is a robotic suite that integrates a laser-induced endogenous fluorescence sensor for tissue diagnosis (referred to as \textbf{TumorID}: single-point tumor identification \cite{tucker2021creation}) and a fiber-coupled laser scalpel for tissue resection based on a parallel configuration (Fig.~\ref{fig_1_system_main}b, Method \ref{method_tumormap_framework_tissue_diagnosis_system} and \ref{method_tumormap_framework_tissue_resection_system}).
This allows free-space laser movements with arbitrary scanning patterns and provides enhanced flexibility and expanded workspaces for tumor searching and resection.
The second module comprises a high-fidelity perception system that consists of a tabletop OCT sensor and a calibrated dual-camera setup to create a volumetric and colorized reconstruction of the tissue object (Fig.~\ref{fig_1_system_main}c and Method \ref{method_tumormap_framework_tissue_perception_system}).
System calibration was performed to jointly align the robot end-effector suite (TumorID and fiber-coupled laser) and the dual-camera system with the OCT sensor frame (Method \ref{method_system_calibration}).

\subsubsection{Automated tumor resection workflow}
\begin{figure}[H]
\centering
\includegraphics[width = 0.99\textwidth]{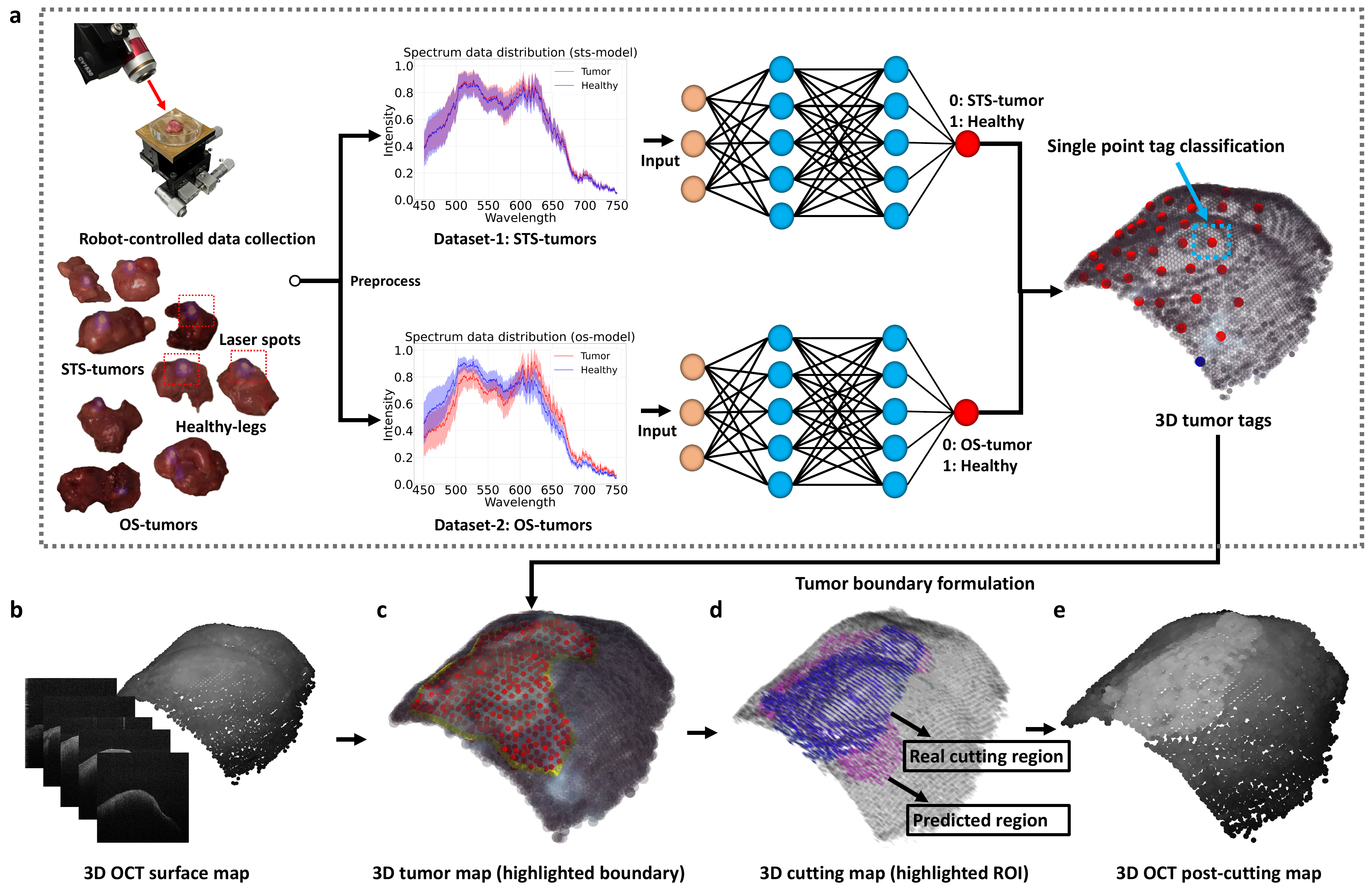}
\caption{
\textbf{Automated tumor resection workflow}: from tumor boundary estimation to resection.
\textbf{a:} The offline workflow of building the tumor classifiers by using the TumorID sensor.
Before the online tumor resection, the tumor classifiers are trained with the TumorID datasets of the soft tissue sarcoma (STS) and osteosarcoma sarcoma (OS) murine models (Method.~\ref{method_doe_murine_robot_data_collection}).  
Specifically, the dissected tumors and healthy samples are first placed at the surgical site.
The TumorID is controlled by the robot to collect spectrum data following a raster scanning. 
The tumor classifier is represented by a multilayer perceptron model for binary classifications and used for online tumor resection (in step \textbf{c}). 
\textbf{b:} OCT-guided tumor reconstruction. 
The tumor is scanned with a tabletop OCT and a 3D surface is reconstructed from sequential B-scan images. 
\textbf{c:} Sensor information alignment from the OCT sensor and the color image through the camera-to-oct calibration in Method \ref{method_calibration_oct_to_camera} (red dots: points in the tumor region; yellow line: tumor boundary).
The tumor boundary is overlaid with the colorized OCT map for intraoperative planning and visualization.
\textbf{d:} 3D tumor boundary and cutting region. 
A sparse tumor map is converted to a 3D tumor boundary following the Method.~\ref{method_model_tumor_roi_geometry}.  
A 3D laser trajectory is generated through the proposed robot kinematics solver (Method.~\ref{method_system_model_ik_and_traj_plan}) to visit target points within the predicted tumor boundary (purple: predicted region; blue: realistic cutting region). 
\textbf{e:} The post-resection OCT surface shows unique pixel features of the cutting region after laser ablation. 
The distinct pixel intensities (lighter gray color) are generated due to tissue coagulation.  
}
\label{fig_2_integrated_workflow}
\end{figure}

Based on the two-module configuration, TumorMap enables a fully-automated workflow (Fig.~\ref{fig_2_integrated_workflow}) for tumor reconstruction, mapping, and resection (\textbf{Movie~S2}). 
This integrated workflow focuses on the creation of a \textbf{3D multimodal tumor map} that simultaneously encodes geometric structure, realistic visual appearance, and pathological signatures (Fig.~\ref{fig_2_integrated_workflow}a-d).  
Initially, the perception system captures tissue surface data using OCT and the dual-camera system, generating a co-registered, colorized 3D surface map (Method~\ref{method_tumormap_framework_tissue_perception_system}). 
The TumorID then performs a raster scan across the tissue surface and captures point-wise fluorescence spectra. 
Each spectrum is processed by a tumor classifier represented by a multilayer perceptron model to assign pathological labels to the point tags (Method~\ref{method_tumor_map_diagnosis_for_map_formulation}). 
These spatially tagged classification results are incorporated into a 3D sparse map (Fig.~\ref{fig_2_integrated_workflow}b-d). 
The convex hull algorithm is then applied to construct the tumor boundary (Method \ref{method_model_tumor_roi_geometry}) and subsequently generate a 3D resection region \cite{yap2013quantitative}. 
The region within the tumor boundary is used to calculate the laser cutting trajectory using our proposed kinematics solver tailored for tumor resection (Method \ref{method_system_model_ik_and_traj_plan}). 
This tumor boundary reconstruction workflow preserves the anatomical fidelity of the tissue by minimizing contact and deformation, facilitating precise surgical planning and the execution of laser resection. 
In this study, TumorMap was validated through comprehensive experiments on tissue phantoms (Section.~\ref{main_phantom_section}), artificial tumors in \textit{ex vivo} porcine and chicken tissues (Section.~\ref{main_exvivo_section}), and genetically engineered murine models of soft tissue sarcoma (STS) and osteosarcoma (OS) tumors (Section.~\ref{section_mice_resection}), establishing its feasibility in a variety of clinically relevant scenarios. 

%%%%%%%%%%%%%%%%%%%%%%%%%%%%%%%%%%%%%%%%%%%%%%%%%%%%

\subsection{System accuracy quantification with phantom experiments}
\label{main_phantom_section}
To characterize TumorMap’s accuracy, we conducted quantitative experiments with custom-designed tissue phantoms.  
These experiments evaluated the accuracy and precision of the system in point location, trajectory tracking, and tumor boundary reconstruction.
Phantom models were fabricated with fiducial markers, S-shaped trajectories, and demarcated spherical regions, which contained well-defined surface targets that could easily be segmented from the OCT data to serve as ground truth (e.g., the black targets absorb most of the light and thus create unique pixel features in the image). 
Depending on the experiment, the objects were scanned by the OCT (fiducial marker and trajectory tracking) or TumorID sensor (region tracking) to locate the targets. 
Optimal robot trajectories were planned using our proposed kinematics solver to trace these targets (Method \ref{method_system_model_all}). 
To evaluate generalization across hardware configurations, we repeated the experiments with three laser modules: a class I green laser diode (wavelength $\lambda=$ 650~nm), a class III B TumorID laser diode ($\lambda=$ 405~nm), and a fiber-coupled class IV laser (wavelength $\lambda=$ 1940~nm). 
All laser modules were installed and calibrated following a similar method (Method.~\ref{method_system_calibration}). 
For error measurement, the 3D surfaces were projected onto the 2D plane along the Z-axis of the OCT frame to calculate the point-to-point errors and the offsets between regions. 
For statistics, the two-sample t-tests were used to compare the results of various groups (laser types).

\subsubsection{Fiducial marker tracking}
The marker tracking experiment aims to assess the point-targeting accuracy of the TumorMap system in an arbitrary 3D space. 
The fiducial markers were printed on a \textbf{$3 \times 3$} meshgrid with unique color-coded signatures, each positioned at varying heights to demonstrate the generalization of the system (Fig.~\ref{fig_3_system_exp}a). 
The OCT sensor was used to scan and reconstruct the center of the marker (as targets).
Successful execution was indicated by a minimal deviation between the predicted laser spot and the targets.
The statistical results in Fig.~\ref{fig_3_system_exp}f-g demonstrate that TumorMap consistently achieved submillimeter accuracy (average error < 1.0 mm) across all laser types, with average errors of $0.57 \pm 0.19$ mm (green diode), $0.51 \pm 0.14$ mm (TumorID, blue diode), and $0.19 \pm 0.11$ mm (fiber-coupled), and the root mean square errors (RMSE) of \textcolor{black}{0.60 mm, 0.53 mm, 0.22 mm}, respectively. 
The results of the two-sample t-test indicate significant differences between the fiber-coupled laser and the two laser diodes ($p < 0.001$), but no significant differences between the green and TumorID diode laser ($p > 0.05$). 
This shows that our proposed kinematics solver can consistently and reliably calculate accurate trajectories to track targets. 
The variations likely stem from differences in laser spot estimations, the OCT-to-laser calibration, and the ablation footprint of the fiber laser.

\subsubsection{Trajectory tracking}
To evaluate continuous motion tracking of 3D paths, we verified TumorMap using a trajectory tracking experiment.
An S-shaped trajectory segmented from the OCT data served as the 3D targets (Fig.~\ref{fig_3_system_exp}b). 
The trajectory accuracy was evaluated by computing the point-to-target error (point correspondence given by the closest neighbors) between the predicted path and the OCT-derived targets.
RMSEs are reported as 0.33 mm (laser diode), 0.17 mm (fiber-coupled laser), and 0.41 mm (TumorID), with a maximum error of $<$ 1.25 mm. 
The two-sample t-test reveals statistically significant differences for all three combinations of laser diode versus fiber-coupled laser ($p<0.001$), laser diode versus TumorID laser ($p<0.05$), and fiber-coupled laser versus TumorID laser ($p<0.001$). 
The largest discrepancies are observed with the fiber-coupled laser, likely due to post-ablation artifacts (heat propagation around the laser spot center) that complicate accurate spot localization. 
Nevertheless, all systems maintain small average errors $\leq 0.50$ mm ($0.26 \pm 0.20$ mm for green laser-diode, $0.14 \pm 0.10$ mm for the fiber-coupled laser, and $0.31 \pm 0.27$ mm for the TumorID laser).
This demonstrates TumorMap's ability to accurately traverse arbitrary trajectories in phantom models that mimic realistic tissue structures (e.g., blood vessels or other skin targets in dermatology). 

\subsubsection{Spherical region tracking}
\label{system_sphere_region_checking}
After demonstrating the system's capability on points and unrestricted trajectories, we evaluated the tumor mapping workflow in Fig.~\ref{fig_2_integrated_workflow} using spherical phantom regions that mimicked realistic tumors.
Unlike the previous two tasks, the targets were generated from the TumorID sensor, while the ground truth was created by the OCT data.
The TumorID sensor performed a raster scan of the phantom surface and predicted a spherical ``tumor boundary", based on a region of $13.0 \times 13.0~mm^2$ with a step size of 1.44 mm (total 100 data points).
Due to the unique fluorescence characteristics of artificial tumors (reduced average spectral intensity in the black region), a simple threshold-based tumor classifier was used. 
The average spectral intensity less than 0.50 was categorized as ``tumor", and vice versa for the healthy label.
The points classified as ``tumor" were used to form the tumor boundary that was further processed to generate a laser path (Method~\ref{method_model_tumor_roi_geometry}) .

\begin{figure}[H]
\centering
\includegraphics[width = 0.80\textwidth]{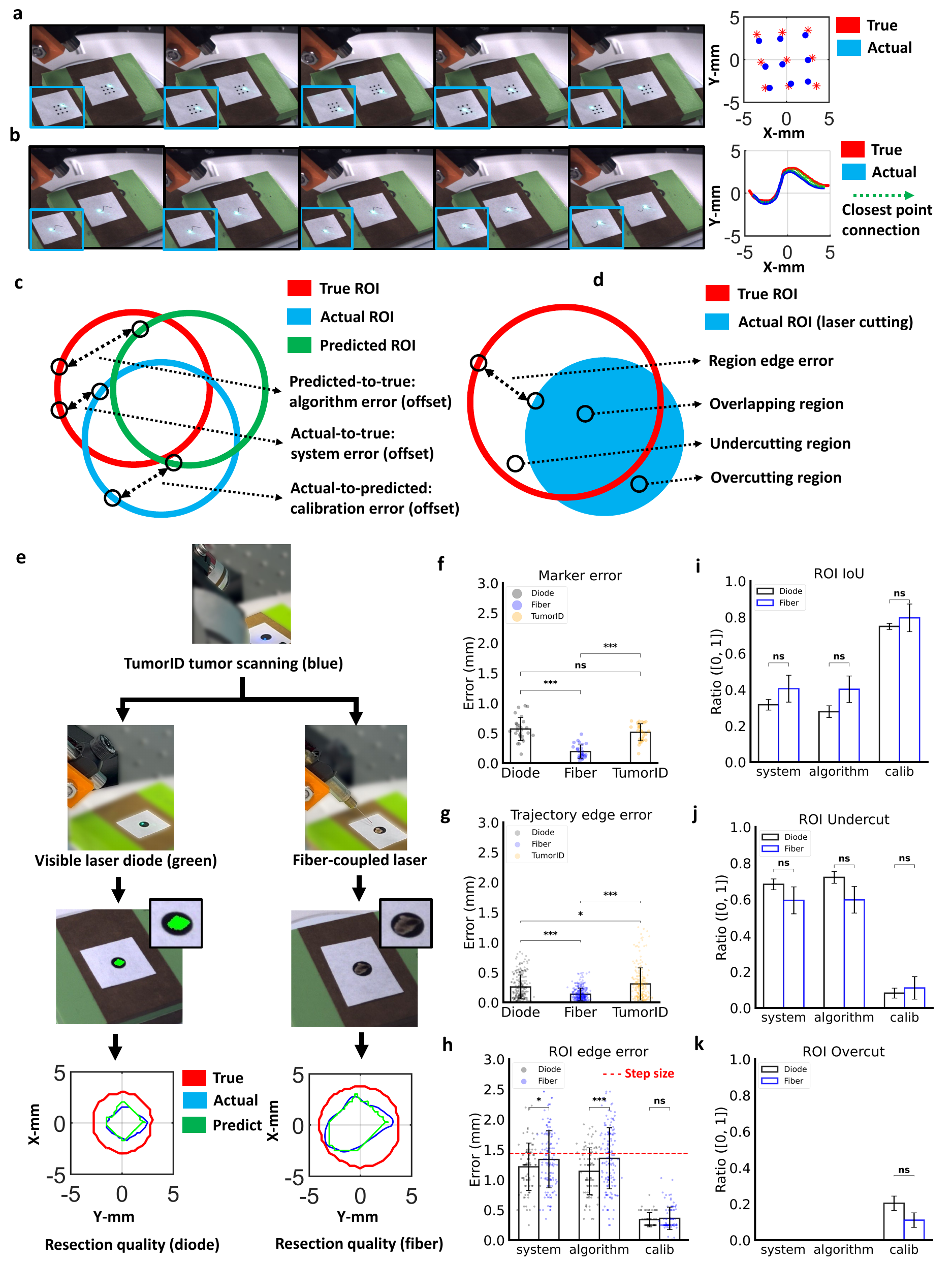}
\caption{
\textbf{Overview of the system accuracy quantification.}
These phantom experiments aim for tracking fiducial marker, 3D trajectory, and the spherical region (ROI) by using three lasers for generalization and reliability testing.
Three laser modules were applied (a fiber-coupled laser with a wavelength of 1940 nm, a green laser diode with a wavelength of 650 nm, and a TumorID laser diode with a wavelength of 405 nm).
\textbf{a:} An example of fiducial marker using the laser diode (visible green dot). 
The global and local views of the images show the error distribution of TumorMap to achieve end-to-end target tracking.
\textbf{b:} 
An example of trajectory tracing using the green laser diode. 
The point-to-target error is calculated by using the correspondence index of nearest neighbors.
\textbf{c:}
The geometric relationship among the true-ROI, prediction-ROI, and actual-ROI, and the definitions of algorithm, system and calibration errors.  
\textbf{d:} The geometric relationship among the overlapping, undercutting and overcutting regions, as well as the point-to-point edge error based on the correspondence of the nearest points.
\textbf{e:} The experimental workflow of using TumorID to determine ``artificial tumor regions" and laser scalpels (laser diode and fiber-coupled laser) to trace targets (laser tools highlighted in figures). 
The laser diode generates a visible laser spot on the surface.  
This can be detected by the camera and thus its center can be accurately estimated for error calculation. 
The fiber-coupled laser generates a resected pattern on the surface and creates a unique feature to differentiate it from the pre-ablated surface. 
\textbf{f:} 
The ``Marker error" graph shows the error barchart between the real-cutting and the true centers.
\textbf{g:} The ``Trajectory edge error" graph depicts the error of the point-based trajectory. 
\textbf{h:} For the subcategories, the barchart ``ROI edge error" shows the point-based boundary error (step size = 1.44 mm highlighted) between the ablated region and the true one \textbf{(system error: actual-to-true offset, algorithm error: predicted-to-true offset, calibration error: actual-to-predicted offset)}. 
\textbf{i:} 
The ``ROI IoU" describes the ratio of the overlapping region versus the true one.  
\textbf{j:} 
The ``ROI Undercut" shows the ratio of the region offset between the predicted region and the true region that should be cut (but has not been cut).
\textbf{k:} 
The ``ROI Overcut" denotes the ratio of the region offset between the predicted region and the true region that should not be cut (but has been cut).
Statistical index: $*: p < 0.05$, $**: p < 0.01$, $***: p < 0.001$, $ns$: not significant). 
}
\label{fig_3_system_exp}
\end{figure}

\noindent
\textbf{Error evaluation metrics:}
\label{error_roi_evaluation_metric}
To evaluate the system's performance with error analysis (Fig.~\ref{fig_3_system_exp}c-d), we define three terms: predicted region (region derived from classifications), actual region (region traced by the laser), and true region (target regions segmented from OCT or other labeled regions). 
The errors were quantified as: the algorithm error (offset between predicted and true regions) reflects the accuracy of the entire algorithm framework; the system error (offset between actual and true regions) indicates the end-to-end system accuracy to trace the target; the calibration error (offset between predicted and actual regions) captures discrepancies between the intended and achieved laser positions.
For region analysis, performance was assessed using 1) point-to-point edge error between two regions, 2) intersection-over-union (IoU) to quantify overlap between predicted and the corresponding true regions (i.e. $\frac{Area ~ of ~ Overlap}{Area ~ of ~ Union}$), 3) undercutting ratio $\frac{Area ~ of ~  Undercutting}{Area ~ of ~ Target}$, and 4) overcutting ratio $\frac{Area ~ of ~ Overcutting}{Area ~ of ~ Target}$ (Fig.\ref{fig_3_system_exp}c–d). 
The undercutting and overcutting reflect potential complications from excess or insufficient tissue removal by the laser scalpel.

Fig.~\ref{fig_3_system_exp}h reveals that the average system and algorithm errors were greater than 1.0 mm for fiber-coupled and diode lasers.
For most data points, the errors are smaller than 1.44 mm (laser step size). 
This suggests that even a single misclassification (of a step) can introduce ambiguity around the tumor margin, contributing to the system error on the scale of the step size.
Although the calibration errors do not show significant differences ($p > 0.05$), both the system ($p < 0.001$) and the algorithm errors ($p < 0.05$) differ significantly, which is mainly due to postablation measurement inconsistencies.
The average system and algorithm IoUs are $\approx$ $30\%$, whereas calibration IoU is significantly higher ($ \approx 80\%$), indicating a strong alignment between predicted and executed resection. 
In ROI IoU, overcutting and undercutting metrics (Fig.~\ref{fig_3_system_exp}i-k), no significant differences are observed between the three region comparisons, and the average overcutting ratios are greater than IoU.  
These findings suggest that the system favors undercutting over overcutting, maintaining precision around the intended resection zone.  
In summary, TumorMap demonstrates reliable and generalizable performance in tracking artificial tumor targets using both TumorID and fiber-coupled resection lasers. 

\subsection{\textit{Ex vivo} tissue validation with artificial tumors}
\label{main_exvivo_section}
To evaluate the tumor resection workflow in Fig.~\ref{fig_2_integrated_workflow} under more realistic conditions, we conducted \textit{ex vivo} experiments using porcine and chicken tissues embedded with colored dyes to simulate artificial tumors (\textbf{Movie~S4}).
These dyes provided clear visual boundaries and distinct fluorescence signatures due to present chromophores, allowing straightforward segmentation of true target regions from color images and the development of a simple threshold-based classifier to distinguish tumors and healthy tissues.
The goal of using a simple classifier in this \textit{ex vivo} experiment is to demonstrate the system feasibility in achieving accurate tissue scanning and resection.
We further demonstrate the integration of a realistic tumor classifier in the next section (Section.~\ref{section_mice_resection}). 

\subsubsection{\textit{Ex vivo} artificial tumor classifier}
\label{exvivo_tumor_classifier}
To formulate the \textit{ex vivo} dataset, we prepared two tissue samples to represent the ``tumor" and ``healthy" regions.
For tumor data, colored dyes were first placed on the tissue surface to mimic the "artificial tumor regions" (images in the Appendix.~\ref{appendix_exvivo_img_dataset}), and the region size was designed to cover all the data points.
The TumorID sensor performed a surface raster scan to collect spectral data following a scanning region of $13.0 \times 13.0~mm^2$ and a step size of 1.44 mm (total of 100 data points). 
For the healthy dataset, the same workflow was applied, and the data points were collected from purely normal tissue samples without adding color dyes.

To build the tumor classifier, the wavelength ranges were empirically selected as 495 to 570 nm for green and 620 to 750 nm for red.
A binary classifier was developed by defining a confidence range calculated as the average of the tumor and healthy spectral intensity, with upper and lower bounds set at 10\% of the entire min-max range (intensity distribution shown in Appendix.~\ref{appendix_exvivo_classification_graph}).
This improves the reliability of the decisions instead of using a single cut-off threshold. 
A new data point with an average spectral intensity greater than or less than the confidence threshold was assigned with the correct classification labels.   

\subsubsection{\textit{Ex vivo} artificial tumor resection}
To evaluate TumorMap's accuracy in \textit{ex vivo} tissue experiments, we tested two laser modules (laser diode and fiber-coupled laser) following the same tumor search and mapping workflow in Fig.~\ref{fig_2_integrated_workflow}.  
As real tissue resection involves ablation, thermal effects could create "burning scars" or thermal injury in the surrounding tissue, making it difficult to accurately quantify errors from post-ablation artifacts.
Thus, initial validation was performed using the laser diode to demonstrate the algorithm's feasibility with visible laser spots.   
Since TumorID was used for both diagnosis and mapping (not for true boundary segmentation), the true tumor region was segmented from the color image and transformed into the OCT frame to serve as the ground truth. 
The laser spot centers of both the TumorID (blue laser diode) and the green laser diode were identified via intensity-based image processing, transformed into the OCT coordinate frame, and converted to 3D coordinates through system calibration.  
For the tumor resection experiment, the TumorID sensor first scanned a region $13.0 \times 13.0~mm^2$ with a step size of 1.44 mm (total 100 data points) and collected fluorescence data. 
The collected data was classified to generate the results of a sparse tumor map (Section.~\ref{exvivo_tumor_classifier}), a tumor boundary, and subsequently a laser cutting trajectory. 
The same workflow was applied to the fiber-coupled laser experiment, where the realistic ablation patterns were produced on the surface.

\begin{figure}[H]
\centering
\includegraphics[width = 0.83\textwidth]{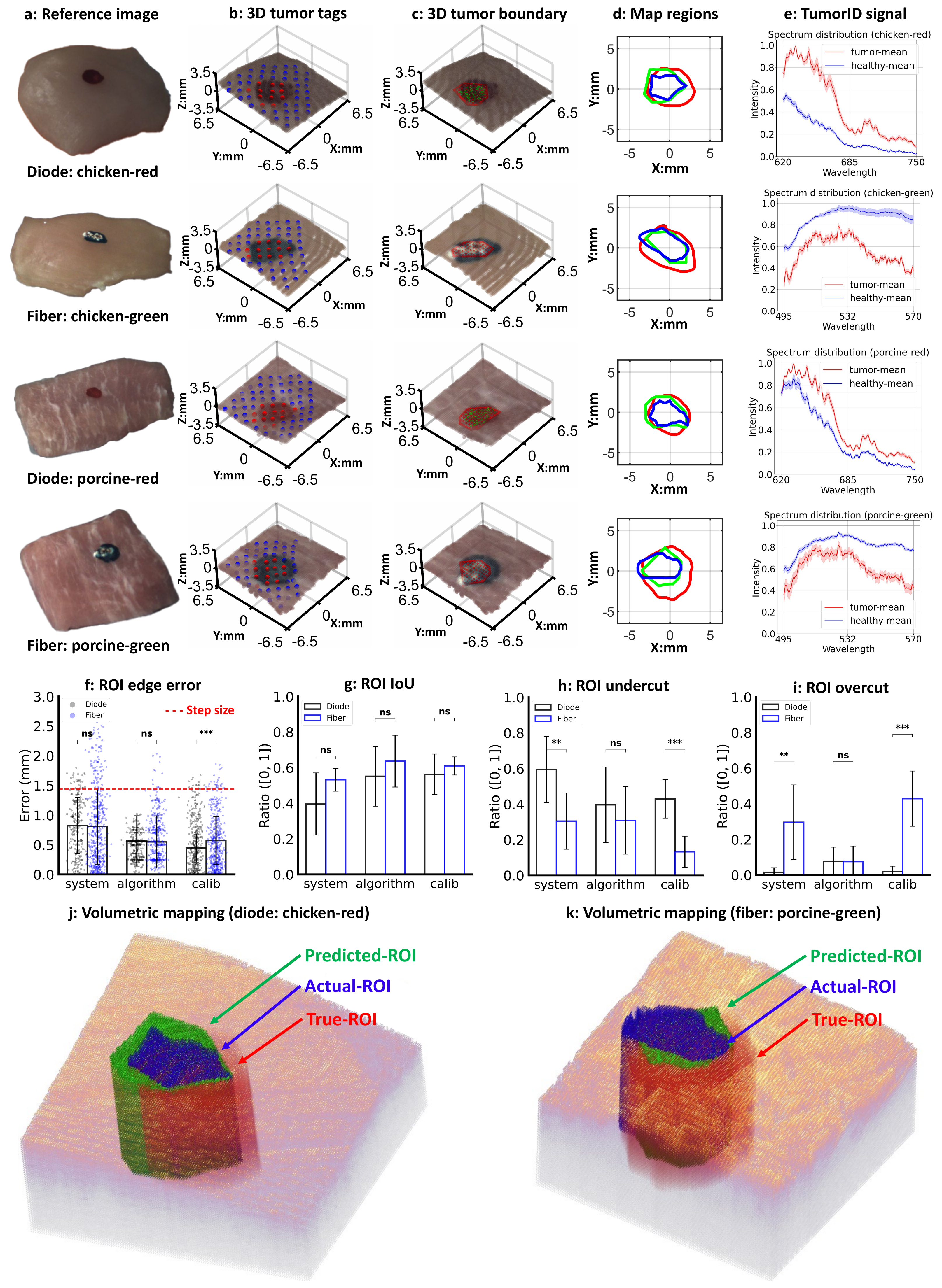}
\caption{
\textbf{\textit{Ex vivo} experiments with the laser diode and the fiber-coupled laser (unit: mm).}
% a 
\textbf{a:} The segmented reference images of the \textit{ex vivo} tissue samples. 
% b. 
\textbf{b:} The tumor tags with classification labels overlapped with the OCT surface reconstruction. 
% c 
\textbf{c:} The actual laser resection signatures overlapped with the post-resection image with the predicted tumor boundary.  
Since the laser diode does not generate cutting patterns, the laser spot is segmented in the image, transformed to the 3D OCT frame, and overlaid on the surface map. 
% d
\textbf{d:} The algorithm evaluation for the predicted tumorous and resection regions (red: true region labeled by operators; green: predicted region from the algorithm framework; blue: realistic resection region).  
% e
\textbf{e:} Spectrum distribution of the TumorID data with various \textit{ex vivo} tissue models. 
The range of wavelengths is selected to show the major difference of the colorized dye to mimic the ``artificial tumors" (495 nm to 570 nm for green color and 620 nm to 750 nm for red color).  
The intensity distribution is shown in Appendix.~\ref{appendix_exvivo_classification_graph}.  
We first calculate the thresholds as the average of the mean tumor and healthy intensities within the wavelength ranges. 
The confidence range of the intensity is defined as the upper and lower bounds of 10\% of the threshold.
% (grey color label).
% 
The average intensity higher than or lower than the confidence bounds is assigned to different classification labels. 
\textbf{f:} The distribution of the edge error for boundary tracking \textbf{(system error: actual-to-true offset, algorithm error: predicted-to-true offset, calibration error: actual-to-predicted offset)}.
\textbf{g:} The distribution of the intersection-over-union (IoU) for the ROI tracking task.
% h
\textbf{h:} The distribution of undercutting ratios for ROI tracking.
\textbf{i:} The distribution of overcutting ratios for ROI tracking.
% i
\textbf{j:} Volumetric mapping of an example for visualization (chicken-red with the diode virtual resection).
% i
\textbf{k:} Volumetric mapping of an example for visualization (porcine-green with the fiber-coupled laser resection).
Statistical index: $*: p < 0.05$, $**: p < 0.01$, $***: p < 0.001$, $ns$: not significant.
}
\label{fig_4_exvivo_exp}
\end{figure}

System evaluation was performed using the same error metrics in Section.~\ref{system_sphere_region_checking}. 
% in average errors 
The ROI edge error (Fig.~\ref{fig_4_exvivo_exp}f) shows no significant difference between the two types of laser for the system and algorithm errors, supporting the system's consistency and reliability in achieving tumor resections without significant variations from resection tools.
However, calibration errors differ significantly ($p < 0.001$), potentially due to challenges in accurately localizing the ablated centers caused by tissue disruption from the fiber laser. 
The measured boundary errors for the diode laser are $0.83 \pm 0.47$ mm (system error), $0.57 \pm 0.43$ mm (algorithm error), and $0.45 \pm 0.24$ mm (calibration error), while for the fiber-coupled laser they are $0.81 \pm 0.65$ mm (system error), $0.55 \pm 0.44$ mm (algorithm error) and $0.57 \pm 0.40$ mm (calibration error), respectively. 
These average error values remain below the laser step size of 1.44 mm.
These results indicate that even isolated misclassifications can propagate into measurable system errors, especially in low-resolution scans.
While the classifier demonstrates reliable performance, the granularity of a scanning region becomes the limiting factor for accuracy and can be tuned with an added time penalty for better results.
The ROI IoU analysis in Fig.~\ref{fig_4_exvivo_exp}g shows no significant difference between the laser diode and the fiber-coupled laser for the system, algorithm, and calibration errors.
This is consistent with the ROI edge error and confirms the system's reliability in achieving tumor boundary mapping and resections.
Variations likely result from single-point misclassifications that can disproportionately impact small-region estimates.  
Undercutting analysis (Fig.~\ref{fig_4_exvivo_exp}h) also indicates that the laser diode shows higher undercutting ratios compared to the fiber-coupled laser.
In addition, the overcutting analysis in Fig.~\ref{fig_4_exvivo_exp}i reveals a significant inconsistency for the fiber-coupled laser in the system ($p < 0.01$) and calibration ($p < 0.001$) error comparisons with the laser diode, but not in the algorithm one. 
The laser-diode study shows significantly smaller overcutting effects than the fiber-coupled laser.  
These inconsistencies are attributed to overirradiation and variability in laser cutting quality, which is influenced by fiber-to-tissue standoff distance, orientation, and resulting post-ablation crater shapes. 
This indicates that system accuracy was affected more by postablation effects than by calibration and prediction algorithms.

In summary, TumorMap exhibits a tendency toward undercutting rather than overcutting, a favorable and safer outcome in the clinical context, as it minimizes the risk of damaging surrounding healthy tissue. 
This behavior reflects conservative calibration and classifier thresholds, although some inaccuracy may come from the ambiguity around the region boundary and the size of the scan step.
Depending on the clinical definition of safe error limit, the step size and classification sensitivity can be modified based on the clinically defined safety margin to balance precision and safety. 
Although with ``optimal tumor classifiers" using colored dyes, the submillimeter accuracy levels in this experiment confirm TumorMap's precision and reliability in targeting artificial tumors of \textit{ex vivo} tissues. 
In the next section (Section.~\ref{section_mice_resection}), a realistic tumor classifier is used from murine sarcoma tumors to demonstrate the feasibility of TumorID sensor in generating tumor boundaries.

\subsection{Murine study with genetically engineered tumor}
\label{section_mice_resection}
The murine experiment aims to demonstrate TumorMap’s translation capability in a realistic surgical scenario for tumor resection, such as soft tissue sarcoma (STS) and osteosarcoma (OS) tumors, two malignant tumors that develop from connective tissues such as muscle and bone \cite{lazarides2016fluorescence, kirsch2007spatially}. 
Following the workflow described in Fig.~\ref{fig_2_integrated_workflow}, we identified the tumor boundary using the TumorID sensor and consequently generated a laser resection trajectory for automated tumor removal \textbf{(Movie~S5)}. 

\subsubsection{Murine data collection}
The first step in developing the tumor classifier involved collecting fluorescence spectral data from murine tumors and healthy tissues.
The murine samples (mice) were induced to develop palpable STS and OS tumors using a standardized tumor induction protocol (Method.~\ref{method_doe_murine_data_preparation}). 
Once tumors were established, superficial skin layers were removed.
Tumor tissues were isolated from the underlying bone to enable data acquisition using the TumorID sensor.

Data were collected from a total of six mice for each tumor model (STS and OS), resulting in paired tumor and healthy samples (\textbf{Movie~S3}). 
To provide labels for model training, an expert manually delineated the pathological region in the tumor-bearing leg, and fluorescence data points within this region were labeled ``tumor" (Appendix.~\ref{appendix_mice_roi_tumor_and_healthy}).  
Data from the contralateral leg, which was anatomically isolated from the tumor site and presumed to be cancer-free, were labeled "healthy". 
For each leg, a raster scan was performed using the TumorID sensor mounted at a 40 ° tilt relative to the OCT Z-axis to minimize interference with perception hardware.
The scans covered a $13.0 \times 13.0~mm^2$ area with a step size of 1.86 mm, generating 64 uniformly spaced fluorescence measurements per sample. Spatial positions and spectra were recorded for each data point.
Thus, each TumorID data capture both the laser-induced endogenous fluorescence spectra and the corresponding spatial coordinates of the laser spots to formulate the dataset.

\subsubsection{Murine tumor classification}
\label{murine_tumor_classification}
In clinical settings, it is rare to observe the co-occurrence of both osteosarcoma (OS) and soft tissue sarcoma (STS) within or adjacent to healthy tissue \cite{demetri2005soft}.
Tumors are typically developed adjacent to healthy tissues, making the binary classification between tumorous and healthy tissue a more relevant modeling approach \cite{demetri2005soft}.
Thus, we developed separate binary classifiers for OS and STS using fluorescence spectra acquired from the TumorID sensor.

A standard vanilla multilayer perceptron (MLP) architecture was used for classification \cite{black2024towards}, with raw spectral data serving as 1D input and expert-annotated labels as output (model architecture in Appendix.~\ref{appendix_MLP_architecture_and_setting}). 
The MLP architecture remained consistent across all experiments.
Prior to training, the input data was normalized with the global mean and standard deviation to reduce non-tumor-specific biases introduced by laser variation and tissue heterogeneity, which was a key challenge involved the inconsistency in data volume between tissue samples. 
To prevent data-rich samples from dominating the model learning process, we restricted the number of usable points from each mouse to no more than three times the global minimum. 
For example, some mouse samples yielded fewer than 15 viable data points due to a small tumor region, while others provided more than 100. 

To promote a rigorous and unbiased evaluation of the tumor classifier, the dataset was organized by mouse ID instead of a single data point, ensuring sample-wise segregation between training and testing sets. 
We used data from six mice for each tumor type.
Four mice were randomly selected for training and two for testing, generating 15 different training-test combinations for the OS and STS groups. 
This data preparation approach prevented data leakage, reduced user selection bias, and allowed analysis of model performance across inter-subject variability. 
Thus, the classification accuracy was attributed to true generalization rather than sample-specific artifacts or class imbalance. 

Our results in Table~\ref{ml_result_40_degree} reveal substantial variability between combinations. 
The performance metrics, such as accuracy, precision, recall, F1 score, and specificity, fluctuate depending on the diversity of tumor subtypes present in the training data. 
While some combinations exceeded 60\% across all metrics (blue-labeled groups), many did not, as expected. 
This variability is attributed to intersample heterogeneity in tumor physiology, arising from differences in tumor growth rate, injection accuracy, immune response, and timing of tissue harvest.
Even with consistent preprocessing, subtle shifts in tumor grade, vascularization, or necrosis can alter fluorescence profiles.
Such heterogeneity reflects, albeit a simplified version of the challenges faced in clinical scenarios, where additional factors such as genetic diversity and anatomical variation add further complexities.

\renewcommand{\arraystretch}{1.5}
\setlength{\arrayrulewidth}{0.25mm}
\begin{table}[h]
\centering
\caption{
\textbf{Murine prediction results with experts' labels as ground truth (non-histopathological analysis)}.  
Data points are grouped according to mouse ID for training and testing datasets.
The red-labeled group is selected for the online tumor resection experiment based on a criterion that the training dataset should cover the majority of the tumor subtypes and statuses (e.g., size, shape, grades, etc.).
The blue-labeled groups represent the acceptable models with a statistical index greater than 0.60 for all statistical metrics of accuracy, precision, recall, F1-score, and specificity. 
TP: counts of true positive; TN: counts of true negative; FP: counts of false positive; FN: counts of false negative.
}
\label{ml_result_40_degree}
\resizebox{\columnwidth}{!}{%
\begin{tabular}{ccccccccccccc}
\toprule
\hline
Group index & Train + Val (counts) & Testing (counts) & TP & TN & FP & FN & Accuracy & Precision & Recall & F1-score & Specificity \\
\midrule
\hline

OS-1 + OS-2 + OS-3 + OS-4 (idx-1) & 209 & 126 & 48 & 29 & 28 & 21 & 0.61 & 0.63 & 0.70 & 0.66 & 0.51 \\
\colorbox{blue!20}{OS-1 + OS-2 + OS-3 + OS-5 (idx-2)} & 201 & 134 & 63 & 35 & 16 & 20 & 0.73 & 0.80 & 0.76 & 0.78 & 0.69 \\
OS-1 + OS-2 + OS-3 + OS-6 (idx-3) & 255 & 80 & 39 & 12 & 16 & 13 & 0.64 & 0.71 & 0.75 & 0.73 & 0.43 \\
\colorbox{blue!20}{OS-1 + OS-2 + OS-4 + OS-5 (idx-4)} & 220 & 115 & 43 & 39 & 12 & 21 & 0.71 & 0.78 & 0.67 & 0.72 & 0.76 \\
OS-1 + OS-2 + OS-4 + OS-6 (idx-5) & 274 & 61 & 17 & 14 & 14 & 16 & 0.51 & 0.55 & 0.52 & 0.53 & 0.50 \\
OS-1 + OS-2 + OS-5 + OS-6 (idx-6) & 266 & 69 & 26 & 10 & 12 & 21 & 0.52 & 0.68 & 0.55 & 0.61 & 0.45 \\
OS-1 + OS-3 + OS-4 + OS-5 (idx-7) & 158 & 177 & 102 & 35 & 32 & 8 & 0.77 & 0.76 & 0.93 & 0.84 & 0.52 \\
\colorbox{red!20}{OS-1 + OS-3 + OS-4 + OS-6 (idx-8)} & 212 & 123 & 77 & 9 & 35 & 2 & 0.70 & 0.69 & 0.97 & 0.81 & 0.20 \\
OS-1 + OS-3 + OS-5 + OS-6 (idx-9) & 204 & 131 & 90 & 9 & 29 & 3 & 0.76 & 0.76 & 0.97 & 0.85 & 0.24 \\
OS-1 + OS-4 + OS-5 + OS-6 (idx-10) & 223 & 112 & 71 & 19 & 19 & 3 & 0.80 & 0.79 & 0.96 & 0.87 & 0.50 \\
OS-2 + OS-3 + OS-4 + OS-5 (idx-11) & 192 & 143 & 24 & 36 & 33 & 50 & 0.42 & 0.42 & 0.32 & 0.37 & 0.52 \\
OS-2 + OS-3 + OS-4 + OS-6 (idx-12) & 246 & 89 & 16 & 24 & 22 & 27 & 0.45 & 0.42 & 0.37 & 0.40 & 0.52 \\
OS-2 + OS-3 + OS-5 + OS-6 (idx-13) & 238 & 97 & 25 & 11 & 29 & 32 & 0.37 & 0.46 & 0.44 & 0.45 & 0.28 \\
OS-2 + OS-4 + OS-5 + OS-6 (idx-14) & 257 & 78 & 17 & 30 & 10 & 21 & 0.60 & 0.63 & 0.45 & 0.52 & 0.75 \\
OS-3 + OS-4 + OS-5 + OS-6 (idx-15) & 195 & 140 & 38 & 7 & 49 & 46 & 0.32 & 0.44 & 0.45 & 0.44 & 0.12 \\

\hline
\hline

STS-1 + STS-2 + STS-3 + STS-4 (idx-1) & 228 & 139 & 23 & 44 & 24 & 48 & 0.48 & 0.49 & 0.32 & 0.39 & 0.65 \\
\colorbox{red!20}{STS-1 + STS-2 + STS-3 + STS-5 (idx-2)} & 265 & 102 & 39 & 34 & 13 & 16 & 0.72 & 0.75 & 0.71 & 0.73 & 0.72 \\
STS-1 + STS-2 + STS-3 + STS-6 (idx-3) & 248 & 119 & 39 & 39 & 10 & 31 & 0.66 & 0.80 & 0.56 & 0.66 & 0.80 \\
STS-1 + STS-2 + STS-4 + STS-5 (idx-4) & 224 & 143 & 42 & 45 & 30 & 26 & 0.61 & 0.58 & 0.62 & 0.60 & 0.60 \\
STS-1 + STS-2 + STS-4 + STS-6 (idx-5) & 207 & 160 & 42 & 47 & 30 & 41 & 0.56 & 0.58 & 0.51 & 0.54 & 0.61 \\
STS-1 + STS-2 + STS-5 + STS-6 (idx-6) & 244 & 123 & 40 & 24 & 32 & 27 & 0.52 & 0.56 & 0.60 & 0.58 & 0.43 \\
\colorbox{blue!20}{STS-1 + STS-3 + STS-4 + STS-5 (idx-7)} & 269 & 98 & 32 & 33 & 16 & 17 & 0.66 & 0.67 & 0.65 & 0.66 & 0.67 \\
STS-1 + STS-3 + STS-4 + STS-6 (idx-8) & 252 & 115 & 38 & 34 & 17 & 26 & 0.63 & 0.69 & 0.59 & 0.64 & 0.67 \\
STS-1 + STS-3 + STS-5 + STS-6 (idx-9) & 289 & 78 & 35 & 11 & 19 & 13 & 0.59 & 0.65 & 0.73 & 0.69 & 0.37 \\
STS-1 + STS-4 + STS-5 + STS-6 (idx-10) & 248 & 119 & 19 & 51 & 7 & 42 & 0.59 & 0.73 & 0.31 & 0.44 & 0.88 \\
\colorbox{blue!20}{STS-2 + STS-3 + STS-4 + STS-5 (idx-11)} & 238 & 129 & 69 & 35 & 14 & 11 & 0.81 & 0.83 & 0.86 & 0.85 & 0.71 \\
\colorbox{blue!20}{STS-2 + STS-3 + STS-4 + STS-6 (idx-12)} & 221 & 146 & 81 & 34 & 17 & 14 & 0.79 & 0.83 & 0.85 & 0.84 & 0.67 \\
STS-2 + STS-3 + STS-5 + STS-6 (idx-13) & 258 & 109 & 78 & 8 & 22 & 1 & 0.79 & 0.78 & 0.99 & 0.87 & 0.27 \\
\colorbox{blue!20}{STS-2 + STS-4 + STS-5 + STS-6 (idx-14)} & 217 & 150 & 75 & 50 & 8 & 17 & 0.83 & 0.90 & 0.82 & 0.86 & 0.86 \\
STS-3 + STS-4 + STS-5 + STS-6 (idx-15) & 262 & 105 & 70 & 15 & 17 & 3 & 0.81 & 0.80 & 0.96 & 0.88 & 0.47 \\

\hline 
\bottomrule
\end{tabular}
}
\end{table}

In this study, deliberate efforts were made to minimize alterations in tissue samples and preserve the inherent challenges of surgical oncology (reflected in the results).  
Though with the same tumor preprocessing steps, different mouse grades and status, and the varieties within the same OS- and STS-tumorous samples can contribute to the inconsistencies of classification performances.
These intraclass variations suggest that some test samples may fortuitously align with training distributions, leading to inflated performance, while others deviate significantly and thus contribute to reduced model reliabilities. 
Despite these challenges, several combinations yield promising results, demonstrating the feasibility of the TumorID platform for efficient point-based classification and 3D tumor mapping. 

\subsubsection{Murine tumor resection}
The murine tumor resection experiment followed the workflow previously established in the \textit{ex vivo} study  (Fig.~\ref{fig_2_integrated_workflow}).
A clinically inspired mouse combination (the red label in Table~\ref{ml_result_40_degree}) was chosen to build the tumor classifier, which covers the widest spectrum of tumor subtypes and morphological diversity.
The OS tumor classifier, although it achieves a respectable accuracy of 0.70 and a precision of 0.69, demonstrated lower specificity scores. 
In contrast, the STS-tumor classifier yields more balanced and higher performance metrics, potentially due to a better representation of STS subtypes in the training set. 
These differences highlight the expected variability in model performance when the tumor subtypes in the test set differ in combination, but not in distribution, from those seen during training, as present in any supervised ML training paradigm. 

\begin{figure}[H]
\centering
\includegraphics[width = 0.93\textwidth]{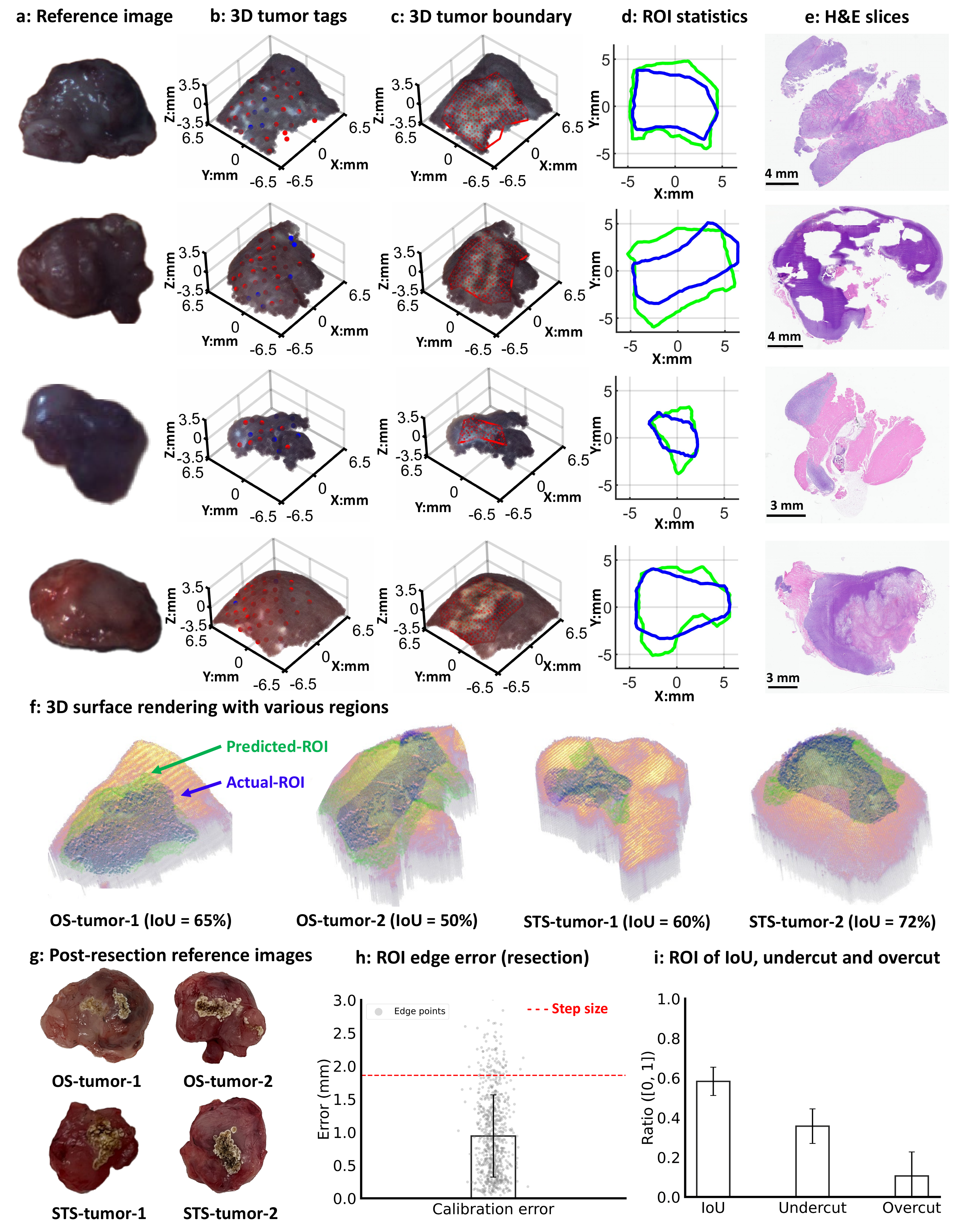}
\caption{
\textbf{Summary of murine tumor resection experiments (unit: mm).}
% a.
\textbf{a:} The segmented reference images for murine tumor models.
% b.
\textbf{b:} Sparse tumor tags with classifications of the scanned points (tumor versus healthy).
This map combines sensor information from the OCT (3D surface reconstruction), the color camera (color information), and the TumorID sensor (tumor classification).
The red dot indicates the tumor tags from predictions, and the blue dot indicates the healthy one.
% c
\textbf{c:} 3D tumor map with highlighted boundary. 
The tumor boundary was formulated with the tumor tags ( from \textbf{(b)} ) using the convex hull algorithm to connect the edge points (Method \ref{method_model_tumor_roi_geometry}). 
% d
\textbf{d:} ROI evaluations of the predicted and resection regions. 
The green color represents the prediction, and the blue one shows the realistic resection region. 
% e
\textbf{e:} Histopathological analysis of resected tumor regions.
As it is difficult to provide an absolute ground truth for the resection region, where the entire surface geometry changes after the histopathological processing procedure, a reference post-resection H\&E slicing image is provided. 
This image shows that most of the cutting region covers the tumorous regions visible as discoloration of purple color (tumor). 
% f
\textbf{f:} Volumetric rendering for the regions of prediction (green) and actual cutting (blue).
% g 
\textbf{g:} Reference images of the post-resection mice tumors.
% h
\textbf{h:} ROI edge error distributions (highlighted step size = 1.86 mm). 
% i
\textbf{i:} Ratio distributions of the Intersection-over-union, undercutting, and overcutting ratios for the laser resection tasks.  
}
\label{fig_5_mice_exp}
\end{figure}

System evaluation was performed using the same error metrics described in Section.~\ref{system_sphere_region_checking}.
The results of murine tumor resection experiments are summarized in Fig.~\ref{fig_5_mice_exp}h-i. 
Since a complete histopathological analysis of the tumor region cannot be performed intraoperatively, due to workflow constraints and tissue deformation during processing, there was no absolute ground truth for the entire tumor region. 
Therefore, we evaluated the performance of the resection by comparing the predicted tumor region (from the TumorID classifier) and the resected region (actual laser ablation). 
This comparison serves as a proxy to assess the effectiveness with which the system could remove the tissue predicted to be tumorous. 
Quantitative metrics used to evaluate performance include boundary edge error, intersection-over-union (IoU), undercutting ratio, and overcutting ratio. 
The average boundary edge error is $0.94 \pm 0.62$ mm, and this is approximately $51\%$ of the laser's step size ($1.86$ mm), with higher variance attributed to the classification uncertainty near tumor margin.
Notably, even a single misclassified point can lead to substantial boundary offsets due to the coarse spatial resolution imposed by the step size and laser spot diameter. 
This can be improved by finer spatial sampling, albeit at the cost of increased acquisition time.

The system achieves an average IoU of $60\%$ between the predicted and resected regions.
The additional 40\% gap for the IoU towards the perfect alignment (i.e., 100\%) is due to the ambiguity of the classification around the boundary based on the current step size setting (i.e., 1.86 mm). 
In addition, a higher undercutting ratio (about $40\%$) is observed compared to overcutting (about $20\%$). 
These results suggest that the system tends to achieve a cautious and tissue-conscious strategy, a desirable trait in clinical practice where preserving healthy tissue is paramount, and residual tumor tissue can be removed in a follow-up pass.

Although intraoperative histopathology was not feasible, post-resection Hematoxylin and Eosin (H\&E) stained tissue sections were obtained for visual confirmation (Fig.~\ref{fig_5_mice_exp}e). 
These images demonstrate that most of the resected areas correspond to visibly tumorous regions, supporting the TumorID classification accuracy. 

In summary, the murine experiments validate the integrated pipeline for 3D tumor mapping and resection. 
The successful deployment of tumor resections in OS and STS shows potential generalizability of TumorMap to target other tumor types. 
This also indicates that TumorID can serve as a viable and noncontact proxy for histopathological diagnosis, enabling precise and automated resection of tumors without human intervention. 
We further establish the validity of TumorID in the next section (Section.~\ref{main_tumorid_histopahology_analysis}). 

\subsection{Histopathological analysis and verification of TumorID}
\label{main_tumorid_histopahology_analysis}
In earlier stages of our study (Section~\ref{section_mice_resection}), tumor labels were generated based on expert-delineated regions on the images, leveraging the known biomechanical differences between sarcoma tumors and healthy tissue in murine models.  
Sarcoma tumors tend to exhibit increased stiffness due to changes in cellular and extracellular matrix composition \cite{jain2014role}, which served as the primary cue for expert annotation.  
While this method facilitated rapid label assignment during live experiments, it lacked histopathological confirmation.
Histopathological analysis remains the clinical gold standard for tumor classification and diagnosis, offering definitive insights into tissue pathology based on cellular morphology and structure \cite{noorbakhsh2020deep}. 
However, integrating this analysis into the development of TumorID classifiers poses significant challenges due to data scarcity and complicated histopathological workflows. 
First, the time-consuming H\&E slicing procedures prevent the generation of sufficient data labels.  
Second, the complexity of tissue preparation, staining procedures, and manual interpretation hinders real-time compatibility and complicates data standardization for training machine learning models.
Therefore, we developed a novel experimental histopathological pipeline to obtain ground truth information for the verification of the proposed TumorID tumor classifier (\textbf{Movie~S6}).

For the histopathological analysis, we retrained the TumorID classifier with different dataset configurations.
We prepared a follow-up dataset where the training labels were derived from visual annotations (as before), but the testing labels were obtained directly through histopathological analysis of H\&E-stained tissue sections (Fig~\ref{fig_6_histopahotlogy_exp}a-d). 
This dataset formulation strategy combined visual labels for training and gold standard histopathology for testing, verified by a pathologist, enabling an unbiased evaluation of TumorID. 
By establishing point-level correspondences between spectral measurements and histologically verified labels, we assessed the extent to which TumorID predictions aligned with clinical diagnostic criteria.

\subsubsection{Training tumor classifier}
To train the tumor classification model, we adopted the dataset formulation method described earlier (Section.~\ref{section_mice_resection}) using expert-generated visual labels to annotate the tumor and healthy regions. 
This labeling method enabled a sufficiently large and diverse dataset to support effective model training. 
The same model architecture in Section~\ref{murine_tumor_classification} was applied.
For this verification study, we employed a modified data acquisition setup to enhance spectral signal quality. 
The TumorID sensor was configured at a 90-degree angle relative to the tissue surface to maximize signal strength, which was different from the 40-degree tilt used in previous resection experiments. 
This orthogonal configuration improved the laser-tissue interaction and maximized the fluorescence signal-to-noise ratio, which was critical in generating high-quality training data. 
Using this updated configuration, we applied the data collection workflow in Section \ref{murine_tumor_classification} for eight murine tumor samples that include five STS and three OS models. 

\subsubsection{Testing dataset from the histopathological analysis}
\label{Testing_dataset_for_histopathological_analysis}
\begin{figure}[H]
\centering
\includegraphics[width = 0.98\textwidth]{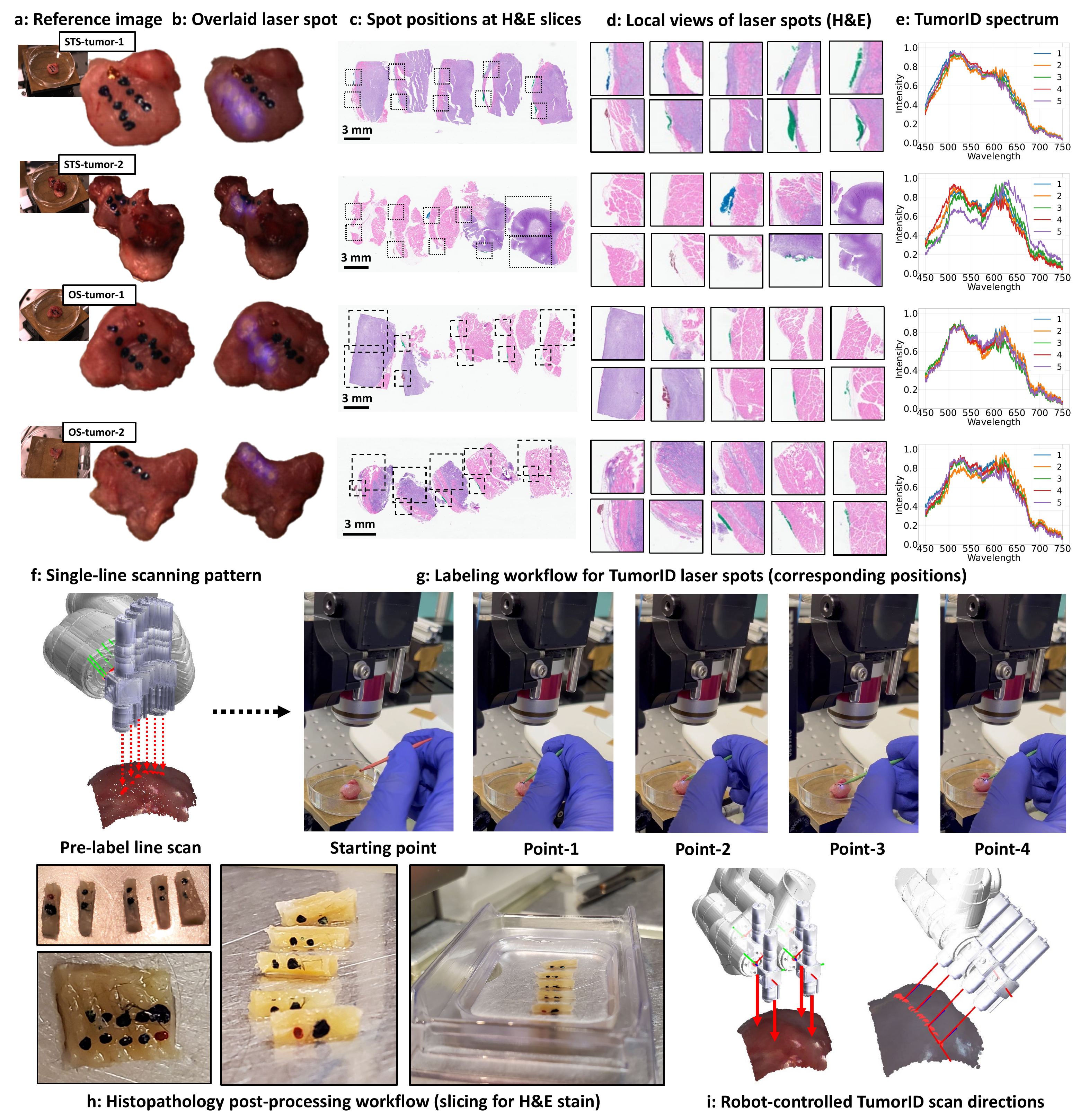}
\caption{
\textbf{Histopathology analysis for point-based fluorescence measurements.}
% a.
\textbf{a:} The reference images of STS- and OS- mice tumors.  
% b
\textbf{b:} The laser spots of the line scanning overlapped with the images correlated with the labeled points (purple color shows the laser spot).   
% c
\textbf{c:} The global view of the histopathology images for the labeled dots (highlighted regions denote the estimation of the spot positions). 
The bounding box does not represent the actual label orders and is used for local visualization only. 
% d
\textbf{d:} The local view of the tumorous and healthy spots, where each region is correlated with a label (i.e. tumor or healthy) for histopathological verification. 
\textbf{e:} The corresponding spectral data for the five overlaid laser spots in \textbf{(b)}.  
% b.
\textbf{f:} Robot-laser line scanning towards the 90-degree scanning. 
% c.
\textbf{g:} The workflow of labeling the laser spot associated with the TumorID sensor. 
The robot achieved a stop-and-scan strategy for data collection. 
The robot stops, and the TumorID emits a visible low-intensity laser mark to guide the manual annotation. 
\textbf{h:} The histopathological post-processing workflow for H\&E slicing and preparation (details in Method.~\ref{method_doe_histopathology}).
\textbf{i:} Geometry of the 40-degree and 90-degree scanning with various laser orientations.
}
\label{fig_6_histopahotlogy_exp}
\end{figure}

We established a reliable ground truth for evaluating the TumorID's classification accuracy. 
This was achieved by correlating the spectral data with the histopathological labels obtained using H\&E method.
To collect the data, we developed a novel stop-and-scan acquisition strategy that enabled point-wise mapping between in situ fluorescence spectra and histologically labeled regions, as illustrated in Fig.~\ref{fig_6_histopahotlogy_exp}b–e. 
The TumorID sensor performed a line-scanning over the tissue and sequentially paused at each point to collect spectral data, generating a visible low-intensity laser mark. 
These laser spots served as fiducial indicators for manual annotation and persisted through the entire histological processing pipeline.
Annotated fiducials generated a point-to-point matching between the fluorescence spectral data and the colorized label. 

To collect a high-quality dataset while minimizing cross-contamination between adjacent points, especially during manual marking with colored dyes, we adopted a two-pass scanning protocol (Fig.~\ref{fig_6_histopahotlogy_exp}). 
In the first pass, a fine-resolution scan was used to collect 10 data points along a line with a step size of 0.75 mm. 
A total of ten data points were collected for each line scan.
In the second pass, a larger 1.5 mm spacing was applied to ensure clear separation of laser spots and avoid overlapping dye artifacts, thereby improving post-staining visibility and label clarity. 
Each scan point was marked with cancer marker dye using a pointed tip and the annotations were used later for H\&E slicing. 
These point-by-point data, paired with the corresponding fluorescence measurements, formed a high-quality testing set for evaluating the classifier's accuracy under histologically verified conditions. 

\renewcommand{\arraystretch}{1.5}
\setlength{\arrayrulewidth}{0.5mm}
\begin{table}[h]
\centering
\caption{
\textbf{Mouse prediction results with histopathological labels as ground truth (with histopathology analysis)}. 
The blue labeled groups represent good model performances with statistics greater than $60\%$.
The train count indicates the number of points in the training dataset with labels given by experts based on visual information. 
Pathology labels indicate the number of points for the pathology-based testing dataset, where each point was labeled by H\&E staining information. 
TP (counts): true positive; TN: true negative; FP: false positive; and FN: false negative.
}
\label{ml_result_90_degree}
\resizebox{\columnwidth}{!}{%
\begin{tabular}{ccccccccccccc}
\hline
\hline
Combinations & Train (counts) & Pathology labels & TP & TN & FP & FN & Accuracy & Precision & Recall & F1-score & Specificity \\
\hline
\hline

OS-1 + OS-2 & 78 & 20 & 12 & 0 & 8 & 0 & 0.60 & 0.60 & 1.00 & 0.75 & 0.00 \\
OS-1 + OS-3 & 76 & 8 & 4 & 0 & 4 & 0 & 0.50 & 0.50 & 1.00 & 0.67 & 0.00 \\
\colorbox{blue!20}{OS-2 + OS-3} & 86 & 10 & 3 & 6 & 0 & 1 & 0.90 & 1.00 & 0.75 & 0.86 & 1.00 \\

\hline
\hline

STS-1 + STS-2 + STS-3 & 202 & 30 & 13 & 2 & 14 & 1 & 0.50 & 0.48 & 0.93 & 0.63 & 0.12 \\
STS-1 + STS-2 + STS-4 & 199 & 30 & 21 & 2 & 4 & 3 & 0.77 & 0.84 & 0.88 & 0.86 & 0.33 \\
STS-1 + STS-2 + STS-5 & 197 & 40 & 24 & 0 & 14 & 2 & 0.60 & 0.63 & 0.92 & 0.75 & 0.00 \\
\colorbox{blue!20}{STS-1 + STS-3 + STS-4} & 196 & 30 & 13 & 9 & 5 & 3 & 0.73 & 0.72 & 0.81 & 0.76 & 0.64 \\
STS-1 + STS-3 + STS-5 & 194 & 40 & 11 & 10 & 12 & 7 & 0.53 & 0.48 & 0.61 & 0.54 & 0.45 \\
STS-1 + STS-4 + STS-5 & 191 & 40 & 27 & 3 & 9 & 1 & 0.75 & 0.75 & 0.96 & 0.84 & 0.25 \\
STS-2 + STS-3 + STS-4 & 186 & 26 & 6 & 10 & 4 & 6 & 0.62 & 0.60 & 0.50 & 0.55 & 0.71 \\
STS-2 + STS-3 + STS-5 & 184 & 36 & 13 & 6 & 16 & 1 & 0.53 & 0.45 & 0.93 & 0.60 & 0.27 \\
\colorbox{blue!20}{STS-2 + STS-4 + STS-5} & 181 & 36 & 19 & 9 & 3 & 5 & 0.78 & 0.86 & 0.79 & 0.83 & 0.75 \\
STS-3 + STS-4 + STS-5 & 178 & 36 & 13 & 11 & 9 & 3 & 0.67 & 0.59 & 0.81 & 0.68 & 0.55 \\

\hline
\hline

STS-1 + STS-2 + STS-3 + STS-4 & 228 & 10 & 4 & 2 & 2 & 2 & 0.60 & 0.67 & 0.67 & 0.67 & 0.50 \\
STS-1 + STS-2 + STS-3 + STS-5 & 226 & 20 & 8 & 0 & 12 & 0 & 0.40 & 0.40 & 1.00 & 0.57 & 0.00 \\
STS-1 + STS-2 + STS-4 + STS-5 & 223 & 20 & 18 & 0 & 2 & 0 & 0.90 & 0.90 & 1.00 & 0.95 & 0.00 \\
STS-1 + STS-3 + STS-4 + STS-5 & 220 & 20 & 9 & 3 & 7 & 1 & 0.60 & 0.56 & 0.90 & 0.69 & 0.30 \\
\colorbox{blue!20}{STS-2 + STS-3 + STS-4 + STS-5} & 210 & 16 & 6 & 8 & 2 & 0 & 0.88 & 0.75 & 1.00 & 0.86 & 0.80 \\

\hline
\end{tabular}%
}
\end{table}

\subsubsection{Histopathological verification of the tumor classification model}
% (Section~\ref{Testing_dataset_for_histopathological_analysis})
Using the labels obtained through histopathological analysis, we formulated a specialized test dataset to perform an unbiased evaluation of the TumorID tumor classifier. 
Due to the limited number of unique subjects, where each mouse ID represented an independent biological sample rather than a single data point, we used a cross-validation-based approach to generate robust training and testing splits. 
For STS tumors, a 3:2 sample-wise split was used to form 10 combinations; for OS tumors, a 2:1 split was used to form 3 combinations. 
Notably, the histopathology-labeled testing data were kept entirely separate from the training data to eliminate any potential data leakage and to ensure generalization beyond the training cohort. 
Each trained ML model was evaluated against its respective histopathology-labeled testing set. 

The results in Table~\ref{ml_result_90_degree} reveal considerable variation in performance across different sample combinations. 
Although some configurations yielded lower scores in standard metrics (accuracy, precision, recall, F1 score, specificity), several combinations demonstrate promising results, with all metrics exceeding 0.60, highlighting the feasibility of fluorescence-based classification in well-represented tumor subtypes. 
The observed discrepancies can be attributed to the complexity of tumor scanning conditions, such as bloody surfaces, and the variety of STS and OS tumors in generating subsets of tumor types that can produce different fluorescence effects. 
Notably, these findings are consistent with the results of the cross-validation experiments in Table~\ref{ml_result_40_degree}. 
The results also indicate that if mouse selection covers a wide variety of tumor subtypes, the TumorID sensor can serve as a proxy for the gold standard histopathological method for the intraoperative tumor diagnosis, while providing noncontact tissue scanning during robotic surgery. 

In summary, we conducted a system-level verification study to demonstrate the feasibility of using TumorID for point-based tumor classification, which was validated against histopathology-derived ground truth. 
The experimental result highlights the effectiveness of robot-controlled data acquisition for rapid and consistent training of machine learning models in a surgical context. 

\section{Discussion}
Automated tumor resection in oncology surgery is an emerging field where the convergence of robotics, artificial intelligence (AI), and advanced sensing is driving significant advances, although current applications are mainly teleoperated or supervised rather than fully autonomous.
While recent breakthroughs in large-scale deep learning models are accelerating research to build intelligent algorithms toward fully autonomous surgery \cite{long2025surgical, kim2025srt}, major challenges still exist in system integration with three key problems of high-fidelity tumor reconstruction, pathological region mapping, and reliable resection approaches. 
To overcome these challenges, we developed TumorMap, a fully-automated robotic laser platform that integrates multimodal sensing information, real-time tissue diagnosis, and noncontact laser resection for automated tumor resection.  
This system represents the first demonstration of solid tumor identification and removal executed through a fully-automated workflow without human intervention.

In contrast to conventional surgical robotic systems, which rely primarily on contact-based tools such as scalpels and forceps \cite{abdelaziz2024fiberbots, kuntz2023autonomous, saeidi2022autonomous, fang2021soft, shademan2016supervised, price2023using}, the three-laser mechanism of TumorMap enables noncontact diagnosis and resection with enhanced precision and geometric preservation. 
This architecture also supports a seamless transition between automated and human-in-the-loop workflows - critical for practical deployment in clinical settings.
Our platform has broad applicability in biomedical research and clinical translation, particularly for 3D mapping of superficial tumors. By integrating anatomical, functional, and pathological information into a unified surgical framework, TumorMap offers a powerful tool for surgical navigation, intraoperative diagnosis, and image-guided tissue resection.

Through comprehensive experimental validation, including structured phantom studies, \textit{ex vivo} tissue tests (chicken and porcine), and murine models with genetically engineered sarcomas, TumorMap has demonstrated excellent potential for 3D tumor mapping and automated resection with submillimeter resection accuracy (average error < 1.0 mm). 
The residual system error primarily came from calibration mismatches and integration artifacts in the diagnostic, perception, and cutting subsystems. 
Errors can be further mitigated through the design of a compact and unified sensor-laser platform \cite{prakash2025portable}.
TumorMap was further validated on murine models of soft tissue sarcoma (STS) and osteosarcoma (OS) to assess its adaptability across diverse tumor pathologies and morphologies. 
These genetically engineered sarcoma models were chosen due to the clinical complexity of sarcoma resection, which is challenged by a broad spectrum of histological subtypes and anatomical locations, including the extremities, retroperitoneum, pelvis, spine, head, and neck.
The successful achievements of tumor resections in mice of different tumor types show that the use of a dexterous robotic platform, laser-induced fluorescence sensing (point-based), and deep learning models allows complete surface tumor mapping and boundary detection. 

A key observation, however, was the difficulty in training robust tumor classifiers due to significant intraclass heterogeneity. 
Even within the same tumor type, variations in grade, vascularity, inflammatory state, and surface characteristics (e.g., presence of blood or necrosis) introduce substantial variability in spectral signatures. 
These factors underscore the need for a broader and more representative training dataset to ensure the generalizability of classification models across the full spectrum of biological variability.
In addition, we developed a novel histopathological workflow to enable accurate and validated identification of tissue characteristics in H\&E stained images. 
This approach establishes a direct correspondence between fluorescence measurements and histological ground truth by allowing operators to precisely label tissue regions scanned by the robot-controlled TumorID sensor.
Our histopathological analysis has successfully demonstrated the feasibility of using a robotic laser-induced endogenous fluorescence sensor for tumor diagnosis, with potential applications in surgical automation, visualization, and intraoperative care.

In summary, TumorMap offers an extensible software and algorithmic framework that bridges machine learning, robotics, and surgical planning, significantly advancing the integration of complementary sensors into medical robotic systems with minimal hardware modifications. 
This modular design enables rapid adaptation to evolving surgical workflows and new sensing modalities. 
Future work can focus on developing a more robust and adaptive laser orientation planner capable of dynamically adjusting the scanning angle based on real-time surface geometry.
Furthermore, incorporating closed-loop feedback control informed by intraoperative sensing data can further enhance the precision and safety of tumor resections.
These advancements will contribute to the continued refinement of TumorMap and its transition to intelligent and autonomous surgical platforms for clinical use.

\section{Materials and Methods}
\subsection{TumorMap framework}
\label{method_tumormap_framework}
The TumorMap framework combines a perception system, a laser-induced tumor diagnosis system, and a laser scalpel in a unified manner to achieve 3D tumor boundary reconstruction. 
The coordination of these components required an accurate system calibration to align their spatial transformation in a global coordinate frame. 
Once calibrated, the proposed robot kinematics and trajectory planning methods were used to generate laser trajectories to visit tumor targets for subsequent resection. 
This section mainly discusses the framework for robot calibration, kinematics, planning, and technical details of experiments on phantom, \textit{ex vivo} tissue of chicken and porcine, and the murine tumor models.

\subsubsection{Tumor diagnosis system} 
\label{method_tumormap_framework_tissue_diagnosis_system}
The tumor diagnosis system consists of a laser-induced endogenous fluorescence sensor integrated into the end-effector (EE) of a 6-DOF robot arm (UR5e, Universal Robots, Denmark) to achieve point-based tissue scanning (Fig.~\ref{fig_1_system_main}b).
This system is controlled to perform predefined scanning patterns above the superficial tissue surface.
The tissue diagnosis system is improved from our previous work \cite{tucker2022creation, tucker2021creation, zachem2025intraoperative} (referred to as \textbf{"TumorID"} sensor, i.e., tumor identification for single spot measurement). 
TumorID is modified with an objective lens of a 0.2 NA, 2X super-apochromatic microscope objective (Model: TL2X-SAP Thorlabs, NJ, United States) to achieve a working distance of 56.3 mm, integrating a compact spectrometer with a spectral bandwidth of 350 to 700 nm (Model: CCS100, Thorlabs, Newton, NJ), and a compact laser diode driver for the 405 nm wavelength (Model: CLD1010LP, Thorlabs, NJ, United States), to produce an effective endogenous fluorescence signal at a longer working distance.
During the experiments, the power of the laser was 220 mW and the integration time of the spectrometer was 1.0 seconds.

Endogenous fluorescence leverages the Warburg effect, where tumor cells have altered metabolic pathways compared to normal cells, to differentiate healthy and neoplastic tissue \cite{liberti2016warburg}. 
The Warburg effect leads to changes in endogenous biomarkers (i.e., NADH and FAD) within cancer cells, and these changes are detected by an appropriate optical setup.
Thus, the difference in signal intensity can be used to detect tumor cells.
For each point-based target irradiated by the laser, a fluorescence spectrum signal (1D signal) is collected and passed through a multilayer perceptron (MLP) network to differentiate tumorous or healthy samples.

\subsubsection{Tumor perception system}
\label{method_tumormap_framework_tissue_perception_system}
The tumor perception system is tasked with generating a colorized 3D tissue surface (Section \ref{method_tumor_map_oct_and_color}). 
It consists of a dual-camera system of two $1280 \times 720$ resolution USB 3.0 industrial cameras with a side-by-side configuration (Model: DFK 33UP1300, The Imaging Source, NC, United States) with 8-mm lenses (Model: TCL 0814, The Imaging Source, NC, United States), as well as an $1310~nm$ optical coherence tomography (OCT) scanner (Lumedica, Inc., OQ-StrataScope, NC, United States) in tabletop configuration.  
For 3D colorized surface reconstruction, the OCT sensor first captures a high-fidelity image volume up to 7.5 mm below the tissue surface.
The colorized information from the dual camera system is initially registered on the 3D OCT surface through the system calibration (Method.~\ref{method_calibration_oct_to_camera}). 
This formulates a 3D colorized surface to initialize the subsequent robot planning.
Combined with data from the TumorID sensor, a tumor map is created to provide multimodal information from color, pathology, and geometry to assist in surgical planning and navigation (workflow in Section \ref{workflow_of_tumor_mapping}). 

\subsubsection{Tumor resection system}
\label{method_tumormap_framework_tissue_resection_system}
The tumor resection system comprises a cutting laser scalpel attached to the robot-EE along with the TumorID (Fig.~\ref{fig_1_system_main}b).  
This design enables surface resection with predefined trajectories while maintaining adequate distance and orientation to the tissue surface.  
The fiber-coupled laser (400 $\mu m$ fiber diameter, 1940 nm wavelength, Thulium fiber laser, IPG Photonics, MA, USA) is attached to the robot-EE to achieve free-space movement.
%  simplified and 
In geometric modeling and ray tracing \cite{pharr2023physically}, a laser beam was modeled as a combination of a laser origin point and a laser vector, allowing for easy implementation of system calibration, kinematics, and planning modeling (Method.~\ref{method_calibration_oct_to_camera}, \ref{method_system_model_ik_and_traj_plan}).
% energy
Laser settings of 0.5 joules (energy), 3 Hz pulse frequency, an approximately 6.5 mm/s fiber distal end speed (controlled by the robot arm) and a working distance of about 3 mm (depending on laser settings and the tissue shapes) were used for tissue resection to ensure efficient removal, which was empirically obtained and adjusted to minimize collateral damage. 

\subsection{Tumor mapping workflow}
\label{workflow_of_tumor_mapping}

\subsubsection{3D OCT surface and color map} 
\label{method_tumor_map_oct_and_color}
The OCT sensor first collects cross-sectional B-scan images of the tissue based on region of about \textcolor{black}{$12.6 \times 12.8 \times 7.5~mm^3$} through an one-time scan of approximately 13 seconds.
For each C-scan, 128 B-scans are obtained. 
The OCT surface map is reconstructed from sequential B-scan images based on an axial pixel scaling of approximately 14.6 $\mu m$ (after calibrating the pixel pitch of the B-scan image with a shape of $512 \times 512$). 
From each B-scan, the surface is segmented by identifying the maximum pixel intensity along the A-scans based on the assumption that the light beam is most strongly reflected or backscattered at the air-tissue interface.
The stitching of the segmented images formulates a 3D surface, and this allows for high-fidelity screening of microscale tumor structures. 
For the color map, the left camera serves as the reference one to provide the colorized information (only one camera is needed to provide color information). 
The right camera is used with the left camera for an accurate estimation of the position of the laser point (see Method.~\ref{method_tumor_map_diagnosis_for_map_formulation}). 
The pixel information is registered with the OCT through the system calibration method (Method.~\ref{method_calibration_oct_to_camera}). 
For surface-only analysis (instead of volume), we assume that most surface points can be mapped to the image frame and have unique correspondences (with the nearest closest point) associated with the pixels.
The pixel color corresponds to the closest 2D projected point associated with the 3D coordinate in the OCT frame. 
The color information is aligned with the surface to formulate a 3D colorized map that provides surgeons with a representation similar to the real world scene.

\subsubsection{3D tumor map}
\label{method_tumor_map_diagnosis_for_map_formulation}
As the colorized map only provides color-related details, and no clear pathological information to localize the tumor region, building the 3D  tumor map aims to classify the points labeled as ``tumors" and stitch these tags to reconstruct the boundary.
To formulate the tumor map, the robot-mounted TumorID first performs a raster scan at a working distance of \textcolor{black}{$56.3~mm$} and an orientation of \textcolor{black}{40 degrees off-perpendicular to the Z-axis of the OCT frame} to minimize potential collisions to the tabletop sensor. 
The TumorID scanning region is denoted by a predefined meshgrid above the tissue surface, based on a configuration of \textcolor{black}{$13.0 \times 13.0~mm^2$ and a step size of $1.86~mm$}, leading to a total of 64 data points.
This scanning region was designed to be greater than the lateral OCT field-of-view (FOV). 
Each point-based measurement is processed by the tumor classifier to assign the label for each 3D tag (healthy versus tumor).
Through system calibration, the TumorID laser position is matched to a 3D surface point (from the OCT reconstruction). 
This formulates two 3D points through projections of left and right cameras. 
As a single laser ray can be mapped to multiple points towards the same laser orientation (Fig.~\ref{fig_7_system_calib}f), the ray tracing algorithm is implemented to determine the intersection point (of the surface point cloud) that lies in the laser beam, which formulates the third representation of the laser spot \cite{moller2005fast}.  
An average filter is implemented to calculate the average value of these three surface points to estimate the accurate laser position.
This increases the accuracy of 3D laser spot estimations and minimizes the uncertainties of the sensor models due to variations of image features and sensor noises.
The concatenation of these discrete tags formulates a 3D tumor map where each position is assigned with colorized, pathological, and geometric information. 
Here, it is assumed that the tumor boundary is convex and all points within the boundary are considered pathological and targeted for resection, which applies to both murine tumors and \textit{ex vivo} tissue models. 

\begin{figure}[H]
\centering
\includegraphics[width = 0.95\textwidth]{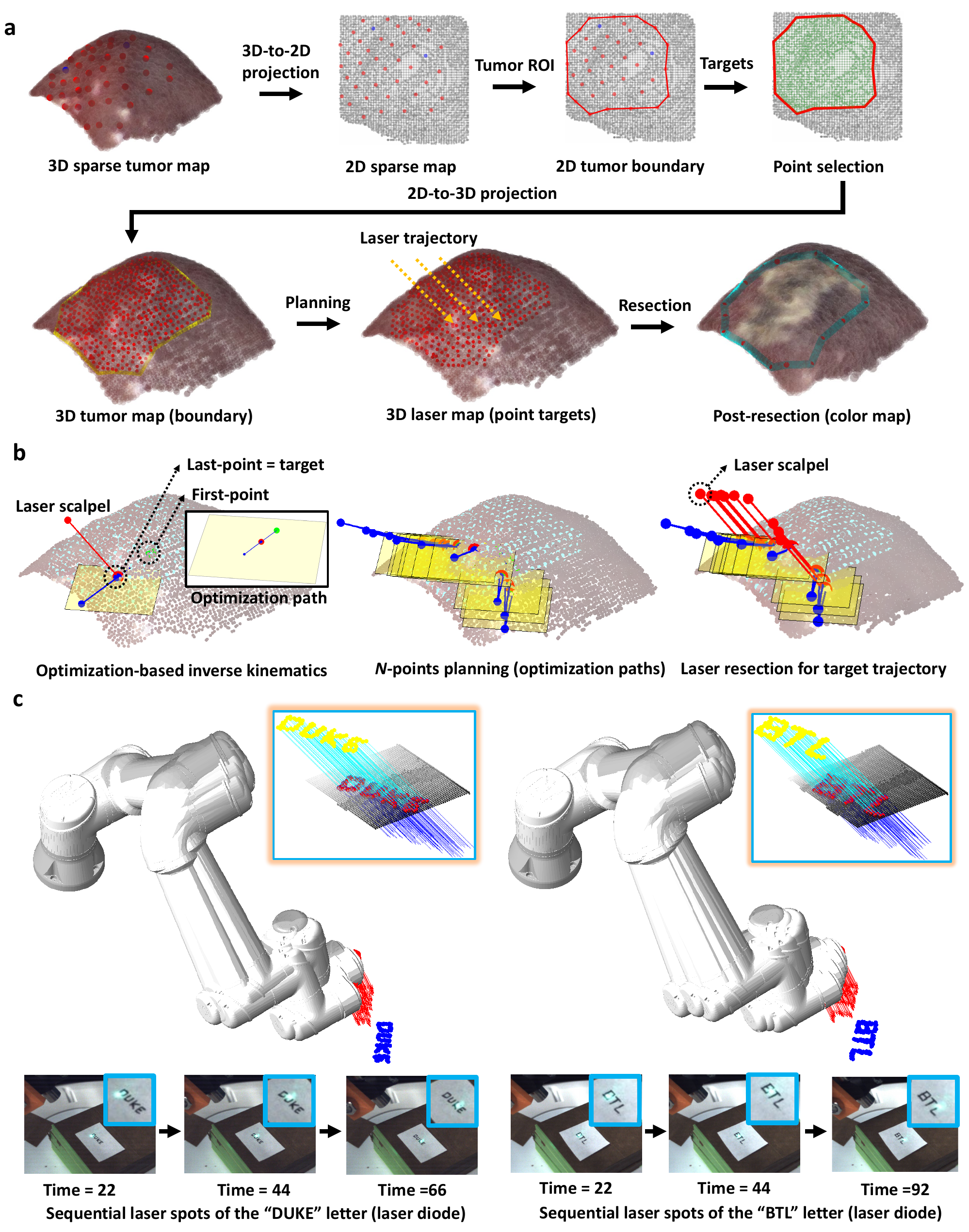}
\caption{
\textbf{Overview of the system calibration, kinematics modeling, and trajectory generation.} % 
% a 
\textbf{a:} The workflow for 3D tumor map generation with the convex-hull based boundary formulation algorithm \cite{yap2013quantitative}. 
The 3D sparse tumor map is first projected to a 2D plane. 
The 2D tumor boundary is formulated using the convex hull algorithm (Method \ref{method_model_tumor_roi_geometry}) and reprojected back to the 3D tumor boundary.
The points within the boundary are selected to formulate the cutting trajectory. 
% b
\textbf{b:} Separated steps of the inverse kinematics solver.
A target point is first selected, and a reference plane is assigned to this point with a surface normal vector and center. 
An optimization-based inverse kinematics solver is applied to solve this problem and generate a sequence of laser waypoints. 
The optimization solver finally produces correct solutions from the starting position (green) to the optimal one (red). 
The blue point depicts the intermediate update.  
All the points are stacked in a output vector from the optimization solver. 
The resultant outputs are the laser positions with the same laser vector targeting the target points.
Although there are necessary components present to perform close-loop resection in future, TumorMap works in an open-loop fashion where each waypoint of the trajectory is pre-computed. 
% c.
\textbf{c:} A demonstration example of the TumorMap system in targeting arbitrary shapes in the OCT scanning region. 
We selected the letters of ``BTL" and ``Duke" to formulate the 3D targets. 
The targets were scanned by the OCT sensor, and the optimization solver was used to generate the optimal laser resection trajectory. 
The laser diode was controlled to visit these points following the trajectory.
}
\label{fig_8_system_model}
\end{figure}

\label{method_model_tumor_roi_geometry}
The classification tags (on the surface pointcloud) are used to formulate the tumor boundary (workflow in Fig.~\ref{fig_8_system_model}a). 
These targets are first projected to a 2D reference plane parallel to the X- and Y- axes in the OCT frame, generating the 2D projected coordinates, which represent the geometry of the tumor boundary. 
A polygon is created from these tumor points using the MATLAB \texttt{boundary} function with a shrinking factor of 0.5 (using the convex hull algorithm \cite{yap2013quantitative}).  
The points within the polygon are selected to formulate the final tumor region using the same index as the 2D boundary points (Fig.~\ref{fig_8_system_model}a). 

\subsubsection{3D laser cutting map}
\label{method_tumor_map_laser_resection}
Given a 3D tumor region, the points within the segmented boundary are selected to formulate a 3D resection map.  
The selected points serve as targets for the calculation of the robot trajectory using the optimization-based inverse kinematics solver (Fig.~\ref{fig_7_system_calib} f to h). 
Each waypoint of the trajectory is assigned an optimal laser position and orientation towards the surface target (Method.~\ref{method_system_model_ik_and_traj_plan}).
Given the calculated trajectory, the fiber-coupled laser is controlled to visit each point sequentially for tumor removal. 

\subsection{System calibration}
\label{method_system_calibration}
System calibration is a necessary step in estimating real-world geometry among sensors, tumor targets, robots, and laser modules. 
This ensures that the robot's actions informed by sensor data are accurately translated to the tool's movements. 
Defining the OCT sensor as the world frame, the system calibration is decomposed as \textbf{1) the OCT-to-camera and 2) OCT-to-laser calibration}.

\subsubsection{OCT-to-camera calibration}
\label{method_calibration_oct_to_camera}
The goal of OCT-to-camera calibration is to determine the spatial relationship between the camera system and the OCT sensor. 
Given this information, a fluorescence signal (with the laser spot) is matched with a 3D point in the OCT frame. 
The OCT-to-camera calibration is decomposed into an intrinsic and an extrinsic step. 
The extrinsic calibration aims to estimate the transformation between the left and right cameras to the OCT sensor.
The camera intrinsic calibration parameters are required for the extrinsic calibration (OCT-to-camera) to project the 3D points onto a 2D image plane (Fig.~\ref{fig_7_system_calib}a). 

\noindent
\textbf{Camera intrinsics calibration:}
Standardized camera intrinsics calibration was performed for the left and right cameras using the MATLAB camera calibration toolbox (Fig.~\ref{fig_7_system_calib}a). 
We used 20 images from the chessboard targets (in various poses) to estimate the intrinsic camera parameters. 
The average pixel errors were reported to be 0.09 for the left camera and 0.07 for the right camera. 
\noindent
\textbf{Camera extrinsics calibration:}
Given the intrinsic calibration information, the MATLAB function \texttt{estimateExtrinsics} was used to estimate the OCT-to-camera matrix. 
The MATLAB function \texttt{world2img} was applied to project the OCT coordinates onto the pixel frame of the left camera, which was defined as the reference coordinate system. 
For extrinsic calibration, a planar calibration board with fiducial markers was positioned in different poses, encompassing a range of orientations and translations relative to the camera. 
The calibration board for each pose was scanned with the OCT sensor and the 3D centers of the fiducial markers were segmented from the OCT-reconstructed images. 
The corresponding 2D pixel coordinates of the cameras were labeled in the image frame.
This created a 2D-to-3D mapping, where the 2D and 3D coordinates were used as inputs for the \texttt{estimateExtrinsics} function. 
For the implementation, six calibration poses were used to cover the orientations and translations (Fig.~\ref{fig_7_system_calib}a).  
Extrinsic calibration reprojection errors were reported as $0.24 \pm 0.15$ mm (left camera) and $0.28 \pm 0.19$ mm (right camera). 

\subsubsection{OCT-to-laser calibration} 
\label{method_calibration_oct_to_laser}
The OCT-to-laser calibration aims to estimate the spatial relation between the robot-controlled lasers and the OCT sensor. 
This is a necessary step for solving the trajectory planning problem to trace targets.  
We first define a reference frame relative to the world frame (i.e., OCT frame) above the tissue surface, by which the robot-laser trajectories are calculated (Fig.~\ref{fig_7_system_calib}e). 
The reference frame is used to build the constraint so that the laser can always move with a fixed working distance to the tissue target. 
The laser is controlled with a starting point from the origin and moves only in the frame defined by the X-axis and the Y-axis (Fig.~\ref{fig_7_system_calib}b-d). 
The incident laser intersects the object's surface and the intersection point (prediction) is associated with the given robot pose.
In this model, the X- and Y- axes and the laser incidence orientations are unknown variables that needed to be calibrated. 

\noindent
\textbf{Step-1: OCT-to-laser axis calibration.} 
The determination of the X-axis and Y-axis in the reference plane requires a collection of two fiducial markers (i.e., starting and ending points) in both directions. 
To solve this problem, we attached a fiber tip to the robot-EE where the end position is detected by the OCT sensor to represent a 3D point (Fig.~\ref{fig_7_system_calib}c).  
Two fiducial markers were collected for each direction to estimate the orientation of the axes.
Specifically, the robot end effector moved only in the reference plane to maintain the specialized orientation and object distance.
The reference frame is defined with an origin of $\textbf{p}_{s}^{org} \in \mathbb{R}^3$, an X-axis vector $\mathbf{v}_x^{oct} \in \mathbb{R}^3$, and a Y-axis vector $\mathbf{v}_y^{oct} \in \mathbb{R}^3$.
A waypoint point $\mathbf{p}_{w} \in \mathbb{R}^3$ in the frame is defined as (Fig.~\ref{fig_7_system_calib}e): 
\begin{equation}
    \begin{aligned}
        \textbf{p}_{w} & = \textbf{p}_{s}^{org} + (\alpha_x + \beta_x)  \mathbf{v}_x^{oct} + (\alpha_y + \beta_y)  \mathbf{v}_y^{oct}
    \end{aligned}
    \label{calib_1}
\end{equation}
Where $\textbf{p}_{s}^{org} = [0, 0, z_{obj}]$ ($z_{obj}$ is the fixed TumorID working distance to the target) is the origin of the reference frame and can be determined by calibration of the system. 
$\{ \alpha_x, \alpha_y \} \in \mathbb{R}^2$ are the fixed parameters that represent the origin. 
$\{ \beta_x, \beta_y \} \in \mathbb{R}^2$ is a pair of variables in the two directions to determine the waypoint coordinate. 
This formulates a full representation of a forward kinematics model that maps a robot configuration to a laser pose. 

\noindent
\textbf{Step-2: OCT-to-laser position and orientation calibration.}
The second step aims to estimate the laser orientation $\mathbf{v}_{w}$ from the reference frame and the parameters of $\alpha_x$ and $\alpha_y$. 
As a single laser ray can be matched to multiple points toward the same orientation (Fig.~\ref{fig_7_system_calib}f), a virtual plane is defined at the target point with its normal surface vector and the center $\mathbf{p}_{n} \in \mathbb{R}^3$ and $\mathbf{v}_{n} \in \mathbb{R}^3 $. 
This formulates a local geometry to describe a unique interaction point.  
Thus, the forward propagation model of a laser spot $\mathbf{p}_{i} \in \mathbb{R}^3$ is described as (at the intersection surface) \cite{li1995laser}: 
\begin{equation}
    \begin{aligned}
        \mathbf{p}_{i} & = \mathbf{p}_{w} - \frac{ \mathbf{v}_n \cdot (\mathbf{p}_{w} - \mathbf{p}_n) }{ \mathbf{v}_n \cdot \textbf{v}_w } \mathbf{v}_w
    \end{aligned}
    \label{point_ik_eqn}
\end{equation}
Where $\textbf{v}_w \in \mathbb{R}^3$ is the laser incidence vector.  
The $\{ \mathbf{p_w}, \mathbf{v_w} \}$ formulates a unique laser configuration to create an intersection point. 
The $\{ \mathbf{p_n}, \mathbf{v_n} \}$ denote the center and vector of a virtual plane defined around the target point to estimate the laser configuration. 
Equation~\ref{point_ik_eqn} can be plugged into Equation~\ref{calib_1} to describe the intersection point with a mapping function $g(\cdot)$: 
\begin{equation}
    \begin{aligned}
        \mathbf{p}_{i} & = g( ~ \underbrace{\textbf{v}_w , \alpha_x, \alpha_y }_{\text{OCT-to-laser calibration}}; ~ \underbrace{\beta_x, \beta_y}_{\text{Waypoint coordinate}}; ~ \underbrace{ \textbf{p}_n, \textbf{v}_n }_{\text{Target point geometry} } ) = \textbf{p}_{s}^{org} + (\alpha_x + \beta_x)  \mathbf{v}_x^{oct} + (\alpha_y + \beta_y)  \mathbf{v}_y^{oct}  - \frac{ \mathbf{v}_n \cdot (\mathbf{p}_s - \mathbf{p}_n) }{ \mathbf{v}_n \cdot \textbf{v}_w } \mathbf{v}_w
    \end{aligned}
    \label{fk_1}
\end{equation}
Where $\textbf{p}_{s}^{org}, \mathbf{v}_x^{oct}, \mathbf{v}_y^{oct}$ are initially given from the calibration of the OCT-to-laser axis ($\textbf{p}_{s}^{org}$ is an arbitrary variable that can be pre-defined).
To model calibration as an optimization problem, we first place a planar calibration board on the surgical site by which a laser can generate a unique spot on the surface.
This formulates geometric constraints for the optimization problem to calculate an optimally unique laser incidence orientation. 
To estimate the optimal laser incidence vector $\textbf{v}_w^*$ and the system parameters $\{ \alpha_x, \alpha_y \}$, an optimization problem is modeled as follows: 
% F_{calib} =
\begin{equation}
    \begin{aligned}
        \min_{\textbf{v}_w^*; ~ \alpha_x^*, ~ \alpha_y^*} \quad  \sum_j ~ || ~ g_j(\textbf{v}_w; \alpha_x, \alpha_y) - \textbf{p}_{j}^* ~ ||_2^2 
    \end{aligned}
    \label{eq_calib_main}
\end{equation}
Where $\textbf{p}_{j}^*$ is the $j$-th laser spot (visible spot by laser diode or ablation spot by fiber-coupled laser) created on the calibration board. 
The center of the laser spot is reconstructed from the OCT scan.
The group of parameters $\{ \beta x, \beta y ; ~ \textbf{p}_n , \textbf{v}_n \}_j$ is initially predefined and assigned for the $j$-th fiducial target on the calibration board. 
For the implementations, the calibration boards were placed at four different heights. 
The laser was controlled to create four ablation craters with centers estimated from the OCT sensor. 
The laser configuration, robot configuration, and 3D fiducial markers were matched and used as inputs for the optimization solver in Equation.~\ref{eq_calib_main}. 
The MATLAB \texttt{trust-region-reflective} solver was used to calculate the optimal laser configurations. 
The system calibration workflow is depicted in Fig.~\ref{fig_7_system_calib}a to d. 

\subsection{System modeling}
\label{method_system_model_all}
This section describes the robot-laser kinematics model and the laser trajectory planning.
The forward kinematics (FK) and inverse kinematics (IK) of the models are defined differently compared to conventional robot modeling approaches.  
% (e.g., a tumor point)
FK maps a robot configuration to a laser vector to determine the unique intersection point at the target surface, whereas, given a fixed target point, IK aims to estimate the optimal laser configuration to move the beam towards the target. 
We developed both the optimization-based IK solver for single-point estimation and a trajectory planner for multiple points (on the surface).
Once the laser configuration was solved with the IK solver, the movement of the robot-EE was determined using the Klampt robot simulator and solver \cite{hauser2016robust}. 

\subsubsection{Optimization-based inverse kinematics solver and laser trajectory formulation}
\label{method_system_model_ik_and_traj_plan}
\begin{figure}[H]
\centering
\includegraphics[width = 0.98\textwidth]{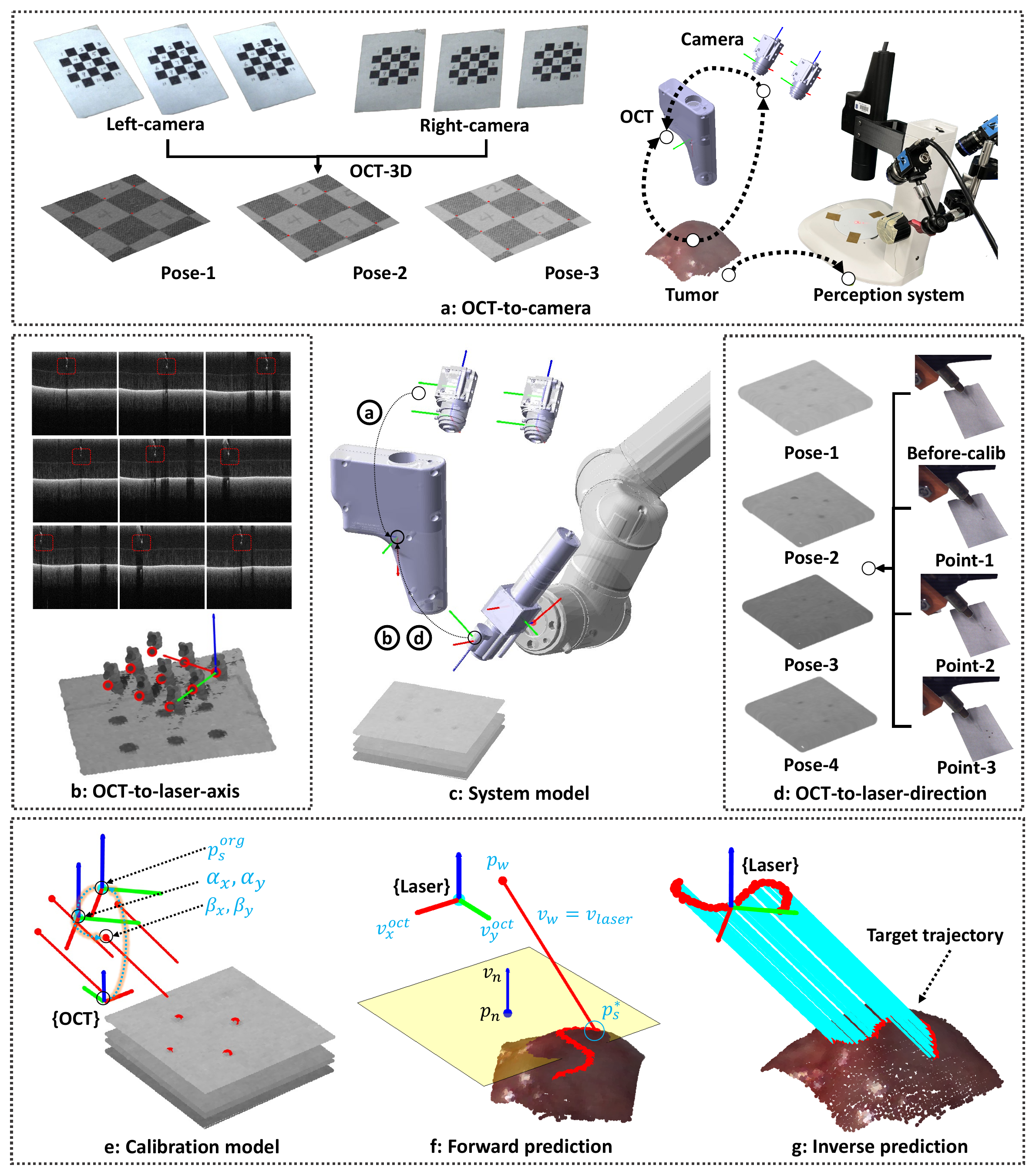}
\caption{
\textbf{Overview of the system calibration and modeling}
% a.
\textbf{a:} OCT-to-camera extrinsic calibration with the checkerboard pattern.
The 2D and 3D fiducial markers are segmented from the camera and the OCT sensor. 
With system calibration, the information (e.g., the fluorescence data from TumorID) can be mapped to the camera and the OCT. 
\textbf{b:} OCT-to-laser axis calibration (in the laser plane) with the fiber-shaped calibration patterns. 
The center of the fiber tooltip is moved to different positions following a $3 \times 3$ meshgrid pattern (toward the two axes). 
The tip center is segmented from the B-scan image and converted to 3D coordinates in the OCT frame through stitching of OCT image sequences. 
\textbf{c:} System calibration model with robot-guided TumorID and the 3D perception system. 
% d
\textbf{d:} OCT-to-laser orientation calibration. 
Four fiducial markers were generated with realistic laser resection on the planar surface that can be scanned by the OCT sensor. 
The centers of the markers serve as inputs for the optimization problem to estimate the optimal laser incidence orientation. 
The single collection covers four different marker positions to generate the laser ablation patterns, which are detected by the OCT sensor.
The center of the resection crater is used for system calibration by putting it into the optimization solver.
The optimal laser orientation vector is calculated, and a similar method is applied for the fiber-coupled laser and the TumorID laser.
% e
\textbf{e:} Calibration model of the laser frames. 
Each component is assigned a coordinate frame with the origin of the OCT sensor defined as the world frame.
The calibration and kinematics methods are developed based on the same world frame. 
% f
\textbf{f:} Reference plane to formulate the laser kinematics model. 
The center and a surface normal vector of the virtual plane are formulated. 
The $p_n$ is defined as $[0, 0, z_{p^*}]$ where $z_{p^*}$ is the z-axis coordinate of target point $p^*$.
This allows the reference surface to intersect with the target point to formulate the kinematics model. 
The surface normal is always defined as $[0, 0, 1]^T$. 
% g
\textbf{g:} The resection laser trajectory defined in the laser frame, which is denoted as a sequence of laser waypoints where the laser beam can be matched with each target on the surface. 
}
\label{fig_7_system_calib}
\end{figure}

For each tumor target point, a virtual plane with a center of $\textbf{p}_n$ and a surface vector of $\textbf{v}_n$ is assigned (Fig.~\ref{fig_7_system_calib}f).  
% the local geometry around the
% 
The $\textbf{p}_n$ is defined as the same as the target point, and $\textbf{v}_n$ is defined as a fixed vector as $\mathbf{v_n} = [0, ~0, ~1]^T$. 
The optimization-based IK is defined as: 
\begin{equation}
    \begin{aligned}
         \min_{ \beta_x, ~ \beta_y  } \quad || ~  g( \beta_x, \beta_y;  \mathbf{v_w^*}) - \textbf{p}^* ~ ||_2^2 
    \end{aligned}
    \label{laser_opt_obj_problem}
\end{equation}  
Where $g(\cdot)$ is the prediction function from Equation.~\ref{fk_1}.
The $\textbf{p}^*$ is a target point and $\mathbf{v_w^*}$ denotes the laser incidence vector calibrated by formulation.~\ref{eq_calib_main}, which is represented by multiplying a unique vector ($[0, 0, 1]$) and a rotation matrix represented by the two rotation angles (without "twisting" for the laser vector).  
The IK solver is used to estimate the optimal laser configuration to visit each point.
% (on the surface). 
% 
Gradient information of the objective function is provided from the MATLAB Symbolic Math Toolbox to speed up the optimization process.
As the laser trajectory can be formulated with individual points associated with robot configurations, the trajectory optimization can be formulated as: 
\begin{equation}
    \begin{aligned}
        \min_{ \Theta = \{ ~ \beta_{x,k}, ~ \beta_{y,k} ~ \}_{i=1}^N } \quad F_{traj} = \sum_k ~ || ~ g( ~ \beta_{x,k}, ~ \beta_{y,k} ~ ) - \mathbf{p}_k^* ~ ||_2^2   
    \end{aligned}
    \label{eq_laser_traj_problem}
\end{equation}
Here $\Theta = \{ ~ \beta_{x,k}, ~ \beta_{y,k} ~ \}_{i=1}^N$ denotes the waypoint coordinates of the laser trajectory at the reference plane. 
The $k$ is the index of the point $\mathbf{p}_k^* \in \mathbb{R}^3$ on the surface targets. 
The gradient $\frac{\partial F_{traj}}{\partial \Theta}$ of the objective function was provided to solve Equation.~\ref{eq_laser_traj_problem} based on the \texttt{Interior-Point} algorithm from the MATLAB optimization solver. 
This trajectory planner generates a sequence of optimal laser configurations to trace tumor targets. 

\subsection{Design of experiment for murine tumor study}
\label{method_doe_murine}

\subsubsection{Animal study protocol}
\label{method_doe_murine_animal_protocol}
\textcolor{black}{
Animal studies were conducted under the protocol and guidance described by IACUC and the Division of Laboratory Animal Resources of Duke University.  
}

\subsubsection{Mouse data preparation}
\label{method_doe_murine_data_preparation}
In this study, two murine tumor models were selected: soft tissue sarcoma (STS) and osteosarcoma (OS). 
The inducible Kras G12D/+, p53 flox/flox murine model \cite{kirsch2007spatially} was used.
Adenovirus-Cre was injected intramuscularly (for STS) and intraosseously (for OS) in the left hind leg of six- to eight-week-old mice. 
Approximately two months post-injection, palpable tumors formed, and mice were sacrificed. 
The tumorous and surrounding healthy muscle tissue were isolated from the femurs, and the dermis and epidermis were removed to expose the layer of tumor tissue.   
% 
% Although many tumors typically reside deep within tissue, we isolated the tumors so the majority of the mass was contained beneath the superficial tissue layer.

\subsubsection{Robot-guided tumor dataset collection}
\label{method_doe_murine_robot_data_collection}
For data collection, we developed a robot-guided sensor system to automatically collect data from tumor and healthy samples. 
A tabletop camera (Prosilica GC1020, 1024 x 768 resolution, 16 mm lens) was used to capture the image of the TumorID 405 nm laser spot for each data point.
TumorID performed a raster scan with fixed orientation (aligned with specific experiments) and distance from the target tissue, based on a scanning region of $8.0~mm \times 8.0~mm$ (step size: about $1.14$ mm), resulting in a total of 64 points for each tissue sample. 
The laser spot image and the fluorescence signal were recorded for each data point. 

\subsubsection{Tumor labeling}
\label{method_doe_murine_tumor_labeling}
Similar to oncological surgeries utilizing visual and palpation-based approaches to localize the tumor boundary in real-time \cite{lazarides2016fluorescence}, we used expert annotation \textbf{(provided by Beatrice Schleupner)} to mark out regions of tumor tissue on the leg containing palpable tumors from the image. 
Specifically, the labeling process began with an expert identifying the tumor location by palpating the mouse leg and sacrificing the specimen. 
Then, the expert estimated the tumor region, and the region was delineated on the corresponding images. 
A fluorescence data point associated with a laser spot within the highlighted region (Appendix.~\ref{appendix_mice_roi_tumor_and_healthy}) was assigned a label based on its spatial location (inside or outside the tumor region).  
The control leg was applied with a similar process to obtain pure normal tissue without tumor mass.
In summary, an expert labeled all tumorous and healthy regions for the dataset with the same workflow. 

\subsubsection{Tumor classifier}
\label{method_doe_murine_tumor_classier_detail}

We used a multilayer perceptron model (MLP) architecture to train the fluorescence spectrum data for binary classification (tumor or healthy). 
This prediction model was trained using a database consisting of OS- and STS- tumor samples. 
%  with 150 epochs and a batch size of 16, based on
The MLP model was trained with a three-layers multilayer perceptron network (Fig.~\ref{appendix_MLP_architecture_and_setting}) with configurations of 1024, 512 and 256 neurons.
\label{mlp_preprocessing_use}
\noindent
\textbf{Preprocessing:}
This section discusses the machine learning model used in the workflow of Fig.~\ref{workflow_of_tumor_mapping}. 
To generate the sparse pathological map, a common training and inference pipeline was devised for both types of tumor (OS and STS). 
% % 
For each data point interrogated by TumorID, spectral information was collected. 
A preprocessing step was used to remove bias stemming from the heterogeneity of tissues, the presence of blood, moisture, variation of tissue contour, laser stability, and spectrometer limits. 
The spectral signal between $450$~nm $- 750$~nm was kept, as this represents the effective fluorescence wavelengths of interest. 
The cutoff signal was performed with maximum normalization and smoothed with a Savitzky-Golay filter \cite{zachem2025intraoperative}.
Before training, the data were randomly split after stratification with appropriate weighting.

\noindent
\textbf{Training:}
% (1*N) 
The filtered and smoothed data (from Section.~\ref{mlp_preprocessing_use}) were used as (1*N) input to the MLP model where N represents the number of wavelength bins, with a tissue pathology label as output. 
During training, the dataset wide normalization was performed.
% 
% 
% % 
The model was trained for 150 epochs with a batch size of 16, using the Adam optimization algorithm and a learning rate of $1e^{-3}$. 
The softmax layer was used to get the final probability and the loss function was implemented with the Negative Log Likelihood Loss method for the binary classification label.
% 
% % 
%

% 
%
\noindent
\textbf{Model deployment:}
The models used in the murine tumor resection experiments were trained on a Nvidia RTX 2070 Super (8 GB) GPU (Nvidia Corporation, CA, USA) and implemented on PyTorch (high-performance deep learning library). 
During the experiment deployment, only the CPU (i9-11900K processor, Intel Corporation, CA, USA) was used for inference. 
For the murine data analysis of Table.~\ref{ml_result_40_degree} and Table.~\ref{ml_result_90_degree}, the models were trained with the NVIDIA RTX A2000 12GB GPU. 
All models were trained and tested on Windows 10 operating systems.

\subsection{Design of experiment for the histopathology analysis}
\label{method_doe_histopathology}
Histopathological analysis was provided to verify the classification accuracy and performance of TumorID compared to gold standard method.
The tumor and healthy data points were collected and labeled with colorized dyes to mark fiducial points for comparison after H\&E staining. 
Once the H\&E processed slides were processed and prepared, high-quality scanned images of histology slides were obtained. 
A trained pathologist \textbf{(provided by Jeffrey Everitt, DVM)} blindly labeled each marked dye spot without prior knowledge of mouse ID or tumor type. 
Each spot was assigned a label between healthy and tumorous. 
Instances with unclear information were not included in the analysis. 
In case both healthy and tumor cells were present within $200 \mu m$ depth of the colorized spot, it was labeled as a tumor. 
As the 405-nm laser penetration depth was limited, any tumor cell beyond the range was not considered. 
Slices generated after the tumor resection experiment (Method.~\ref{method_tumor_map_laser_resection}) were also qualitatively analyzed for the presence of thermal damage. 
These results were used as ground truth and compared with the predictions from the tumor classifier.

\subsubsection{Post-labeled sample processing for H\&E histopathology}
% \subsubsection{Automated tumor labeling: line-scanning for pathology analysis}
% 
\label{method_doe_histopathology_tumor_label}
The postlabeled sample preservation process involved the gentle spraying of vinegar on the tissue sample for dye fixation, transferring of the tissue samples to histological cassettes, soaking in phosphate-buffered saline solution for 24 hours, followed by transferring to $10\%$ neutral buffer formalin (NBF) for tissue fixation. 
The tissue samples underwent substantial shrinkage during the formalin fixation step. 
NBF also bleached part of the colorized dye markers.
With tissue shrinkage, the spatial relationships between marked points were lost, and identifying the spot where the laser biosensor had been scanned before became very challenging to correlate in the absence of colorized dyes.
Hematoxylin and eosin were used to stain the tissue slides. 
For converting the tissue specimen to histology slides, the specimen was sliced at \textcolor{black}{$50~ \mu m$} slice width. 
To obtain accurate tissue information in the presence of dye, we assume that the pathology of cells right below the dye up to a depth of $200 ~\mu m$ corresponds to the prediction obtained from the pathology prediction pipeline. 
This is because 405-nm light is almost completely attenuated beyond $200 ~\mu m$ depth in tissue. 
Moreover, the top surface of the tissue information was occluded by the dye. 
We devised a two-step process for slicing the specimens. 
In the first step, the technician removed unnecessary tissue parts that did not contain any dye, and cut the center of the dye color spot. 
The cut specimen was rotated by \textcolor{black}{$90$ degree} to expose the tissue right below the dye. 
This exposed the scanned tissue region and allowed the pathologist to make a judicious decision. 
In summary, each H\&E slice provided a representation of all the single-point information, where a pathologist would label them individually with the corresponding data from the fluorescence spectrum. 
Each H\&E slice was correlated with a classification label to provide absolute ground truth. 

\section*{Acknowledgment}

The authors thank Paula Newell for support of histopathological analysis.  
This work was supported by the National Institutes of Health (NIH) under Award R01EB030982. 
The content is solely the responsibility of the authors and does not necessarily represent the official views of the National Institutes of Health.

\section{Author Contributions}

GM, RP, WCE and PJC conceptualized the study. 
GM and RP contributed to system design, algorithm and software development, implementation of experiments, and data analysis.
BS and WCE contributed to mouse tumor preparation and tumor data labeling.
RP, BS and GM contributed to the pathological processing of mice tumors. 
JE contributed to the pathological analysis and labeling.
GM and RP drafted the paper. 
GM, RP, AM, BS, JE, JC, BM, BC, LB, PZ, MD, WCE and PJC contributed to the edits and comments of the paper. 
WCE and PJC supervised the project. 
All authors revised and approved the final paper.

\section{Competing Interests}

The authors declare that they have no competing interests.

\section{Additional information}

The code will be made publicly available and datasets can be requested upon request from the authors.

% 
% 

% \bibliographystyle{IEEEtran}
% \bibliography{IEEEexample.bib}


\clearpage 
\newpage

% \subfile{appendix}

% 

% 

\section{Supplementary Materials}

\vspace{+6mm}

\begin{center}
% \centering
\textbf{ {\fontsize{20pt}{24pt}\selectfont TumorMap: A Laser-based Surgical Platform for 3D Tumor Mapping and Fully-Automated Tumor Resection} }
\end{center}

\begin{center}
    \vspace{+6mm}
    \normalfont \normalsize
    Guangshen Ma, PhD,\textsuperscript{1,2 $^{\dagger}$, *}
    Ravi Prakash,\textsuperscript{1, $^{\dagger}$, *}
    Beatrice Schleupner,\textsuperscript{3}
    Jeffrey Everitt, DVM,\textsuperscript{4}
    Arpit Mishra, PhD,\textsuperscript{1} \\
    Junqin Chen, PhD, \textsuperscript{1} 
    Brian Mann, PhD, \textsuperscript{1}
    Boyuan Chen, PhD, \textsuperscript{1}
    Leila Bridgeman, PhD, \textsuperscript{1}
    Pei Zhong, PhD, \textsuperscript{1} \\
    Mark Draelos, MD, PhD, \textsuperscript{2,5} 
    William C. Eward, DVM, MD \textsuperscript{3}
    and
    Patrick J. Codd, MD\textsuperscript{1,6, $^{\dagger}$}
    \\[18pt]

\itshape
\textsuperscript{$*$}G. Ma and R. Prakash contributed equally to this work. \\
\textsuperscript{$^{\dagger}$} Corresponding email: guangshe@umich.edu, ravi.prakash@duke.edu, patrick.codd@duke.edu. \\
\textsuperscript{1}Thomas Lord Department of Mechanical Engineering and Materials Science, Duke University\\
\textsuperscript{2}Department of Robotics, University of Michigan, Ann Arbor\\
\textsuperscript{3}Department of Orthopaedic Surgery, School of Medicine, Duke University\\
\textsuperscript{4}Department of Pathology, School of Medicine, Duke University\\
\textsuperscript{5}Department of Ophthalmology and Visual Sciences, University of Michigan Medical School, Ann Arbor\\
\textsuperscript{6}Department of Neurosurgery, School of Medicine, Duke University
% \\[18pt]

\end{center}

\vspace{+3mm}
\begin{flushleft}
    \noindent
    \textbf{ {\fontsize{12pt}{12pt}\selectfont The Supplementary Documents incorporate the following: } } 
    \begin{itemize}
        \item Appendix of Materials and Methods.
        \item Appendix of Results.
    \end{itemize}
\end{flushleft}

% \\[18pt]
\vspace{+3mm}
\begin{flushleft}
    \noindent
    \textbf{ {\fontsize{12pt}{12pt}\selectfont Other Supplementary Materials for this manuscript: } } 
    \begin{itemize}
        \item Movie S1 (main video).
        \item Movie S2 (main workflow).
        \item Movie S3 (data collection method).
        \item Movie S4 (\textit{ex vivo} experiments). 
        \item Movie S5 (mice experiments).
        \item Movie S6 (histopathology experiments).
    \end{itemize}
\end{flushleft}

\clearpage
\newpage

\subsection{System calibration and kinematics modeling}
\subsubsection{OCT-to-camera calibration generalization testing}
\label{appendix_general_test_oct_to_cam_calib}
\begin{figure}[H]
\centering
\includegraphics[width = 0.98\textwidth]{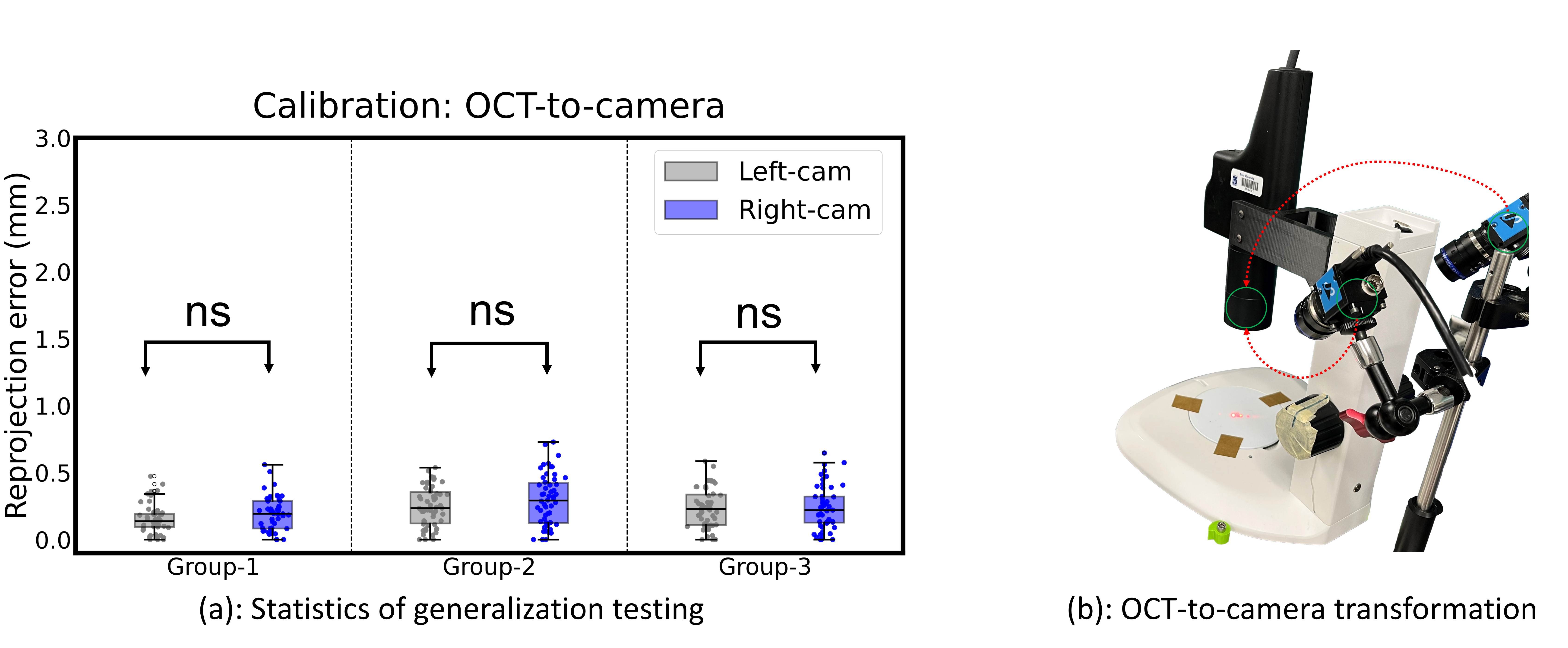}
\caption{\textbf{Statistics of OCT-to-camera calibration for the generalization testing.}
\textbf{a}: Three repeated studies were conducted in this simulation experiment to evaluate system calibration performance of the two cameras. 
\textbf{b}: The transformation between the camera to the OCT sensor. 
($*: p < 0.05$, $**: p < 0.01$, $***: p < 0.001$, $ns$: no significant).
}
\label{fig_general_test_oct_to_cam}
\end{figure}

Further testing was conducted to evaluate the accuracy of the OCT-to-camera calibration algorithm (Method.~\ref{method_calibration_oct_to_camera}) through three repeated studies. 
The calibration method was applied for the left and right cameras with three groups of data. 
The error statistics for the system calibration are reported in Fig.~\ref{fig_general_test_oct_to_cam}a. 
The errors for the three groups are reported as $0.21 \pm 0.12$ mm, $0.27 \pm 0.18$ mm for the left and right cameras, respectively. 
A two-sample t-test was performed and the results show that there are no significant differences among the three data groups, indicating that the algorithm performs reliably for both cameras. 

\subsubsection{OCT-to-laser calibration generalization testing}
\label{appendix_general_test_oct_to_laser_calib}
\begin{figure}[H]
\centering
\includegraphics[width = 0.95\textwidth]{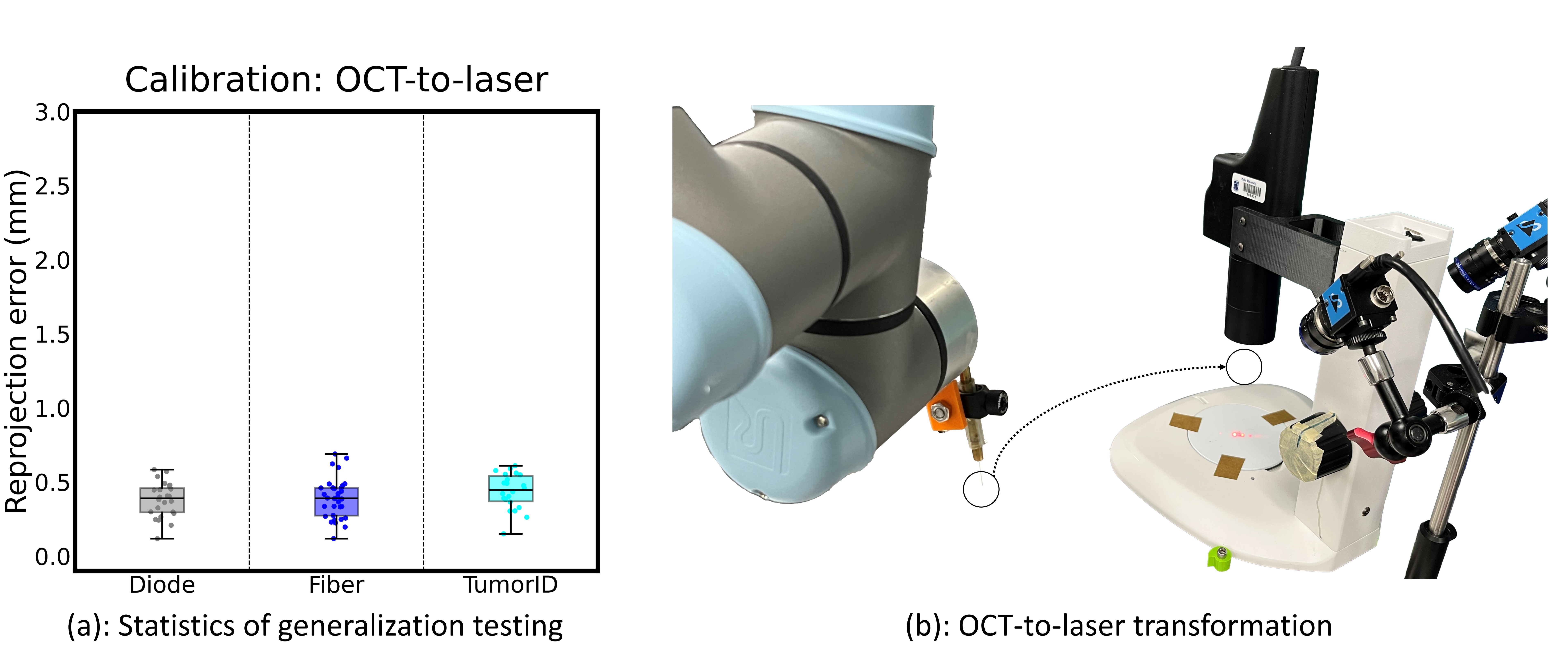}
\caption{
\textbf{Statistics of OCT-to-Laser calibration for the generalization testing. }
\textbf{a}: The studies were conducted for the three types of laser including a 650 nm laser diode, 405 nm laser diode and 1940 nm fiber-coupled laser. 
The result indicates that there is no statistically difference among the group of laser-diode, fiber-coupled laser and TumorID-laser. 
\textbf{b:} The working space of the laser scalpel is within the camera and the OCT scanning region. 
}
\label{fig_general_test_oct_to_laser}
\end{figure}

We conducted the further testing to evaluate the OCT-to-laser calibration algorithm discussed in Method.~\ref{method_calibration_oct_to_laser}.
We repeated the studies for the OCT-to-laser calibration for three different laser types, including the laser diode (green), TumorID (blue), and the fiber-coupled laser. 
The error statistics are reported in Fig.~\ref{fig_general_test_oct_to_laser}a.
The errors for the three groups are reported as $0.38 \pm 0.12$ mm, $0.39 \pm 0.13$ mm, $0.44 \pm 0.12$ mm for the laser-diode, the fiber-coupled laser and the TumorID laser. 
A two-sample t-test was performed and the results show that there are no significant differences among the three groups of data. 
This indicates that the algorithm performs in a reliable way for the OCT-to-laser calibration.  

\newpage
\clearpage

\subsubsection{Generalization testing: optimization-based inverse kinematics and robot planning}
\label{appendix_general_test_ik_and_planner}
\begin{figure}[H]
\centering
\includegraphics[width = 0.95\textwidth]{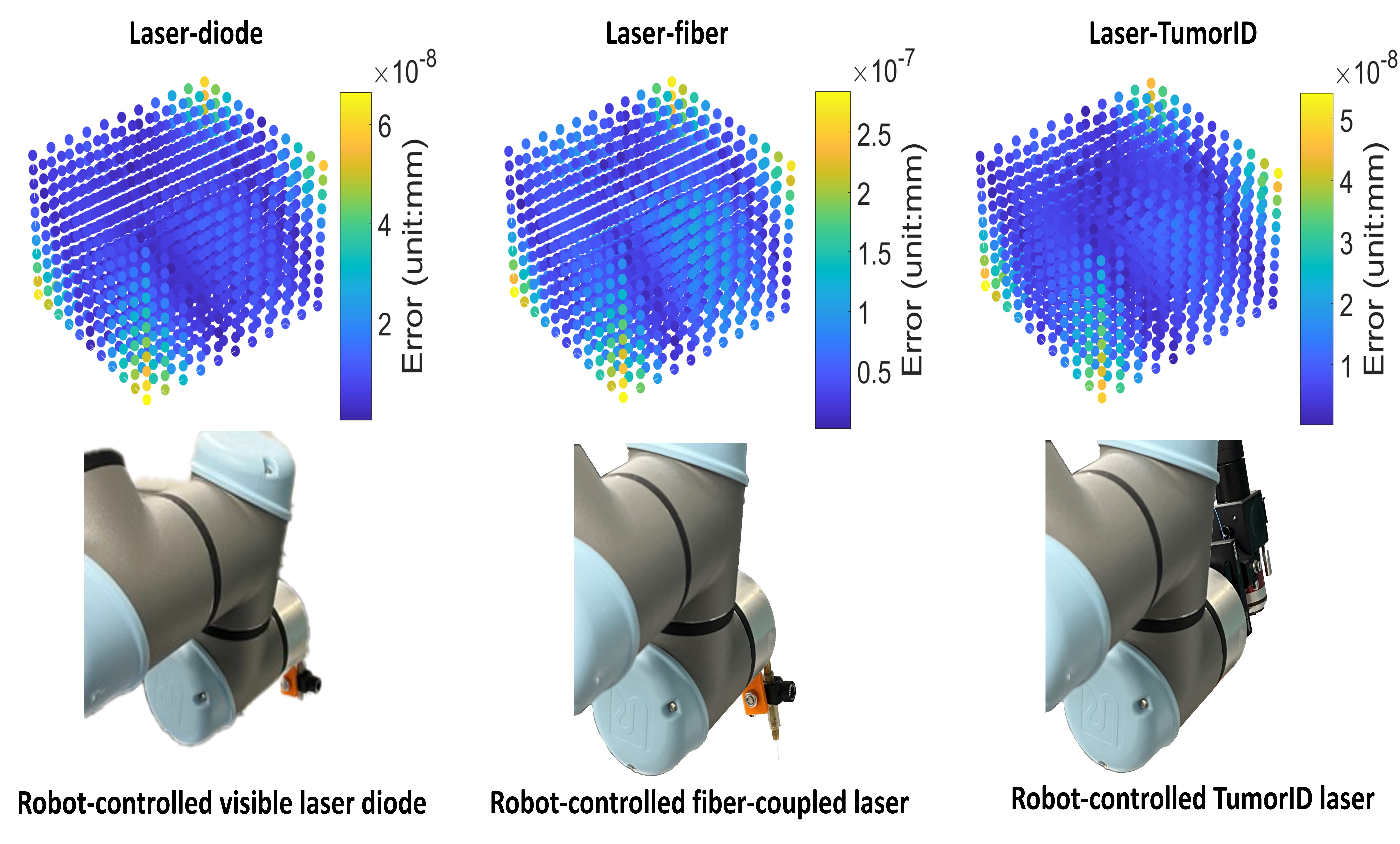}
\caption{
\textbf{Statistics of IK general testing for the laser-diode, laser-fiber and laser-TumorID.}
The 3D heatmaps shows the errors between the predictions (through calculations from the IK solver) and the ground truth (synthetic targets in the simulator). 
Each error was assigned with a color based on a range limit. 
The maximum error is smaller than the $10^{-6}$ mm.
% (in the order of of magnitude 10-5 mm).
% 
}
\label{fig_general_test_ik_solver}
\end{figure}

We conducted the generalization testing to verify the algorithm proposed in Method.~\ref{method_system_model_ik_and_traj_plan} with three calibrated laser types of laser diode, fiber-coupled laser, and TumorID. 
First, a 3D mesh grid was generated to cover the scanning region and serve as 3D targets (for the IK solvers), which simulated the realistic scanning region defined in the actual experiment.   
The IK solver was used to calculate the optimal robot configurations to visit each target point.
We repeated the experiments for the three lasers.
The error was calculated on the basis of the absolute point-based distance between the prediction and the 3D targets.
For the analysis, the prediction-to-target error was calculated for each point to formulate the 3D error heatmaps (Fig.~\ref{fig_general_test_ik_solver}a-c).
For the three types of lasers, the maximum error of all the results is smaller than $10^{-6}$ mm. 
This suggests that the IK solver is capable of calculating the correct robot joint angles to control the laser to visit the targets.
This simulation experiment shows the feasibility of the proposed IK solver being used in various laser configurations. 

\clearpage
\newpage

\subsection{Phantom Studies}

\subsubsection{Fiducial marker tracking}
\label{appendix_result_of_marker_task}
\begin{figure}[H]
\centering
\includegraphics[width = 0.62\textwidth]{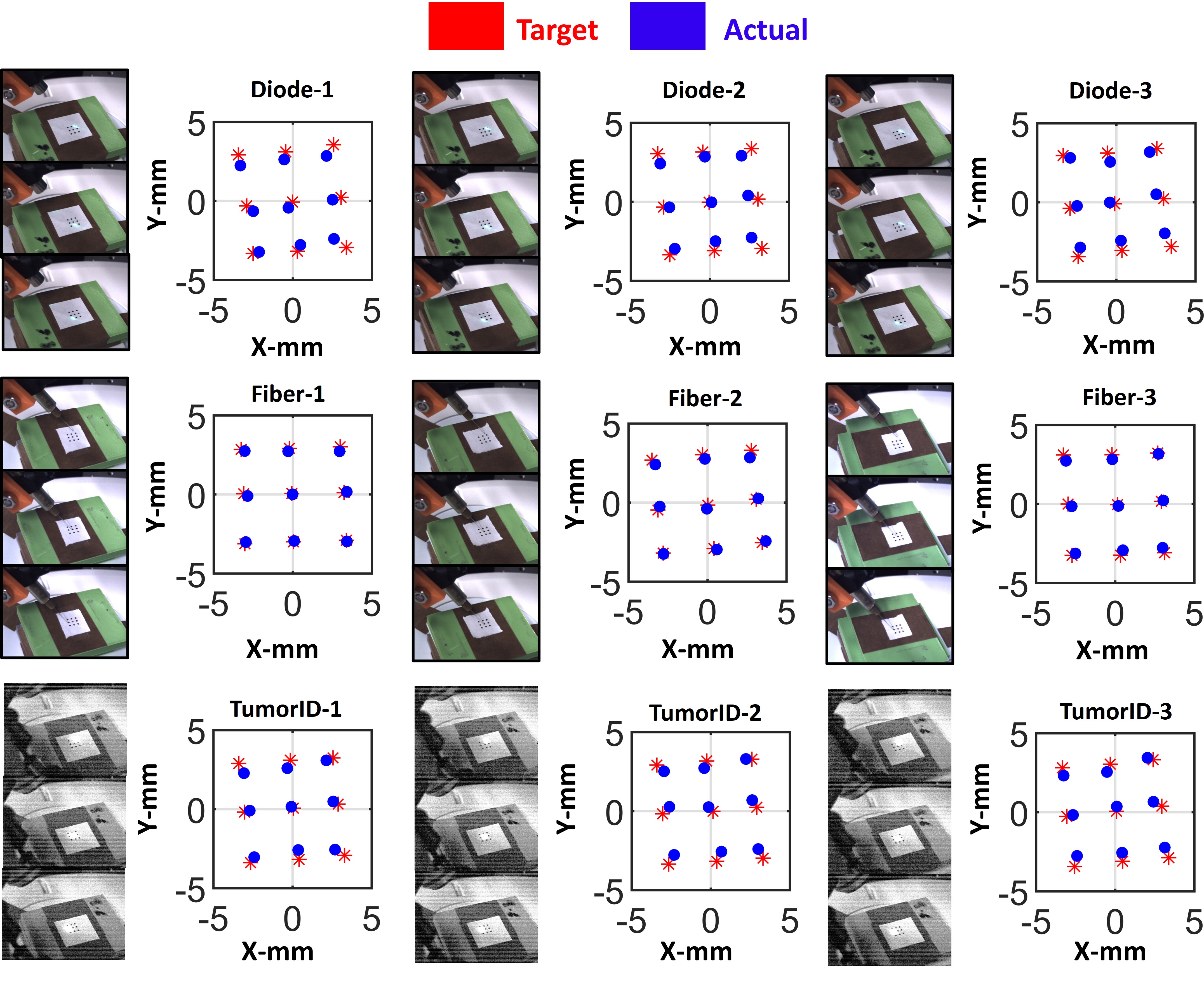}
\caption{
\textbf{Supplementary results of the phantom experiment in Fig.~\ref{fig_3_system_exp}.}
For the experiment, the calibration board were placed in different heights.
For the TumorID results, the exposure time was adjusted to highlight the visible spot from the camera in order to accurately detect the laser spot from the system.
For error statistics, the offsets of the two centers can be the represented end-to-end errors for the TumorMap system (red: target; blue: actual cutting points). 
The targets were first labeled by the operator from the 2D image (after adjusting the image intensity to highlight the pixel difference).
The 2D labeled centers were transformed to 3D coordinates in the OCT frame (Method.~\ref{method_calibration_oct_to_camera}). 
The resulted 2D-to-3D points were projected to the 2D plane along the Z-axis of the OCT frame.
Thus, the point-to-point error can be measured in millimeter.
}
\label{fig_system_result_fid}
\end{figure}

\subsubsection{Trajectory tracking}
\label{appendix_result_of_traj_task}

\begin{figure}[h]
\centering
\includegraphics[width = 0.62\textwidth]{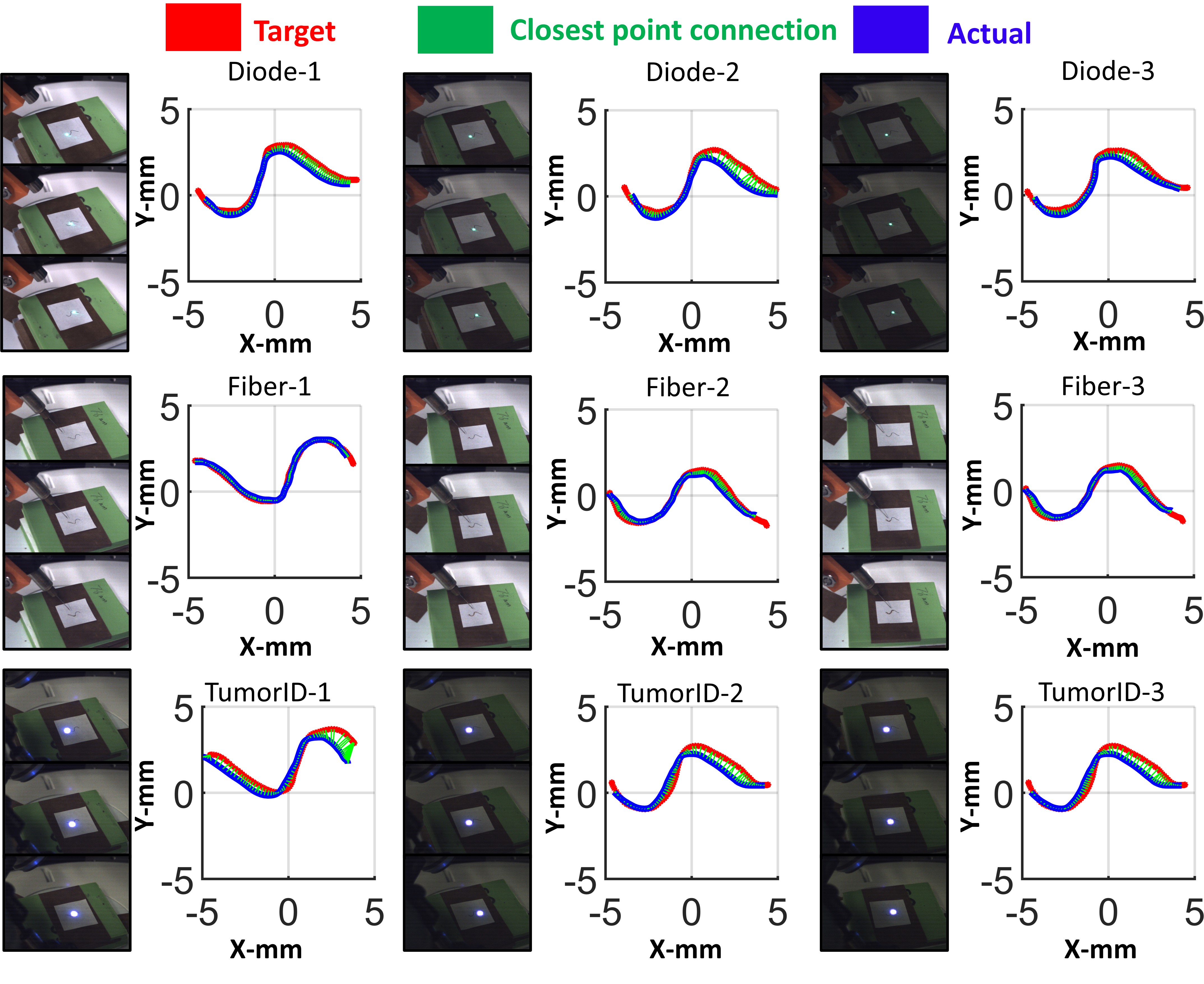}
\caption{
\textbf{Supplementary results of the trajectory tracking experiment in Fig.~\ref{fig_3_system_exp}. }%
Three repeated studies were conducted with three different heights for the calibration boards.
The first column shows the reference images in different time steps towards the complete trajectory tracking. 
The green lines denote the connected points from the detected closest points between the two paths. 
The point correspondence was calculated by looking for the nearest neighbor candidates between the actual and target trajectories. 
For the TumorID laser testing, the exposure time was adjusted to highlight the visible spot from the camera in order to accurately detect the laser spot from the system (visualization only). 
}
\label{fig_system_result_traj}
\end{figure}

\subsubsection{ROI tracking task}
\label{appendix_result_of_roi_task}

\begin{figure}[H]
\centering
\includegraphics[width = 0.75\textwidth]{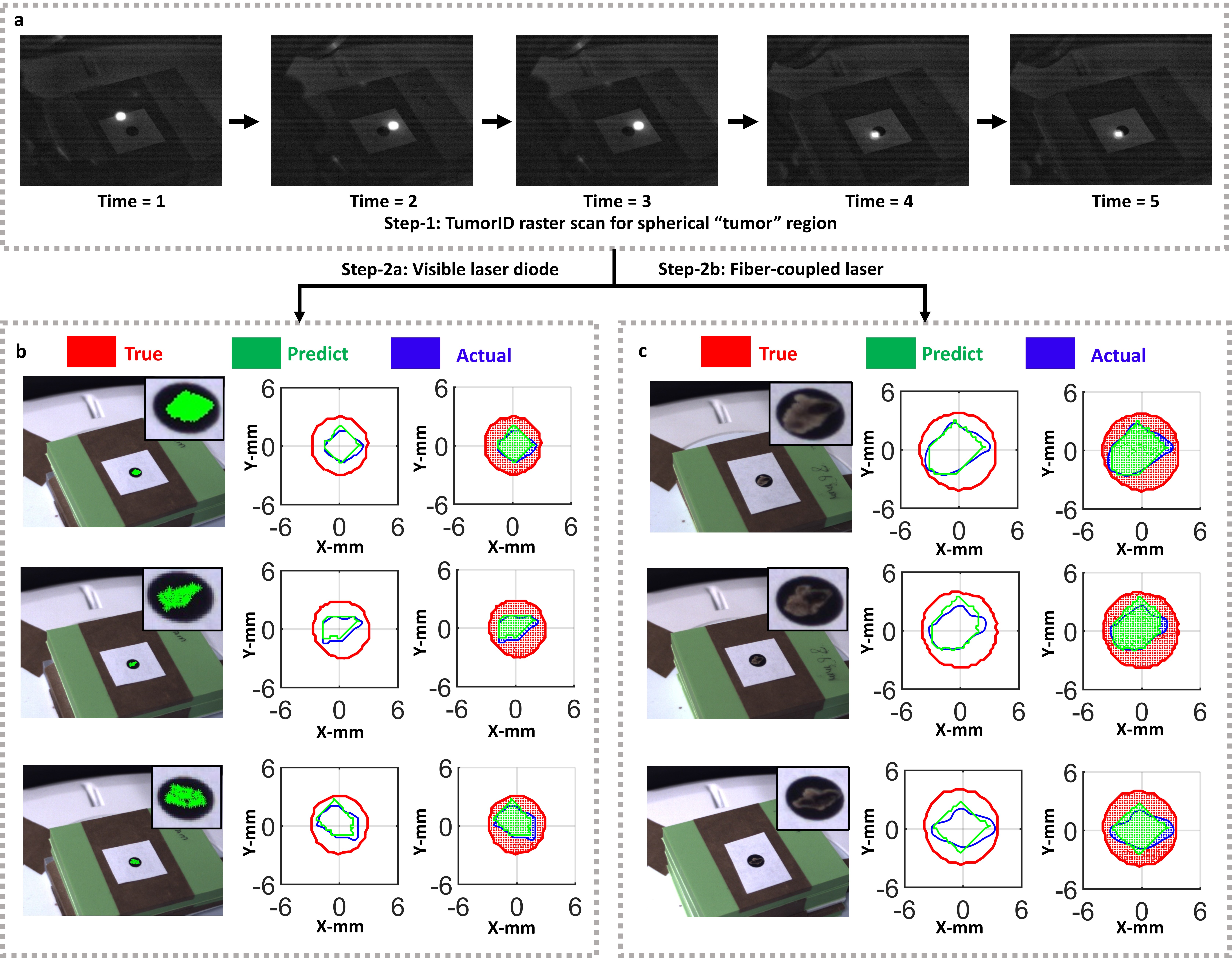}
\caption{
\textbf{Supplementary results for the ROI tracking task. }
We summarize the results of the phantom study for tracking the spherical ROI in Fig.~\ref{fig_system_result_roi}.
\textbf{(a):} The tumor searching procedure with the TumorID sensor (laser spot highlighted). 
The TumorID was controlled to first scan towards the entire surface to determine the tumor boundary (prediction algorithms). 
The 3D tags (labels) were provided from the simple threshold-based classifier from the fluorescence data. 
The tumor boundary was formulated by using the convex hull algorithm (Method.~\ref{method_model_tumor_roi_geometry}). 
A laser trajectory was calculated by the proposed trajectory planner (Method.~\ref{method_system_model_ik_and_traj_plan}). 
Both the visible laser diode and the fiber-coupled laser were used for generalization testing. 
\textbf{(b):} The green spot indicates the ``ablated region" of the visible laser spot. 
\textbf{(c):} The ablated region with the fiber-coupled laser with a clear boundary compared with the other regions. 
}
\label{fig_system_result_roi}
\end{figure}

\clearpage
\newpage

\subsection{\textit{Ex vivo} Tissue Study}
\label{appendix_exvivo_img_dataset}

\subsubsection{Classification performance of the \textit{ex vivo} tissue model}
\begin{figure}[H]
\centering
\includegraphics[width = 0.75\textwidth]{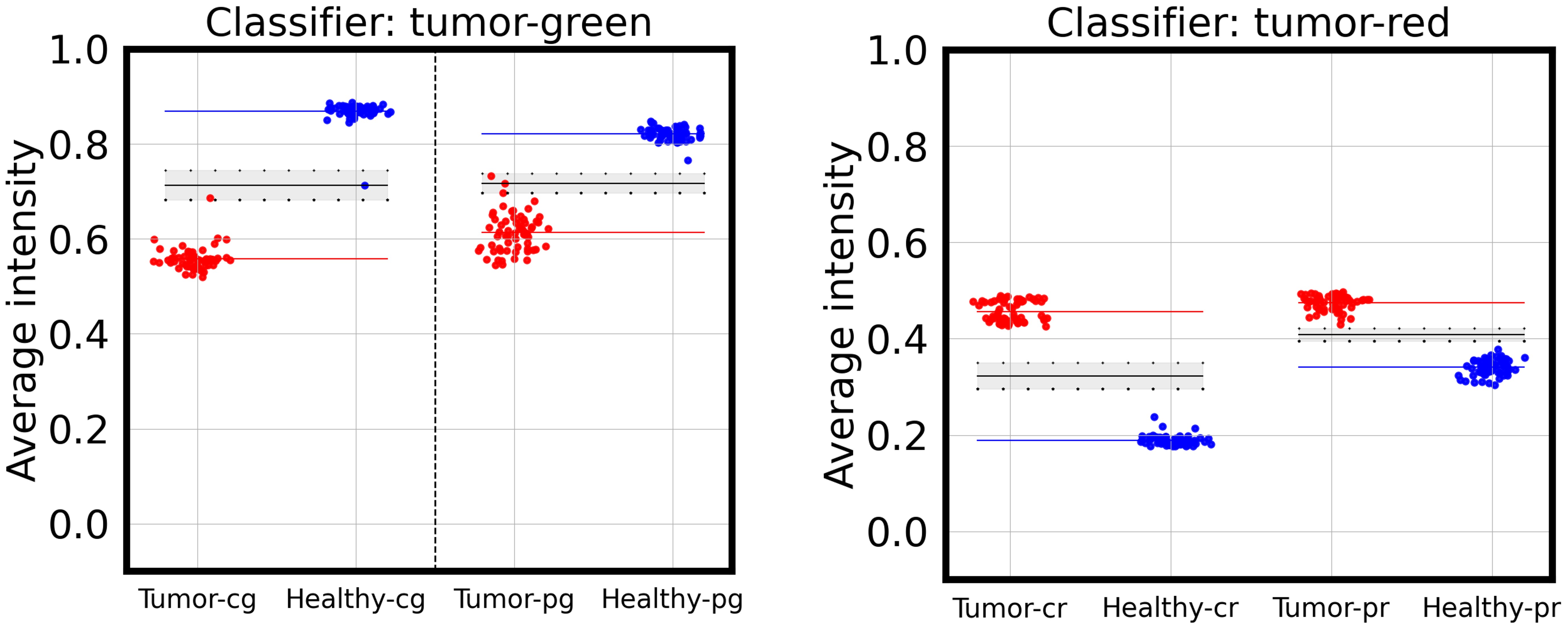}
\caption{
\textbf{Tumor classification graph for the \textit{ex vivo} experiment in Fig.~\ref{fig_4_exvivo_exp}}. 
This classification graph summarizes the \textit{ex vivo} tissue models for the models of chicken-green (cg), chicken-red (cr), porcine-green (pg), porcine-red (pr). 
This is a simple binary classifier to calculate the average intensity of the TumorID signal within a specialized wavelength range. 
}
\label{appendix_exvivo_classification_graph}
\end{figure}

\subsubsection{Summary of the data collection for \textit{ex vivo} tissue (chicken model)}
\begin{figure}[H]
\centering
\includegraphics[width = 0.90\textwidth]{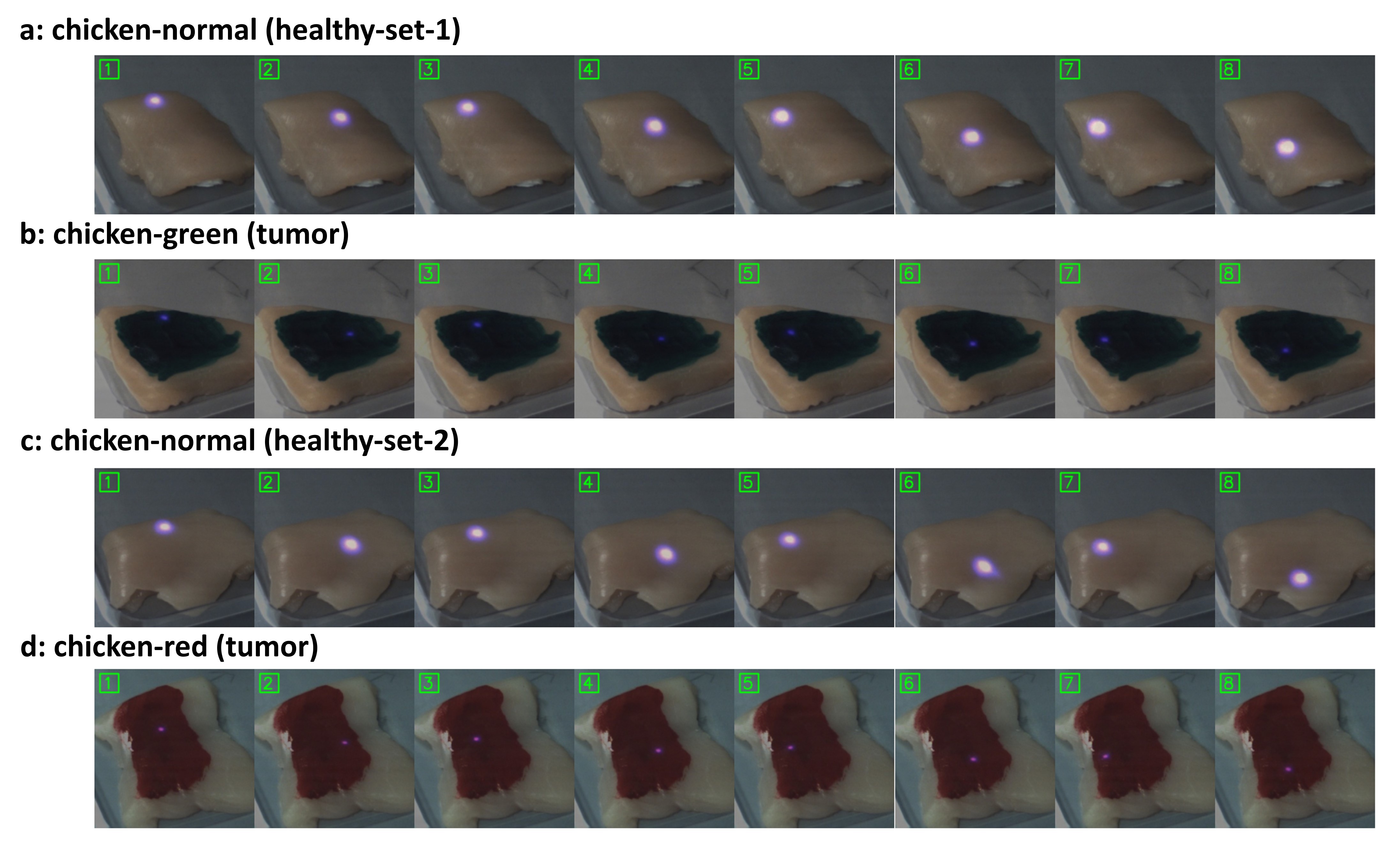}
\caption{
\textbf{Summary of training datasets for the supplementary results in Fig.~\ref{fig_4_exvivo_exp} (\textbf{chicken tissue model})}. 
\textbf{a:} Chicken normal tissue dataset (set-1). 
\textbf{b:} Chicken tumor model (chicken-green).  
\textbf{c:} Chicken normal tissue dataset (set-1). 
\textbf{d:} Chicken tumor model (chicken-red).  
The images were selected uniformly in the dataset for visualization. 
The green and red colorized dyes were used to simulate the artificial tumor models with a homogeneous configuration (points within the regions are clustered with the same labels). 
For the healthy samples, only normal chicken tissue data points were collected without dyes. 
}
\label{fig_dataset_chicken}
\end{figure}

\subsubsection{Summary of the data collection for the results in Fig.~\ref{fig_4_exvivo_exp} \textit{ex vivo} tissue (porcine model)}
\begin{figure}[H]
\centering
\includegraphics[width = 0.90\textwidth]{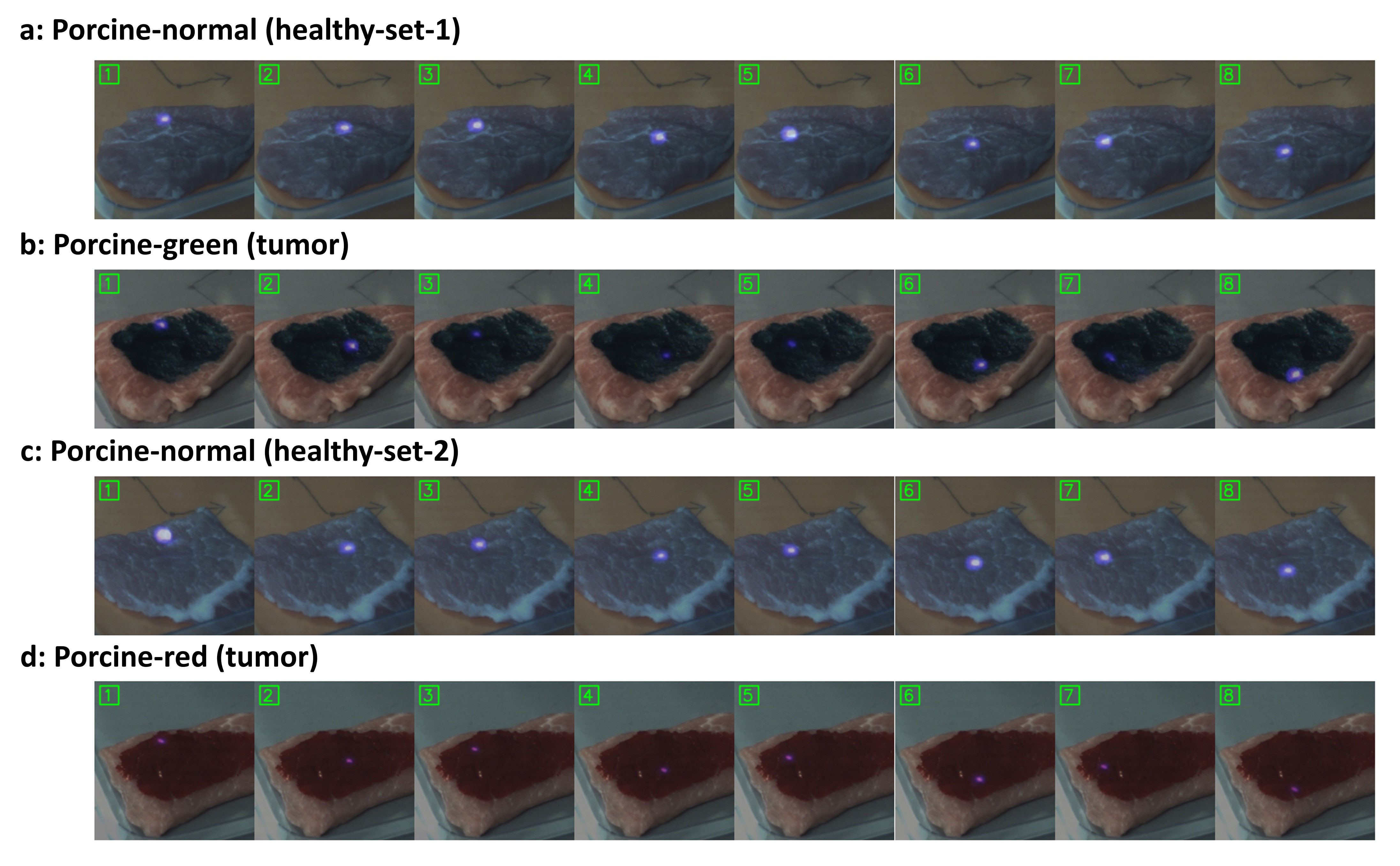}
\caption{
\textbf{Summary of training datasets for the supplementary results in Fig.~\ref{fig_4_exvivo_exp} (\textbf{porcine tissue model}).} 
\textbf{a:} Porcine normal tissue dataset (set-1). 
\textbf{b:} Porcine tumor model (porcine-green).  
\textbf{c:} Porcine normal tissue dataset (set-1). 
\textbf{d:} Porcine tumor model (porcine-red).  
The images were selected uniformly in the dataset for visualization. 
The green and red colorized dyes were used to simulate the artificial tumor models with a homogeneous configuration (points within the regions are clustered with the same labels). 
For the healthy samples, only normal porcine tissue data points were collected without adding any colorized dyes. 
}
\label{fig_dataset_porcine}
\end{figure}

\clearpage
\newpage

\subsubsection{Summary of the resection experiment for \textit{ex vivo} tissue (visible laser diode)}
\begin{figure}[H]
\centering
\includegraphics[width = 0.95\textwidth]{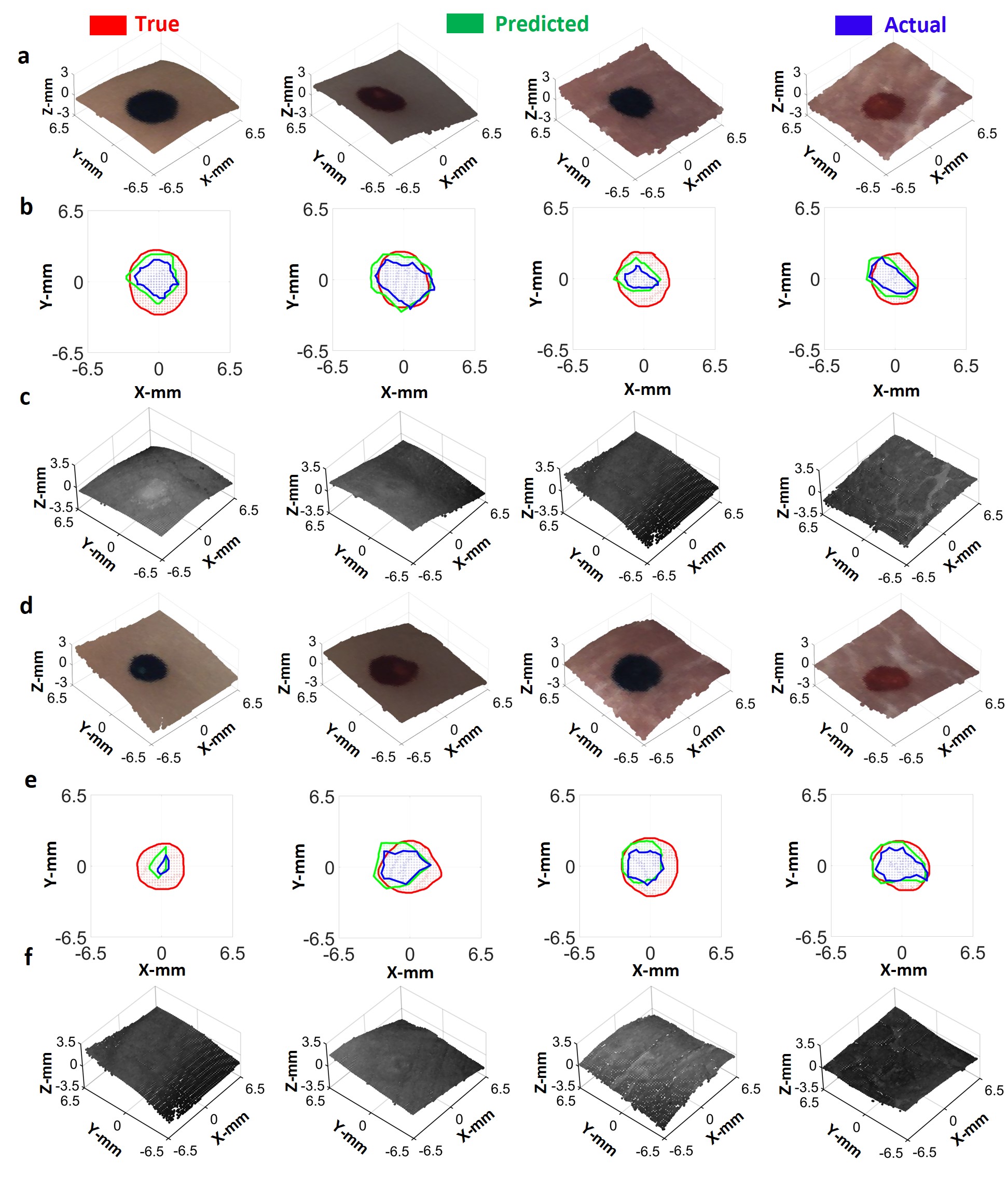}
\caption{
\textbf{Summary of \textit{ex vivo} tissue experiments of the visible laser-diode (unit: mm). }
\textbf{a:} Trial-1: 3D colorized maps for the \textit{ex vivo} tissue model.
%
% % 
\textbf{b:} Trial-1: ROI performance with the true (red), predicted (green) and actual (blue) regions. 
% % 
\textbf{c:} Trial-1: OCT surface image of the pre-resection scanning. 
% % 
\textbf{d:} Trial-2: 3D colorized maps for the \textit{ex vivo} tissue model.
% 
% 
% % 
\textbf{e:} Trial-2: ROI performance with the true (red), predicted (green) and actual (blue) regions. 
% % 
\textbf{f:} Trial-2: OCT surface image of the pre-resection scanning. 
% % 
}
\label{fig_res_exvivo_set_1}
\end{figure}

\subsubsection{Summary of the resection experiment \textit{ex vivo} tissue (fiber-coupled laser)}
\begin{figure}[H]
\centering
\includegraphics[width = 0.95\textwidth]{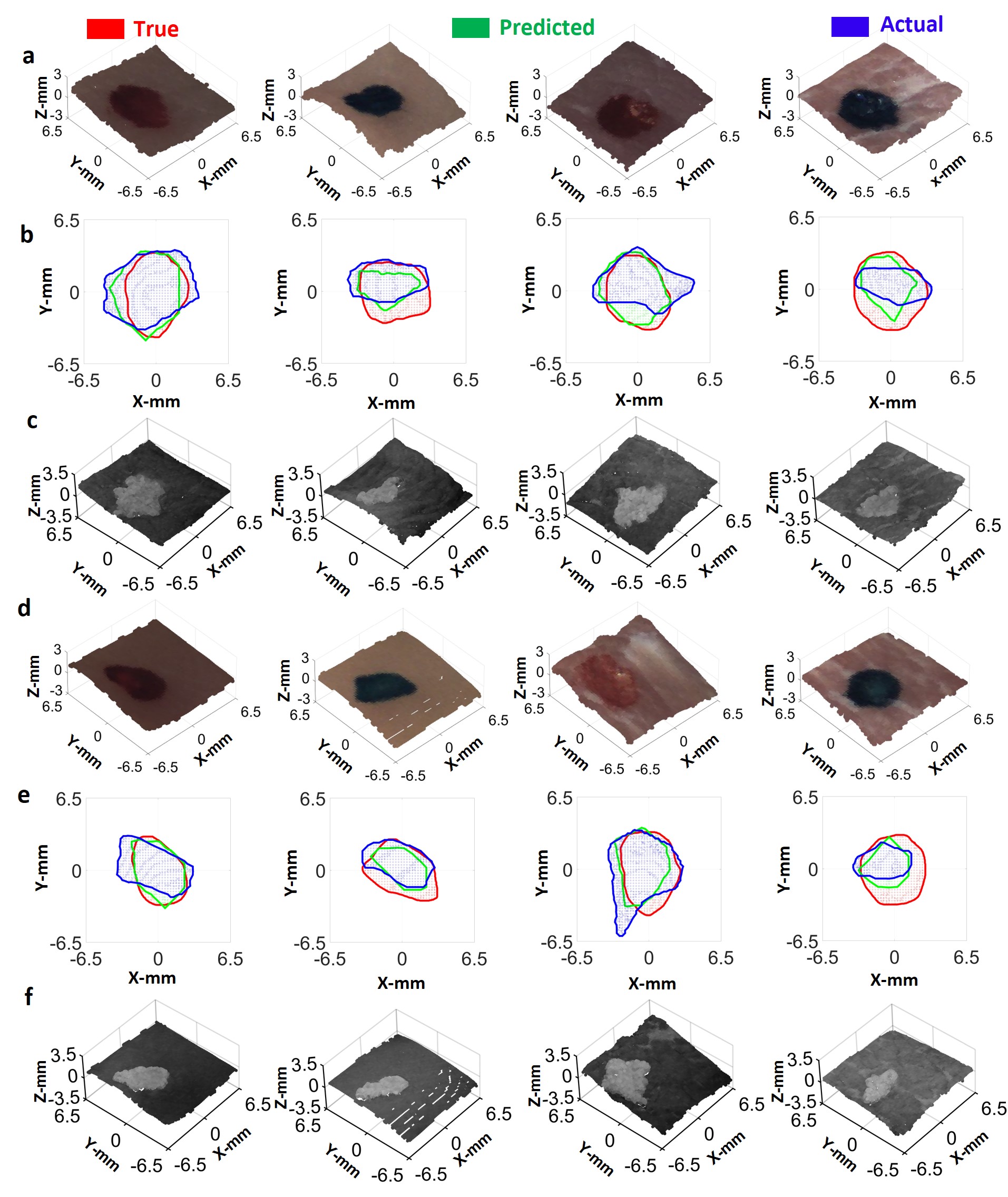}
\caption{
\textbf{ Summary of \textit{ex vivo} tissue experiments of the fiber-coupled laser (unit:mm). }
\textbf{a:} Trial-1: 3D colorized maps for the \textit{ex vivo} tissue model.
\textbf{b:} Trial-1: ROI performance with the true (red), predicted (green) and actual (blue) regions.
\textbf{c:} Trial-1: OCT surface image of the post-resection scanning. 
The intensity change on the surface correlates with the ablation features from the tissue resection.  
\textbf{d:} Trial-2: 3D colorized maps for the \textit{ex vivo} tissue model.
\textbf{e:} Trial-2: ROI performance with the true (red), predicted (green) and actual (blue) regions.
\textbf{f:} Trial-2: OCT surface image of the post-resection scanning. 
The intensity change on the surface correlates with the ablation features from the tissue resection.
}
\label{fig_res_exvivo_set_2}
\end{figure}

\subsection{Murine Tumor Experiments}

\subsubsection{Summary of the OS-tumor resection experiments}

\begin{figure}[H]
\centering
\includegraphics[width = 0.95\textwidth]{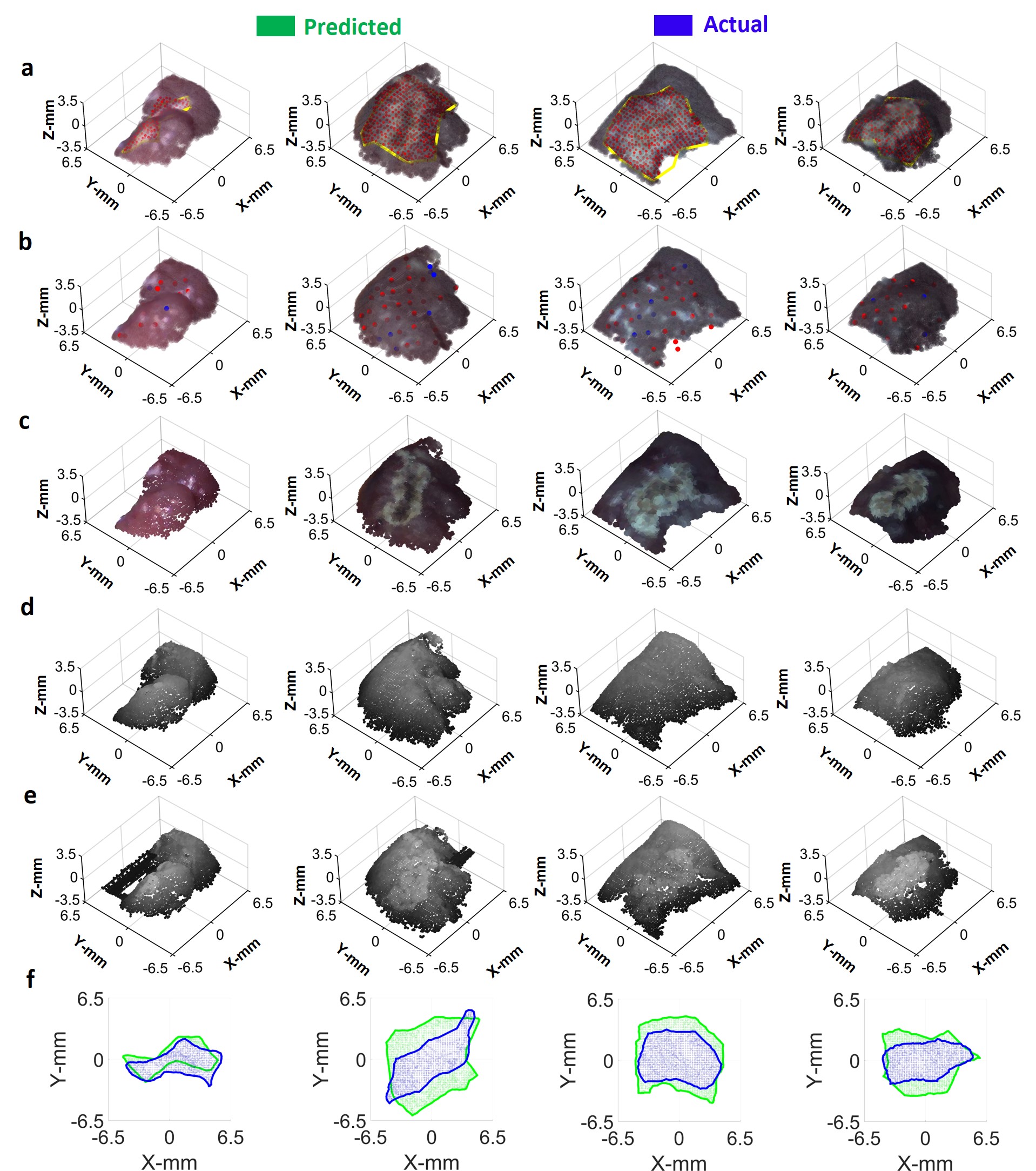}
\caption{
\textbf{Summary of \textit{mice tumor} tissue experiments (\textbf{OS-tumor} model).}
\textbf{a:} 3D colorized maps for the murine tumor model with the ablated regions and the highlighted tumor boundary. 
\textbf{b:} 3D tumor tags with point-based classification labels from the MLP models. 
\textbf{c:} 3D Colorized map for the post-resection image. 
\textbf{d:} OCT surface image of the pre-resection scanning. 
\textbf{e:} OCT surface image of the post-resection scanning. 
The intensity change on the surface correlates with the ablation features from the tissue resection.  
\textbf{f:} ROI evaluation between the predicted (green) and actual (blue) regions.
}
\label{appendix_mice_summary}
\end{figure}

\subsubsection{Summary of the STS-tumor resection experiments}

\begin{figure}[H]
\centering
\includegraphics[width = 0.95\textwidth]{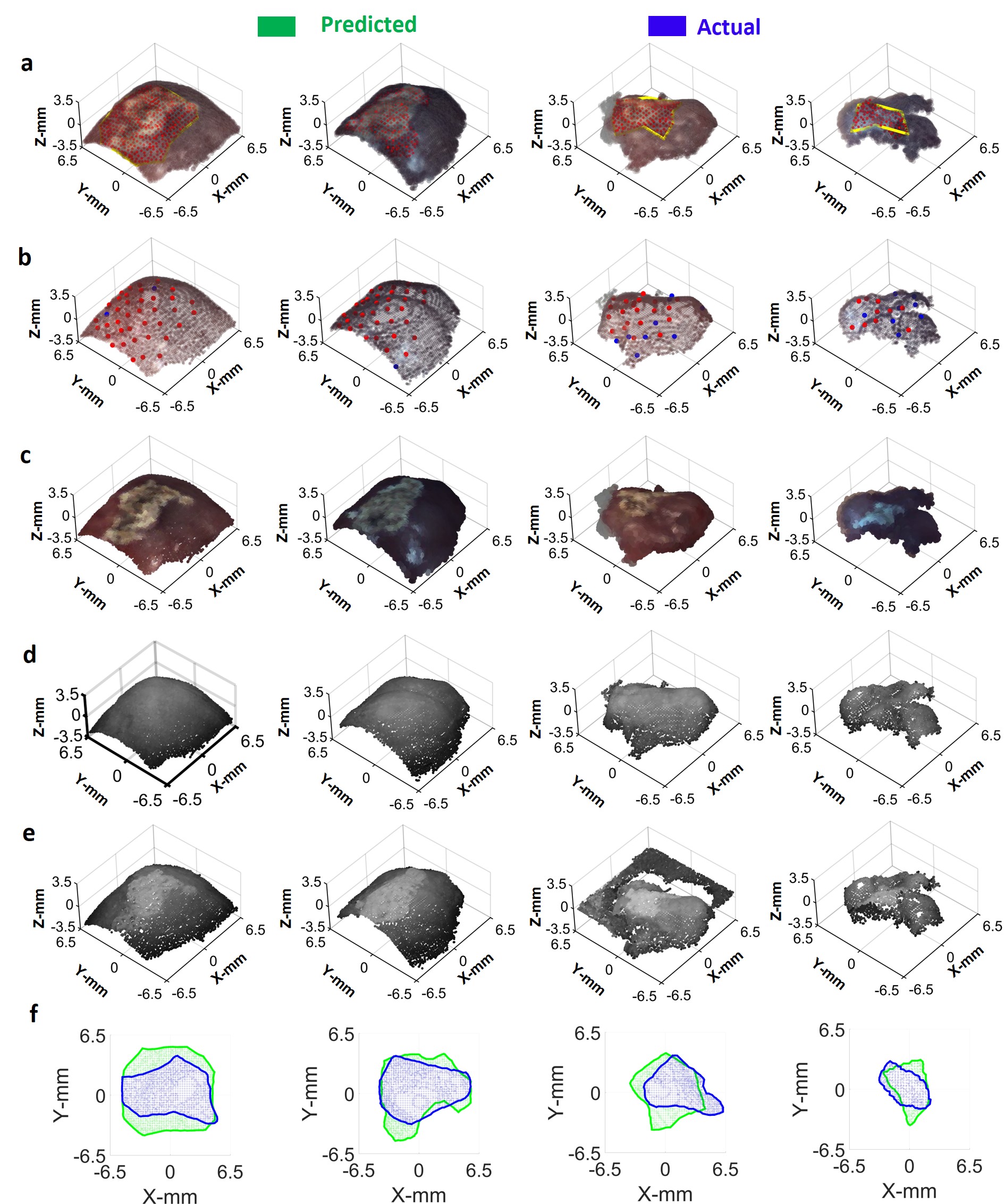}
\caption{
\textbf{Summary of \textit{mice tumor} tissue experiments (\textbf{STS-tumor} model).}
\textbf{a:} 3D colorized maps for the murine tumor model with the ablated regions and the highlighted tumor boundary. 
\textbf{b:} 3D tumor tags with point-based classification labels from the MLP models. 
\textbf{c:} 3D Colorized map for the post-resection image. 
\textbf{d:} OCT surface image of the pre-resection scanning. 
\textbf{e:} OCT surface image of the post-resection scanning. 
The intensity change on the surface correlates with the ablation features from the tissue resection.  
\textbf{f:} ROI evaluation between the predicted (green) and actual (blue) regions.
}
\label{appendix_mice_summary}
\end{figure}

\newpage
\clearpage

\subsection{\textbf{Summary of machine learning model}}
\label{appendix_MLP_architecture_and_setting}

\begin{figure}[H]
\centering
\includegraphics[width = 0.95\textwidth]{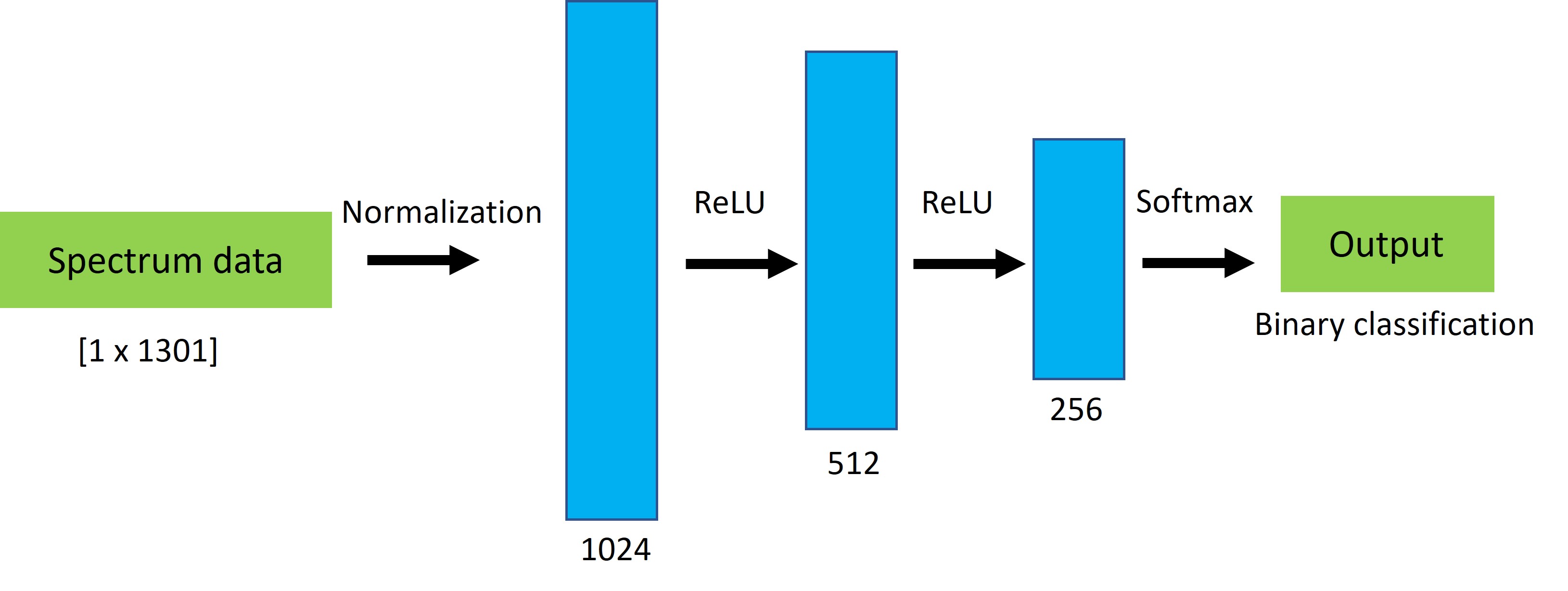}
\caption{\textbf{Tumor classifier model architecture.} 
The input signal is passed through a 3-layer multi-layer perceptron (MLP) architecture.
The softmax function is used in the last layer to produce a binary classification label to differentiate the tumorous and healthy samples. 
}
\label{mlp_architecture_model}
\end{figure}

\clearpage
\newpage

\subsubsection{Summary of the murine tumor samples for dataset formulation}

% mice-40-set-1
\begin{figure}[H]
\centering
\includegraphics[width = 0.85\textwidth]{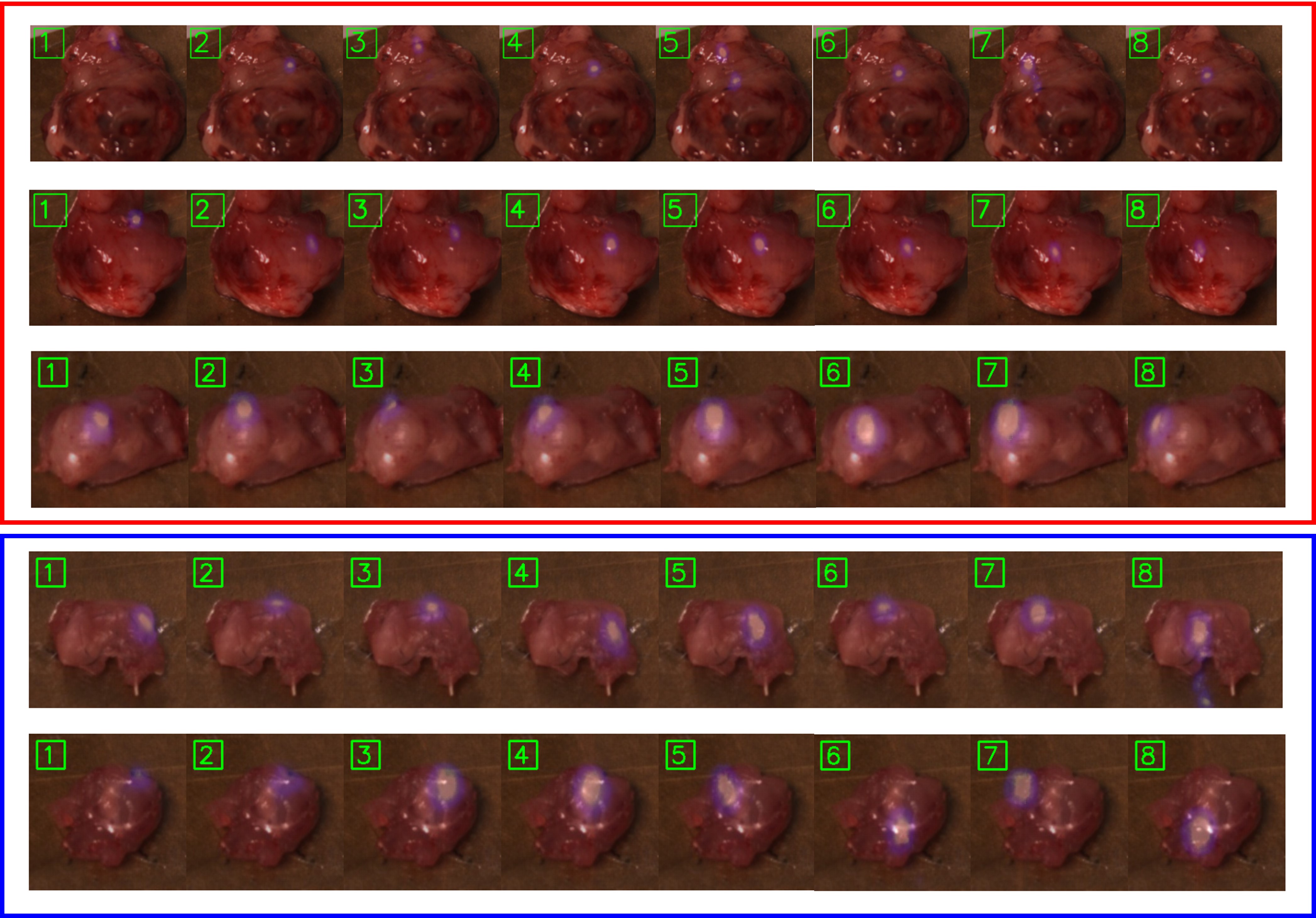}
\caption{
\textbf{Represented examples of dataset generation for tissue objects in Table.~\ref{ml_result_40_degree}}. 
The sequential images show laser spots overlaid on tissue samples. 
% (the first set of reference images for the experiment described in Table.~\ref{ml_result_40_degree}). 
% 
The tumor samples are within the red region and the healthy ones in the blue region.
}
\label{appendix_mice_40_set_1}
\end{figure}

\begin{figure}[H]
\centering
\includegraphics[width = 0.85\textwidth]{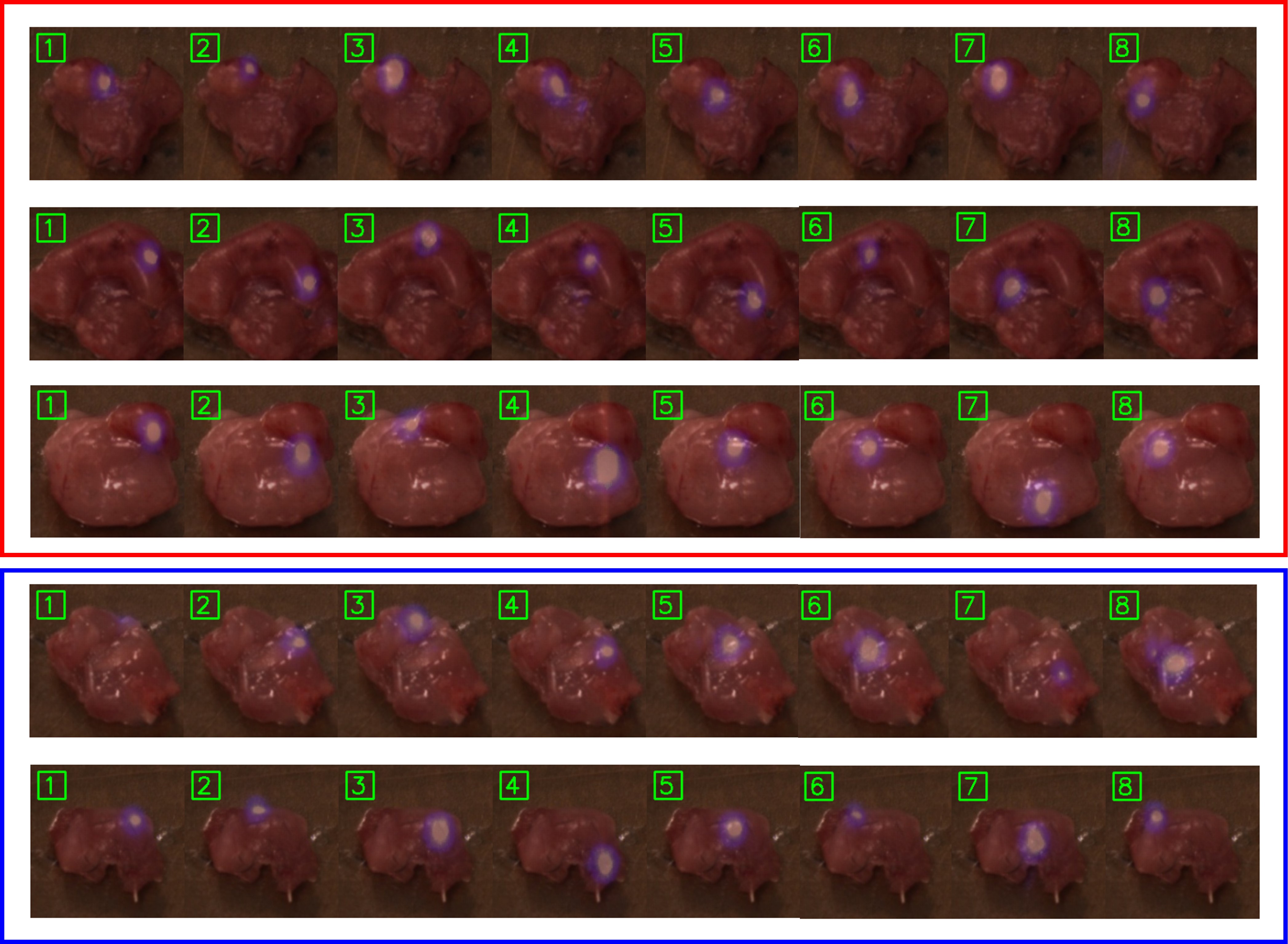}
\caption{
\textbf{Represented examples of dataset generation for tissue objects in Table.~\ref{ml_result_90_degree}}. 
The sequential images show laser spots overlaid on tissue samples. 
% (the first set of reference images for the experiment described in Table.~\ref{ml_result_90_degree}). 
% 
The tumor samples are within the red region and the healthy ones in the blue region.
}
\label{appendix_mice_40_set_1}
\end{figure}

\subsubsection{Summary of the histopathological analysis}

\begin{figure}[H]
\centering
\includegraphics[width = 0.98\textwidth]{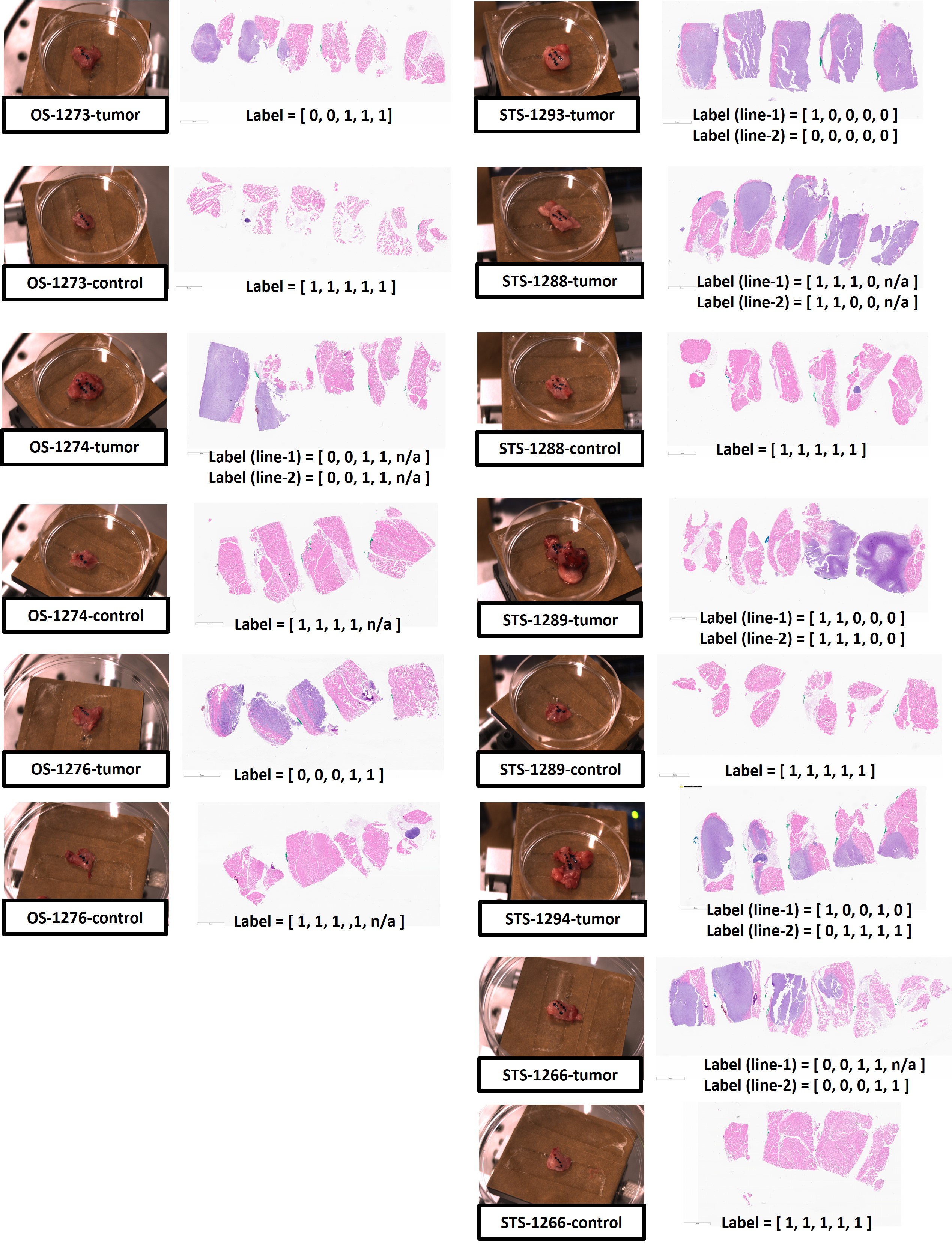}
\caption{
\textbf{Summary of the labeling locations for the murine tumor samples for histopathology analysis}. 
The robot-guided TumorID sensor was controlled to perform a line-scan towards the tumor surface.  
The line-trajectory was manually defined to cover both the tumorous and healthy regions in order to perform a non-biased evaluation of the tumor classifier. 
The labeled tumors undergo the hematoxylin and eosin staining technique (H\&E) processing to highlight regions of possible tumor presence. 
% 
% For the labels, $n/a$ in
}
\label{appendix_pathology_label_reference}
\end{figure}

\begin{table}[H] 
\centering 
\begin{tabular}{  >{\raggedright} m{3.5cm}  m{3.5cm}  m{3.5cm}  m{3.5cm} }      % centered columns (3 columns) 
\toprule                                   %inserts double horizontal lines 
\centering 
 \textbf{OS-tumor} & \textbf{STS-tumor} & \textbf{OS-control} & \textbf{STS-control} \\  % inserts table heading 
\midrule\addlinespace[0.3ex]
        \includegraphics[height=0.80in]{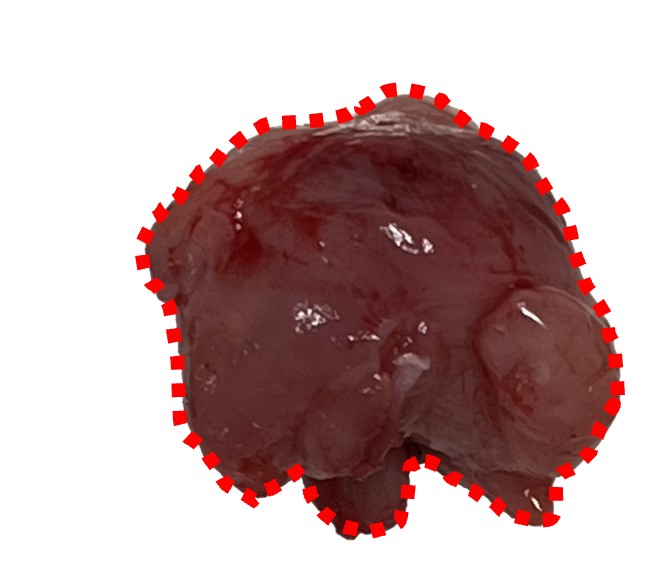} 
        & 
        \includegraphics[height=0.85in]{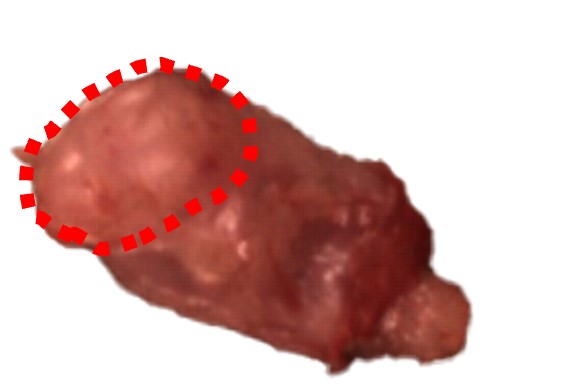} 
        & 
        \includegraphics[height=0.8in]{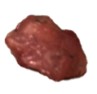} 
        & 
        \includegraphics[height=0.65in]{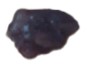}
        \\
\midrule\addlinespace[0.3ex]
        \includegraphics[height=1in]{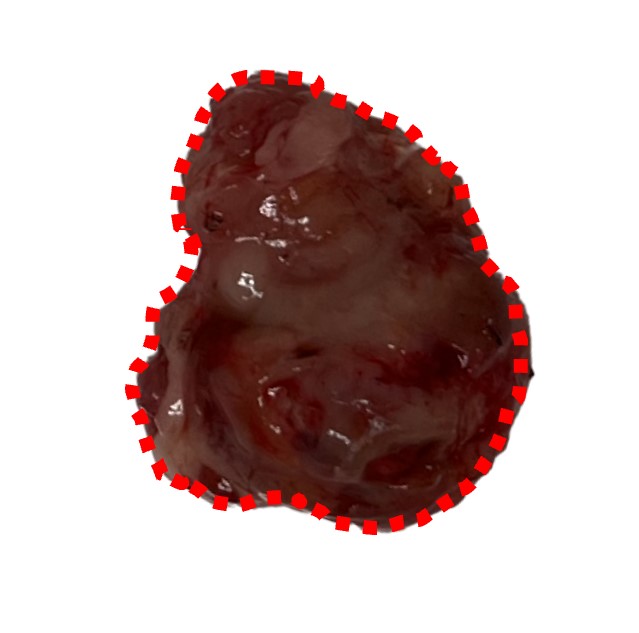} 
        & 
        \includegraphics[height=1in]{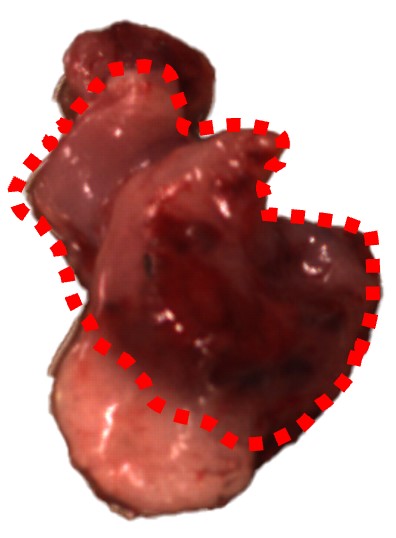}
        & 
        \includegraphics[height=0.8in]{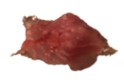} 
        & 
        \includegraphics[height=0.8in]{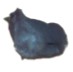}
        \\
\midrule\addlinespace[0.3ex]
        % idx-1241
        \includegraphics[height=1in]{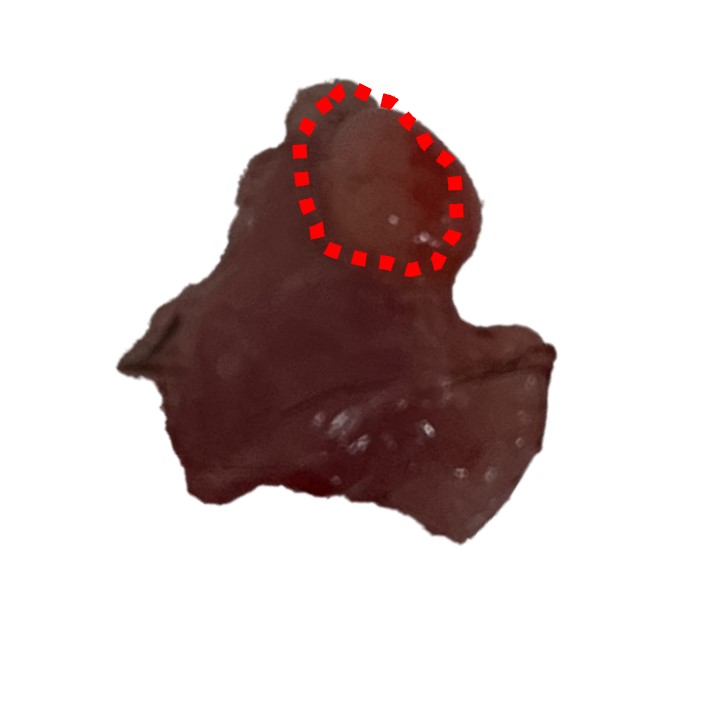} 
        & 
        \includegraphics[height=1in]{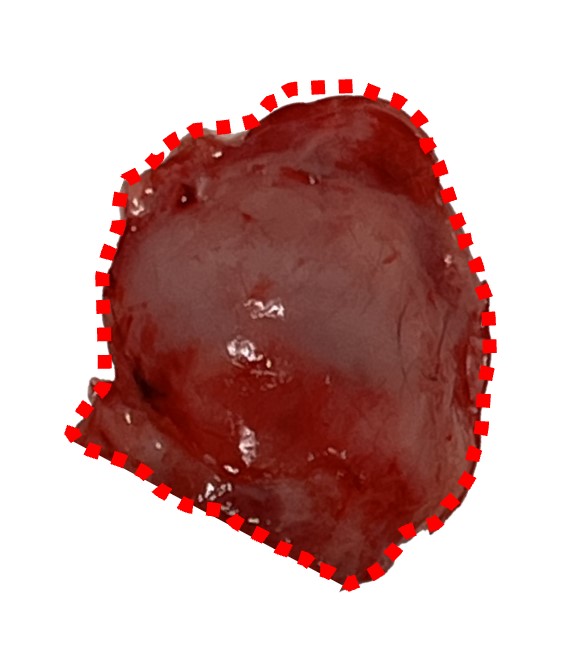} 
        & 
        \includegraphics[height=0.8in]{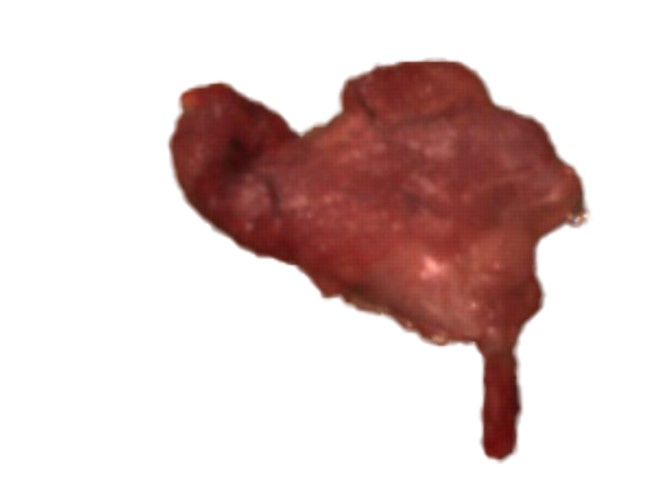} 
        & 
        \includegraphics[height=0.70in]{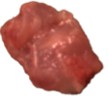} 
        \\
\midrule\addlinespace[0.3ex]
        % idx-1241
        \includegraphics[height=0.85in]{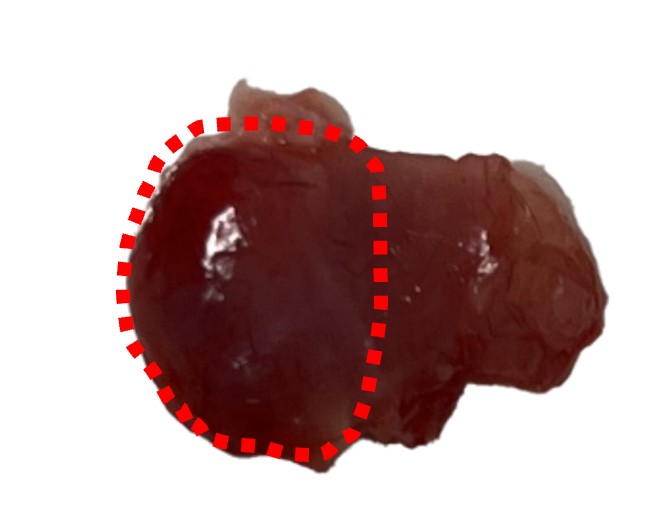} 
        & 
        \includegraphics[height=0.65in]{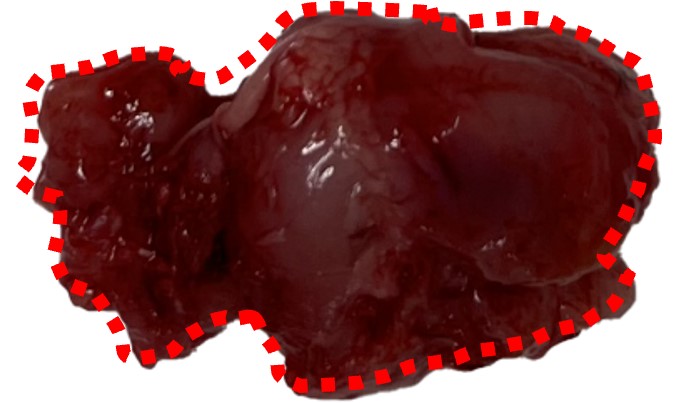} 
        & 
        \includegraphics[height=0.65in]{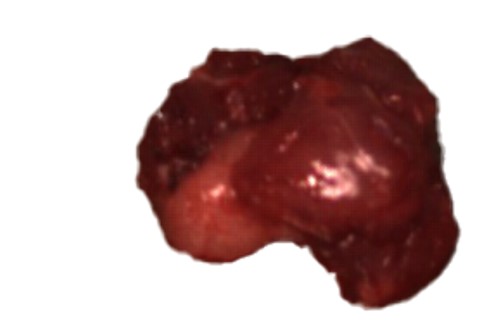} 
        & 
        \includegraphics[height=0.65in]{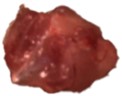}
        \\
\midrule\addlinespace[0.3ex]
        % idx-1241
        \includegraphics[height=0.65in]{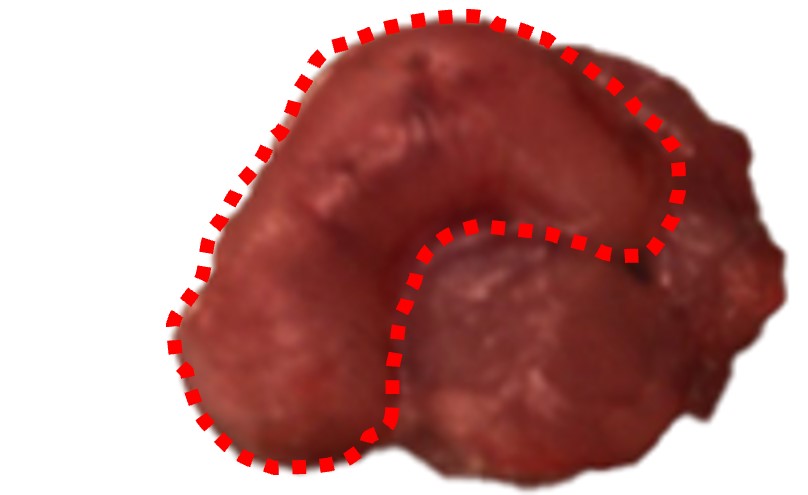} 
        & 
        \includegraphics[height=1in]{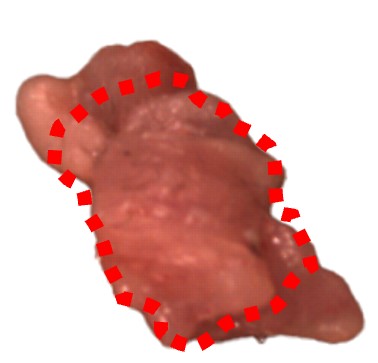} 
        & 
        \includegraphics[height=0.65in]{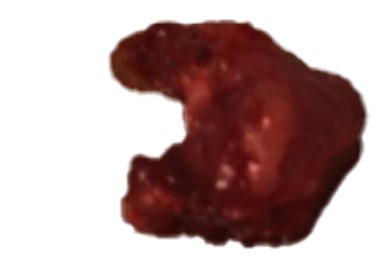} 
        & 
        \includegraphics[height=0.70in]{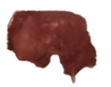}
        \\
\midrule\addlinespace[0.3ex]
        % idx-1241
        \includegraphics[height=0.65in]{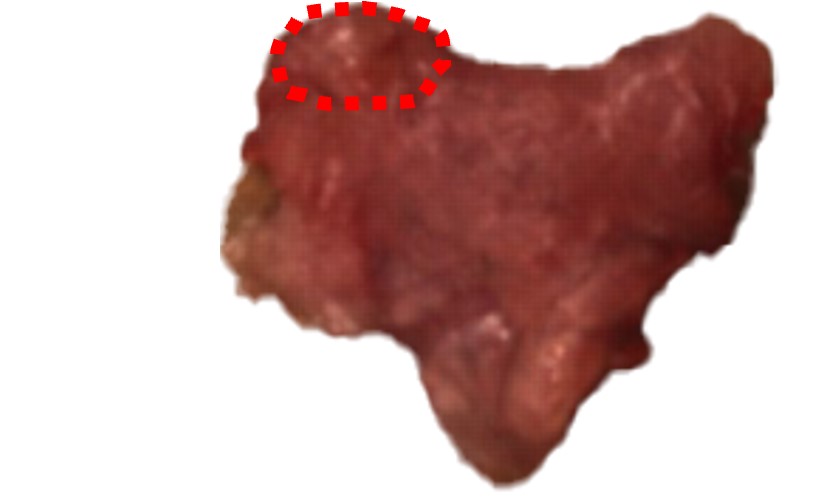} 
        & 
        \includegraphics[height= 0.85in]{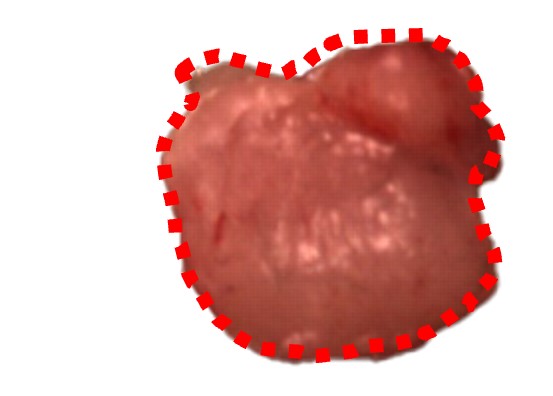} 
        & 
        \includegraphics[height=0.8in]{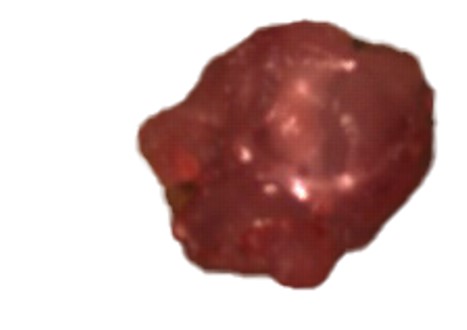} 
        & 
        \includegraphics[height=0.70in]{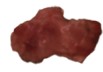}
        \\
\midrule\addlinespace[0.3ex]
        % idx-1241
        \includegraphics[height=1.2in]{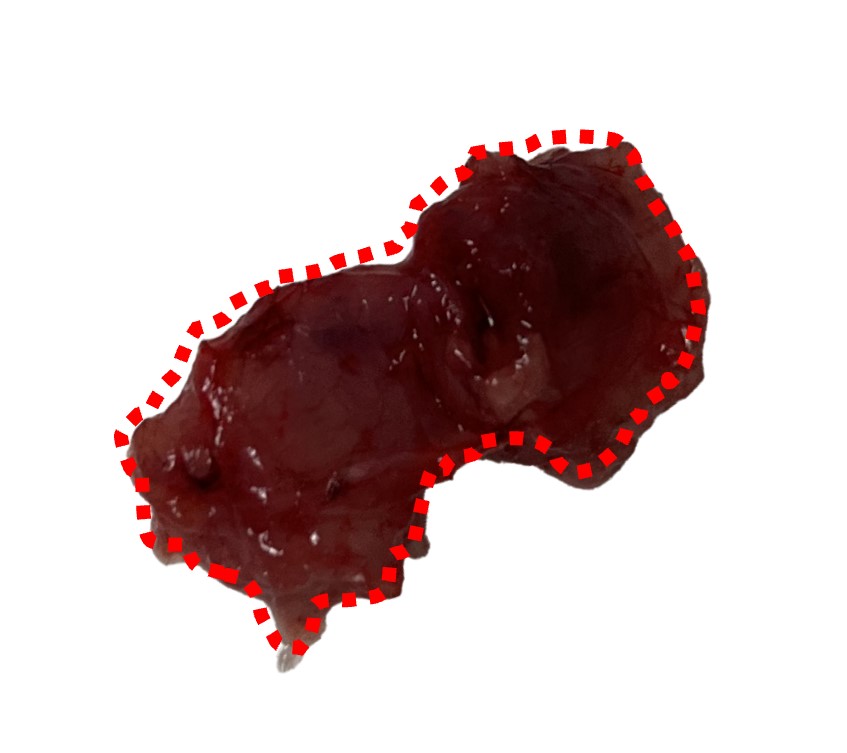} 
        & 
        \includegraphics[height=1.2in]{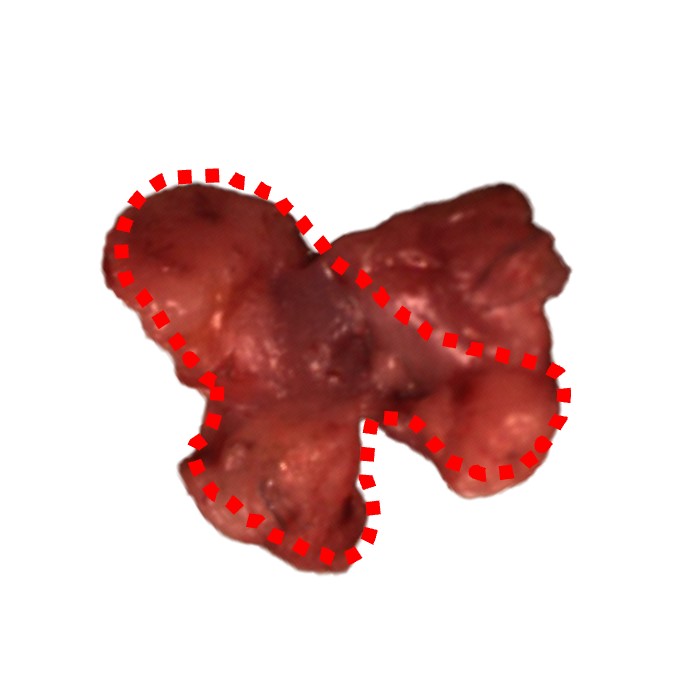}  
        \\
\bottomrule
    \end{tabular}
    % \label{appendix_mice_roi_tumor_and_healthy}
    \caption{Represented examples of murine tumor and control (healthy) samples for OS- and STS- models (red labels: potential tumorous regions)}
    \label{appendix_mice_roi_tumor_and_healthy}
\end{table}

\clearpage
\newpage

\end{document}